\documentclass{template}

\newcommand{\templateoption}{option1}
    
\usepackage[T1]{fontenc}
\usepackage{lmodern}
\usepackage{xspace}

\usepackage{graphicx,amsmath,amssymb,hyperref}
\usepackage[authoryear,round]{natbib}
\usepackage{xcolor}

\usepackage{subcaption}
\usepackage{booktabs}
\usepackage{longtable}
\usepackage{xltabular}
\usepackage{array}
\usepackage{colortbl}
\usepackage{eqparbox}
\usepackage{pgf}
\usepackage{float}
\usepackage{placeins}
\usepackage{bm}
\usepackage{multirow}
\usepackage[most]{tcolorbox}
\usepackage{mathtools}
\usepackage{amsthm}
\usepackage{enumitem}
\usepackage[capitalize,noabbrev]{cleveref}

\theoremstyle{plain}

\theoremstyle{definition}

\theoremstyle{remark}

\newcommand{\pdelta}{\%\Delta}

\definecolor{posgreen}{HTML}{1A6F1A}
\definecolor{negred}{HTML}{C00000}
\definecolor{neutral}{HTML}{434343}
\definecolor{heatgreen}{HTML}{34a853}
\definecolor{heatred}{HTML}{ea4335}
\definecolor{tabgreen}{HTML}{009B56}
\definecolor{tabred}{HTML}{ea4335}
\definecolor{LightGray}{gray}{0.975}
\definecolor{finding1border}{HTML}{1e3a5f}
\definecolor{finding2border}{HTML}{5c1a2a}
\definecolor{finding3border}{HTML}{3d1a5c}
\definecolor{finding1shadow}{HTML}{8ca6cf}
\definecolor{finding2shadow}{HTML}{c87fa3}
\definecolor{finding3shadow}{HTML}{a591bb}

\newlength{\valboxwidth}
\newcommand{\val}[5]{
  \pgfmathsetmacro{\trueval}{#2}%
  \def\valcontent{%
    \makebox[\valboxwidth][c]{%
      \begin{tabular}[t]{@{}c@{}}
        \rule{0pt}{9pt}{#1\%}%
        \\[-2.5pt]{\scriptsize\color[HTML]{555555}(#3)}\rule[-3pt]{0pt}{3pt}%
      \end{tabular}%
    }%
  }%
  {\setlength{\fboxsep}{1pt}%
  \ifdim\trueval pt>0pt
    \pgfmathtruncatemacro{\intensity}{min(\trueval / #4 * 50, 50)}%
    \edef\docolorbox{\noexpand\colorbox{heatgreen!\intensity!white}}\docolorbox{\valcontent}%
  \else
    \pgfmathtruncatemacro{\intensity}{min(abs(\trueval) / #5 * 25, 25)}%
    \edef\docolorbox{\noexpand\colorbox{heatred!\intensity!white}}\docolorbox{\valcontent}%
  \fi}%
}

\newcommand{\valb}[5]{%
  \pgfmathsetmacro{\trueval}{#2}%
  \def\valcontent{%
    \makebox[\valboxwidth][c]{%
      \begin{tabular}[t]{@{}c@{}}
        \rule{0pt}{9pt}{\textbs{#1\%}}%
        \\[-2.5pt]{\scriptsize\color[HTML]{555555}(#3)}\rule[-3pt]{0pt}{3pt}%
      \end{tabular}%
    }%
  }%
  {\setlength{\fboxsep}{1pt}%
  \ifdim\trueval pt>0pt
    \pgfmathtruncatemacro{\intensity}{min(\trueval / #4 * 50, 50)}%
    \edef\docolorbox{\noexpand\colorbox{heatgreen!\intensity!white}}\docolorbox{\valcontent}%
  \else
    \pgfmathtruncatemacro{\intensity}{min(abs(\trueval) / #5 * 25, 25)}%
    \edef\docolorbox{\noexpand\colorbox{heatred!\intensity!white}}\docolorbox{\valcontent}%
  \fi}%
}

\newcommand{\valu}[5]{%
  \pgfmathsetmacro{\trueval}{#2}%
  \def\valcontent{%
    \makebox[\valboxwidth][c]{%
      \begin{tabular}[t]{@{}c@{}}
        \rule{0pt}{9pt}{\underline{#1\%}}%
        \\[-2.5pt]{\scriptsize\color[HTML]{555555}(#3)}\rule[-3pt]{0pt}{3pt}%
      \end{tabular}%
    }%
  }%
  {\setlength{\fboxsep}{1pt}%
  \ifdim\trueval pt>0pt
    \pgfmathtruncatemacro{\intensity}{min(\trueval / #4 * 50, 50)}%
    \edef\docolorbox{\noexpand\colorbox{heatgreen!\intensity!white}}\docolorbox{\valcontent}%
  \else
    \pgfmathtruncatemacro{\intensity}{min(abs(\trueval) / #5 * 25, 25)}%
    \edef\docolorbox{\noexpand\colorbox{heatred!\intensity!white}}\docolorbox{\valcontent}%
  \fi}%
}

\newcommand{\valbu}[5]{%
  \pgfmathsetmacro{\trueval}{#2}%
  \def\valcontent{%
    \makebox[\valboxwidth][c]{%
      \begin{tabular}[t]{@{}c@{}}
        \rule{0pt}{9pt}{\underline{\textbs{#1\%}}}%
        \\[-2.5pt]{\scriptsize\color[HTML]{555555}(#3)}\rule[-3pt]{0pt}{3pt}%
      \end{tabular}%
    }%
  }%
  {\setlength{\fboxsep}{1pt}%
  \ifdim\trueval pt>0pt
    \pgfmathtruncatemacro{\intensity}{min(\trueval / #4 * 50, 50)}%
    \edef\docolorbox{\noexpand\colorbox{heatgreen!\intensity!white}}\docolorbox{\valcontent}%
  \else
    \pgfmathtruncatemacro{\intensity}{min(abs(\trueval) / #5 * 25, 25)}%
    \edef\docolorbox{\noexpand\colorbox{heatred!\intensity!white}}\docolorbox{\valcontent}%
  \fi}%
}

\newcommand{\valna}[4]{
  \pgfmathsetmacro{\trueval}{#2}%
  \def\valcontent{\makebox[\valboxwidth][c]{\rule{0pt}{10pt}{#1\%}\rule[-4pt]{0pt}{4pt}}}%
  {\setlength{\fboxsep}{1pt}%
  \ifdim\trueval pt>0pt
    \pgfmathtruncatemacro{\intensity}{min(\trueval / #3 * 50, 50)}%
    \edef\docolorbox{\noexpand\colorbox{heatgreen!\intensity!white}}\docolorbox{\valcontent}%
  \else
    \pgfmathtruncatemacro{\intensity}{min(abs(\trueval) / #4 * 65, 65)}%
    \edef\docolorbox{\noexpand\colorbox{heatred!\intensity!white}}\docolorbox{\valcontent}%
  \fi}%
}
\newcommand{\valnab}[4]{
  \pgfmathsetmacro{\trueval}{#2}%
  \def\valcontent{\makebox[\valboxwidth][c]{\rule{0pt}{10pt}{\textbf{#1\%}}\rule[-4pt]{0pt}{4pt}}}%
  {\setlength{\fboxsep}{1pt}%
  \ifdim\trueval pt>0pt
    \pgfmathtruncatemacro{\intensity}{min(\trueval / #3 * 50, 50)}%
    \edef\docolorbox{\noexpand\colorbox{heatgreen!\intensity!white}}\docolorbox{\valcontent}%
  \else
    \pgfmathtruncatemacro{\intensity}{min(abs(\trueval) / #4 * 65, 65)}%
    \edef\docolorbox{\noexpand\colorbox{heatred!\intensity!white}}\docolorbox{\valcontent}%
  \fi}%
}
\newcommand{\valnau}[4]{
  \pgfmathsetmacro{\trueval}{#2}%
  \def\valcontent{\makebox[\valboxwidth][c]{\rule{0pt}{10pt}{\underline{#1\%}}\rule[-4pt]{0pt}{4pt}}}%
  {\setlength{\fboxsep}{1pt}%
  \ifdim\trueval pt>0pt
    \pgfmathtruncatemacro{\intensity}{min(\trueval / #3 * 50, 50)}%
    \edef\docolorbox{\noexpand\colorbox{heatgreen!\intensity!white}}\docolorbox{\valcontent}%
  \else
    \pgfmathtruncatemacro{\intensity}{min(abs(\trueval) / #4 * 65, 65)}%
    \edef\docolorbox{\noexpand\colorbox{heatred!\intensity!white}}\docolorbox{\valcontent}%
  \fi}%
}
\newcommand{\valnabu}[4]{
  \pgfmathsetmacro{\trueval}{#2}%
  \def\valcontent{\makebox[\valboxwidth][c]{\rule{0pt}{10pt}{\underline{\textbf{#1\%}}}\rule[-4pt]{0pt}{4pt}}}%
  {\setlength{\fboxsep}{1pt}%
  \ifdim\trueval pt>0pt
    \pgfmathtruncatemacro{\intensity}{min(\trueval / #3 * 50, 50)}%
    \edef\docolorbox{\noexpand\colorbox{heatgreen!\intensity!white}}\docolorbox{\valcontent}%
  \else
    \pgfmathtruncatemacro{\intensity}{min(abs(\trueval) / #4 * 65, 65)}%
    \edef\docolorbox{\noexpand\colorbox{heatred!\intensity!white}}\docolorbox{\valcontent}%
  \fi}%
}

\newcommand{\valpha}[2]{%
  \makebox[0pt][l]{#1}\makebox[3.3em][r]{\raisebox{-6pt}{\scriptsize\textcolor{gray}{($\alpha$=#2)}}}%
}

\definecolor{heatgrey}{HTML}{6b7280}
\definecolor{rowpurple}{HTML}{f5f3ff}
\definecolor{rowblue}{HTML}{eff6ff}
\definecolor{rowgreen}{HTML}{f0fdf4}
\definecolor{rowyellow}{HTML}{fefce8}
\definecolor{roworange}{HTML}{fff7ed}
\definecolor{rowpink}{HTML}{fdf2f8}
\definecolor{rowcyan}{HTML}{ecfeff}
\definecolor{rowlime}{HTML}{f7fee7}
\definecolor{rowviolet}{HTML}{f5f3ff}
\definecolor{rowrose}{HTML}{fff1f2}
\definecolor{rowteal}{HTML}{f0fdfa}
\definecolor{rowamber}{HTML}{fffbeb}
\definecolor{rowsky}{HTML}{f0f9ff}
\definecolor{rowfuchsia}{HTML}{fdf4ff}
\definecolor{rowemerald}{HTML}{ecfdf5}
\definecolor{rowindigo}{HTML}{eef2ff}
\definecolor{rowred}{HTML}{fef2f2}
\definecolor{rowmint}{HTML}{f0fff4}
\definecolor{rowpeach}{HTML}{fff5f5}
\definecolor{rowlavender}{HTML}{f8f5ff}
\definecolor{rowapricot}{HTML}{fff8e7}
\definecolor{rowaqua}{HTML}{e6ffff}
\definecolor{rowcoral}{HTML}{fff0ed}
\definecolor{rowperiwinkle}{HTML}{e8e8ff}
\definecolor{rowchampagne}{HTML}{fff8e1}
\definecolor{rowgray}{HTML}{f9fafb}

\newlength{\cvalwidth}
\newlength{\cvalpwidth}
\newcommand{\cval}[3]{
  \pgfmathsetmacro{\cellval}{#1}%
  \pgfmathtruncatemacro{\intensity}{(#3 - \cellval) / (#3 - #2) * 45}%
  {\setlength{\fboxsep}{0pt}%
  \edef\docolorbox{\noexpand\colorbox{heatgrey!\intensity!white}}\docolorbox{\makebox[\cvalwidth][c]{\rule{0pt}{9.3pt}#1\rule[-3.3pt]{0pt}{3.3pt}}}}%
}
\newcommand{\cvalp}[3]{
  \pgfmathsetmacro{\cellval}{#1}%
  \pgfmathtruncatemacro{\intensity}{(\cellval - #2) / (#3 - #2) * 45}%
  {\setlength{\fboxsep}{0pt}%
  \edef\docolorbox{\noexpand\colorbox{heatgrey!\intensity!white}}\docolorbox{\makebox[\cvalpwidth][c]{\rule{0pt}{9.3pt}#1\%\rule[-3.3pt]{0pt}{3.3pt}}}}%
}

\definecolor{link}{HTML}{307ef1}
\hypersetup{
  colorlinks=true,
  urlcolor=link,
  linkcolor=link,
  citecolor=link,
  filecolor=link,
  pdfborder={0 0 0}
}

\providecommand{\equationname}{Equation}
\providecommand{\sectionname}{\S}

\newcommand{\figref}[1]{%
  \figurename~\hyperref[#1]{\textcolor{link}{\ref*{#1}}}%
}
\newcommand{\tabref}[1]{%
  \tablename~\hyperref[#1]{\textcolor{link}{\ref*{#1}}}%
}
\newcommand{\eqrefc}[1]{%
  \equationname~\hyperref[#1]{\textcolor{link}{\ref*{#1}}}%
}
\newcommand{\secrefc}[1]{%
  \sectionname \hyperref[#1]{\textcolor{link}{\ref*{#1}}}%
}

\makeatletter
\fancypagestyle{firststyle}{
  \fancyhead[L]{}
  \fancyhead[C]{}
  \fancyhead[R]{}
  \fancyfoot[L]{\footerfont \the\correspondingauthor}
  \fancyfoot[R]{}
}
\makeatother

\makeatletter
\DeclareRobustCommand\bfseries{%
  \not@math@alphabet\bfseries\mathbf
  \fontseries\bfdefault\selectfont
  \sffamily
}
\makeatother

\DeclareTextFontCommand{\textbf}{\bfseries\sffamily}
\newcommand{\textbs}[1]{%
  {\normalfont\fontfamily{\rmdefault}\fontseries{b}\selectfont #1}%
}

\title{\centering \fontsize{18}{24}\selectfont{Strong Teacher Not Needed? \\ On Distillation in LLM Pretraining}}

\author{
    \vspace{.2cm}
    \parbox{\textwidth}{\centering
        \textbs{Taiming Lu} \quad \quad
        \textbs{Zhuang Liu}
    }
    \\
    \vspace{-0.3cm}
    Princeton University
}

\makeatletter
\renewcommand{\abscontent}{}%
\makeatother

\newenvironment{abstractblock}{%
  {\centering\large\bfseries\sffamily Abstract\par}
  \vspace{0.2em}
  \begin{list}{}{%
      \setlength{\leftmargin}{2em}
      \setlength{\rightmargin}{2em}
      \setlength{\topsep}{0pt}
      \setlength{\parsep}{0pt}
  }
  \item[]
}{%
  \end{list}
  \par\normalfont\vspace{1em}
}

\begin{document}

\begingroup
\makeatletter
\let\raggedright\centering
\makeatother

\maketitle
\endgroup

\newcommand{\abstractcontent}{%
Knowledge distillation generally assumes a strong-to-weak relationship where stronger teachers yield better students. In this work, we examine this assumption about distillation in large language model pretraining. By varying architecture sizes and training token budgets, we create strong-to-weak, same-level, and weak-to-strong teacher-student relationships, and study distillation's effectiveness under each. We find that the teacher need not be strong: with proper mixing of the language modeling and knowledge distillation losses, even small and undertrained teachers improve larger students. At the same time, a stronger teacher is not always better: pushing the teacher further, through more parameters or more training tokens, can saturate or even reverse the distillation gains. We further observe that distillation improves generalization (out-of-distribution and downstream performance) more readily than in-domain fitting. Together, these results challenge the common belief that distillation pretraining always requires a strong teacher. Our code is avaliable \href{https://github.com/zlab-princeton/strong-distill}{here}.
}

\newcommand{\teaserfigure}[2]{%
    \begin{figure}[#1]
        \centering
        \includegraphics[width=#2,trim=0 70pt 0 23pt,clip]{pics/teaser_v6.pdf}
        \vspace{-2pt}
        \makebox[\textwidth][l]{\hspace*{0.005\textwidth}%
        \begin{minipage}[t]{0.23\textwidth}\centering\small \textbf{(a)} A smaller teacher can still guide a larger student.\end{minipage}\hspace{0.022\textwidth}
        \begin{minipage}[t]{0.20\textwidth}\centering\small \textbf{(b)} A teacher can help a same-level student.\end{minipage}\hspace{0.028\textwidth}
        \begin{minipage}[t]{0.24\textwidth}\centering\small \textbf{(c)} A compatible strong teacher helps the most.\end{minipage}\hspace{0.012\textwidth}
        \begin{minipage}[t]{0.24\textwidth}\centering\small \textbf{(d)} An overly strong teacher can overwhelm the student.\end{minipage}}
        \vspace{.3cm}
        \caption{\textbf{Effective distillation in LLM pretraining depends on teacher--student compatibility, not strong teacher.} Useful supervision can come from a smaller, same-level, stronger, or even much stronger teacher, but the resulting gains depend on how the teacher fits the student. This paper systematically studies this and shows that weak-to-strong and same-level distillation can improve pretraining, while stronger teachers can saturate and underperform.}
        \label{fig:new_teaser}
    \end{figure}%
}


\makeatletter
\ifthenelse{\equal{\templateoption}{option1}}{
    \vspace{-0.1cm}
    \teaserfigure{!ht}{\textwidth}
    \vspace{0.3cm}
    \begin{abstractblock}
    \abstractcontent
    \end{abstractblock}
    \newpage
}{
    \ifthenelse{\equal{\templateoption}{option2}}{
        \vspace{-0.3cm}
        \begin{abstractblock}
        \abstractcontent
        \end{abstractblock}
        \vspace{0.0cm}
        \teaserfigure{!bh}{.65\textwidth}
        \newpage
    }{
        \begin{abstractblock}
        \abstractcontent
        \end{abstractblock}
        \vspace{0.5cm}
        \vspace{-0.5cm}
    }
}
\makeatother


\section{Introduction}
\label{sec:intro}

\vspace{.1cm}

A long-standing assumption governs knowledge distillation: a stronger teacher yields a better student~\citep{hintonKD,ModelCompression}. When a large model is compressed into a smaller one, the more capable the teacher, the more knowledge there is available to transfer~\citep{KD_Efficacy,generativeKD}. Across a decade of work spanning both vision~\citep{FitNets,Att2Att,RDK} and language modeling~\citep{distillbert,tinyBERT,PKD,MiniLM,seq2seq}, this strong-to-weak framing has shaped teacher selection~\citep{KDscalinglaw}, loss design~\citep{DAKD,RLKD,AdaKD}, and the broader premise that distillation serves primarily as a compression technique~\citep{minitron,TinyLlama}.

\vspace{.1cm}

Our experiments suggest this account is incomplete. In large language model (LLM) pretraining, we find that weaker teachers can in fact improve stronger students, and that continuing to strengthen a teacher, through additional parameters or additional training tokens, can degrade rather than improve the resulting student.

\vspace{.1cm}

The stakes extend well beyond theoretical interest. As LLMs continue to scale, two practical scenarios make the teacher-student relationship increasingly consequential. First, when training frontier models~\citep{llama3,qwen3,deepseek-R1}, the best available teacher is often drawn from the previous generation, trained with a smaller architecture or less compute than the target student~\citep{w2s_generalization}. Second, alongside the push toward larger models, demand persists for small, efficient systems~\citep{Phi-3,MobileLLM} capable of serving lightweight tasks at low latency~\citep{SpeculativeDecoding}.

\vspace{.1cm}

Under the conventional strong-to-weak view, the former scenario forecloses distillation entirely, while the latter~\citep{OpenELM,TinyLlama} reduces teacher selection to a single rule: the larger and more trained, the better~\citep{minitron}. Whether these assumptions hold determines both how broadly distillation can be applied in LLM development and how the teacher should be chosen in practice.

Despite its centrality, the strong-to-weak view has remained largely unchallenged in the context of LLM pretraining~\citep{ModelCompression}: weaker teachers are presumed unable to meaningfully guide stronger students, and scaling up the teacher is presumed to monotonically improve the student. Prior work on LLM distillation~\citep{seq2seq,PKD,MiniLM,Self-Instruct,distillbert,tinyBERT} has not systematically examined these assumptions across the full range of teacher configurations. 

\vfill
\begin{center}
    \includegraphics[width=\textwidth]{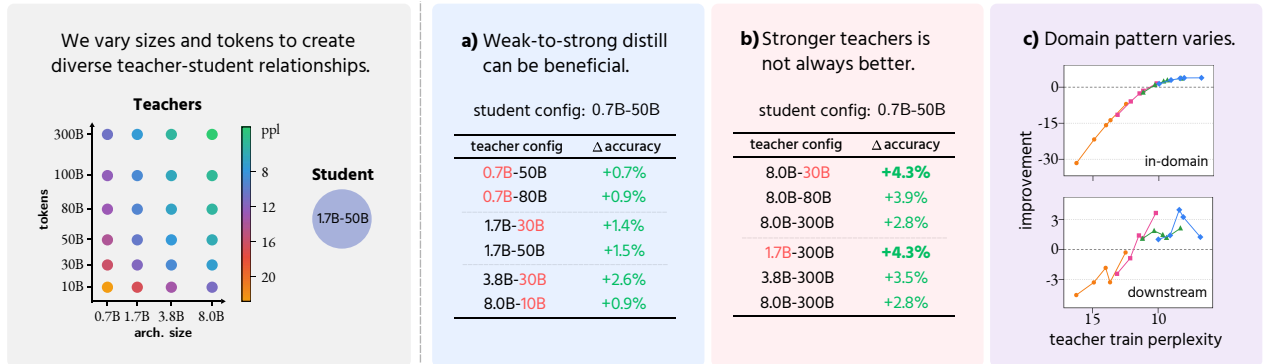}
    \captionof{figure}{\textbf{A strong teacher is not always needed in distillation pretraining.} We vary architecture size and training tokens to create diverse teacher-student relationships. We have three key findings. \textbf{a)} Weak-to-strong and same-level distillation can be beneficial: teachers with smaller architecture or fewer tokens (marked red in figure) than the student still yield improvement. \textbf{b)} Stronger teachers do not always yield better students: at high token budgets, same-size teachers outperform larger ones; at large sizes, teachers trained with fewer tokens outperform those trained with more. \textbf{c)} Different evaluation domains show different patterns: in-domain improvement follows a universal trend and is hardest to improve, while out-of-distribution and downstream exhibit architecture-specific or noisy patterns and are easier to improve, suggesting distillation pretraining transfers useful knowledge for generalization instead of fitting.}
    \label{fig:teaser}
    \vspace{.7cm}
\end{center}

\newpage

We confront them directly by asking two central questions:

\vspace{-.3cm}
\begingroup
  \setlength{\leftmargini}{0.8em}
  \begin{quote}
      \centering
      \textit{Can weak-to-strong distillation be helpful? 
      Does a stronger teacher always yield a better student?}
  \end{quote}
\endgroup
\vspace{-.3cm}


To address these questions, we conduct a systematic study of teacher-student relationships (see \figref{fig:teaser}). We vary teacher and student architecture sizes, teacher training token budgets, and the loss mixing coefficient, spanning strong-to-weak, same-level, and weak-to-strong configurations. Our findings are threefold:
\begin{itemize}[leftmargin=*, topsep=0pt, itemsep=3pt, parsep=0pt, partopsep=0pt]
\item \textbf{Weak teachers can improve strong students.} With appropriate loss mixing, teachers smaller and less trained than the student still yield measurable gains.
\item \textbf{Stronger teachers do not always produce better students.} Architecture mismatch and teacher overtraining can degrade transfer even when the teacher attains a lower training loss.
\item \textbf{Generalization improves more readily than in-domain fit.} Out-of-distribution perplexity and downstream accuracy exhibit larger and more consistent gains than in-domain perplexity.
\end{itemize}
Collectively, these findings unsettle the prevailing account of distillation in LLM pretraining and offer concrete implications for which teacher to select and which loss-mixing to use under realistic compute constraints.



\section{Background}
\label{sec:background}

We review the standard autoregressive language modeling and knowledge distillation (KD) objectives. 
Given a token sequence $x = (x_1,\ldots,x_T)$ drawn from a pretraining corpus $\mathcal{D}$, a student model with parameters $\theta_S$ defines a next-token distribution
$p_{\theta_S}(\cdot \mid x_{<t})$, and a teacher model with parameters $\theta_T$ defines
$p_{\theta_T}(\cdot \mid x_{<t})$.
We formalize the student's training objectives in standard and distillation pretraining below.

\paragraph{Autoregressive language modeling.}
Standard pretraining learns $\theta_S$ by maximizing the likelihood of the observed next token given a preceding context, as in \eqrefc{eq:lm_loss}. We refer to this as the language modeling loss (LM loss), the cross-entropy between the student distribution and a one-hot target (the ground-truth next token). This objective trains the student to model the distribution of the training corpus.

\begin{equation}
\mathcal{L}_{\mathrm{lm}}(\theta_S)
= \mathbb{E}_{x \sim \mathcal{D}}
\left[
\frac{1}{T}\sum_{t=1}^{T}
-\log p_{\theta_S}(x_t \mid x_{<t})
\right]
\label{eq:lm_loss}
\end{equation}

\paragraph{Knowledge distillation for language modeling.}
In distillation pretraining~\citep{hintonKD}, the target is no longer a one-hot vector but the teacher's relative preferences among all tokens. The teacher provides a full distribution over the vocabulary for the same context $x_{<t}$. Let
$q_t(\cdot) := p_{\theta_T}(\cdot \mid x_{<t})$ and $p_t(\cdot) := p_{\theta_S}(\cdot \mid x_{<t}).
$
Student distribution $p_t$ will try to match the teacher distribution $q_t$ by minimizing the forward KL divergence, as in \eqrefc{eq:kd_loss}. We refer to this as the knowledge distillation loss (KD loss). Intuitively, KD trains the student to imitate not only the teacher's top prediction, but also relative preferences among alternative tokens. 

\begin{equation}
\mathcal{L}_{\mathrm{kd}}(\theta_S; \theta_T)=
\mathbb{E}_{x \sim \mathcal{D}}
\left[
\frac{1}{T}\sum_{t=1}^{T}
\mathrm{KL}\!\left(q_t \,\|\, p_t\right)
\right]
\label{eq:kd_loss}
\end{equation}

\paragraph{Mixing ground-truth and teacher supervision.}
Distillation typically combines the standard language modeling loss with the distillation loss~\citep{hintonKD}, as in \eqrefc{eq:mixed_objective}. 
The loss mixing coefficient $\alpha$ controls the trade-off:
$\alpha=0$ recovers standard pretraining, while larger $\alpha$ places more weight on matching the teacher distribution. In practice, $\alpha$ determines how much the student follows the teacher compared to the ground-truth data signal. We refer to this as the mixed loss and use it for all distillation pretraining in this work.

\begin{equation}
\mathcal{L}(\theta_S; \theta_T)=
(1-\alpha)\,\mathcal{L}_{\mathrm{lm}}(\theta_S)
+
\alpha\,\mathcal{L}_{\mathrm{kd}}(\theta_S; \theta_T), \quad \alpha \in [0,1]
\label{eq:mixed_objective}
\end{equation}

\newpage

\section{Experimental Design}
\label{sec:design}

\vspace{-.1cm}


\vspace{-0cm}

\subsection{Teacher-Student Configurations}
\label{sec:config}
\vspace{-.3cm}

\paragraph{Architecture sizes.}
We use the Llama3~\citep{llama3} architectures. To enable systematic comparison across sizes, we create four architecture variants (see \secrefc{appendix:archs_details}), resulting in four model sizes: 0.7B, 1.7B, 3.8B, and 8.0B parameters.
Given a student of a particular size, the teacher can be larger, the same size, or smaller.

\vspace{-.15cm}

\paragraph{Training token budgets.}
Teachers are pretrained on varying amounts of data: 10B, 30B, 50B, 80B, 100B, or 300B tokens. Students are trained on all 50B tokens for distillation pretraining. This creates another dimension of variation: the teacher sees more, the same number, or fewer tokens than the student.

\vspace{-.15cm}

\paragraph{Teacher-student relationships.}
\sloppy
Together, size and tokens create a spectrum of teacher-student relationships:
\vspace{-.1cm}
\begin{itemize}[leftmargin=2em, labelsep=0.2em, topsep=0pt, itemsep=1pt, parsep=0pt, partopsep=0pt]
\item \underline{Architecture}: large-to-small, same-size, or small-to-large,
\item \underline{Tokens}: more-to-less, same-number, or less-to-more,
\item \underline{Train loss}: lower-to-higher or higher-to-lower, defined by teacher train loss compared to student baseline.
\end{itemize}
\vspace{-.1cm}
These dimensions do not always align. A smaller architecture trained on more tokens may achieve lower train loss than a larger one trained on fewer tokens. A same-size teacher trained longer may outperform a larger one trained briefly.
This allows us to disentangle the effects of architecture capacity from training compute in knowledge transfer. More broadly, we refer to a teacher-student relationship as \textit{strong-to-weak} when a teacher is stronger than the student (by any dimensions above), \textit{same-level} if identical, and \textit{weak-to-strong} for the opposite. A single relationship may be strong-to-weak in one dimension and weak-to-strong in another.

\newlength{\catskip}\setlength{\catskip}{14pt}

\begin{table*}[t]
\centering
\small
\setlength{\tabcolsep}{0pt}
\setlength{\abovetopsep}{0pt}
\setlength{\belowbottomsep}{0pt}
\setlength{\aboverulesep}{0pt}
\setlength{\belowrulesep}{0pt}
\begin{tabular*}{\textwidth}{@{\hspace{\catskip}\extracolsep{\fill}} r @{\hspace{1.5\catskip}} c@{}c@{}c@{}c @{\hspace{1.3\catskip}} c@{}c@{}c@{}c @{\hspace{1.3\catskip}} c@{}c@{}c@{}c @{\hspace{\catskip}}}
\toprule
\rule[-3pt]{0pt}{12pt}
 & \multicolumn{4}{c}{\hspace{-18pt}in-domain ppl ($\downarrow$)} & \multicolumn{4}{c}{\hspace{-23pt}out-of-distribution ppl avg. ($\downarrow$)} & \multicolumn{4}{c}{\hspace{-21pt}downstream acc avg. ($\uparrow$)} \\
\cmidrule(l{0pt}r{23pt}){2-5} \cmidrule(l{0pt}r{23pt}){6-9} \cmidrule(l{0pt}r{18pt}){10-13}
\rule{0pt}{9pt}& 0.7B & 1.7B & 3.8B & 8.0B & 0.7B & 1.7B & 3.8B & 8.0B & 0.7B & 1.7B & 3.8B & 8.0B \\[1pt]
\midrule
\multicolumn{13}{@{}l@{}}{{\setlength{\fboxsep}{0pt}\colorbox{rowpurple}{\makebox[\textwidth][l]{\hspace{0.6\catskip}\rule[-3.5pt]{0pt}{12pt}$\rightarrow$ \textit{standard pretrained teachers}}}}} \\
10B & \cval{21.2}{6.8}{21.2} & \cval{16.8}{6.8}{21.2} & \cval{14.4}{6.8}{21.2} & \cval{12.8}{6.8}{21.2} & \cval{39.8}{10.1}{39.8} & \cval{26.3}{10.1}{39.8} & \cval{22.0}{10.1}{39.8} & \cval{18.5}{10.1}{39.8} & \cvalp{34.9}{34.9}{46.6} & \cvalp{37.0}{34.9}{46.6} & \cvalp{38.5}{34.9}{46.6} & \cvalp{39.4}{34.9}{46.6} \\
30B & \cval{16.3}{6.8}{21.2} & \cval{13.1}{6.8}{21.2} & \cval{11.2}{6.8}{21.2} & \cval{10.0}{6.8}{21.2} & \cval{23.6}{10.1}{39.8} & \cval{17.5}{10.1}{39.8} & \cval{14.4}{10.1}{39.8} & \cval{12.8}{10.1}{39.8} & \cvalp{37.8}{34.9}{46.6} & \cvalp{39.4}{34.9}{46.6} & \cvalp{41.5}{34.9}{46.6} & \cvalp{42.7}{34.9}{46.6} \\
50B & \cval{14.9}{6.8}{21.2} & \cval{12.1}{6.8}{21.2} & \cval{10.3}{6.8}{21.2} & \cval{9.1}{6.8}{21.2} & \cval{20.6}{10.1}{39.8} & \cval{15.6}{10.1}{39.8} & \cval{13.0}{10.1}{39.8} & \cval{11.7}{10.1}{39.8} & \cvalp{38.1}{34.9}{46.6} & \cvalp{39.9}{34.9}{46.6} & \cvalp{42.9}{34.9}{46.6} & \cvalp{43.7}{34.9}{46.6} \\
80B & \cval{14.0}{6.8}{21.2} & \cval{11.5}{6.8}{21.2} & \cval{9.6}{6.8}{21.2} & \cval{8.4}{6.8}{21.2} & \cval{18.4}{10.1}{39.8} & \cval{14.5}{10.1}{39.8} & \cval{11.9}{10.1}{39.8} & \cval{11.0}{10.1}{39.8} & \cvalp{38.5}{34.9}{46.6} & \cvalp{41.9}{34.9}{46.6} & \cvalp{43.6}{34.9}{46.6} & \cvalp{44.9}{34.9}{46.6} \\
100B & \cval{13.7}{6.8}{21.2} & \cval{11.2}{6.8}{21.2} & \cval{9.4}{6.8}{21.2} & \cval{8.1}{6.8}{21.2} & \cval{17.8}{10.1}{39.8} & \cval{13.9}{10.1}{39.8} & \cval{11.6}{10.1}{39.8} & \cval{10.6}{10.1}{39.8} & \cvalp{38.8}{34.9}{46.6} & \cvalp{41.9}{34.9}{46.6} & \cvalp{43.7}{34.9}{46.6} & \cvalp{45.2}{34.9}{46.6} \\
300B & \cval{12.5}{6.8}{21.2} & \cval{10.2}{6.8}{21.2} & \cval{8.3}{6.8}{21.2} & \cval{6.8}{6.8}{21.2} & \cval{16.1}{10.1}{39.8} & \cval{12.8}{10.1}{39.8} & \cval{10.9}{10.1}{39.8} & \cval{10.1}{10.1}{39.8} & \cvalp{40.1}{34.9}{46.6} & \cvalp{43.2}{34.9}{46.6} & \cvalp{45.4}{34.9}{46.6} & \cvalp{46.6}{34.9}{46.6} \\
\midrule
\multicolumn{13}{@{}l@{}}{{\setlength{\fboxsep}{0pt}\colorbox{rowblue}{\makebox[\textwidth][l]{\hspace{0.6\catskip}\rule[-3.5pt]{0pt}{12pt}$\rightarrow$ \textit{standard pretrained student baseline}}}}} \\
50B & \cval{14.9}{6.8}{21.2} & \cval{12.2}{6.8}{21.2} & \cval{10.3}{6.8}{21.2} & \cval{9.0}{6.8}{21.2} & \cval{20.6}{10.1}{39.8} & \cval{15.7}{10.1}{39.8} & \cval{13.0}{10.1}{39.8} & \cval{11.7}{10.1}{39.8} & \cvalp{38.1}{34.9}{46.6} & \cvalp{40.1}{34.9}{46.6} & \cvalp{42.2}{34.9}{46.6} & \cvalp{43.5}{34.9}{46.6} \\
\bottomrule
\end{tabular*}
\caption{\textbf{Performance of standard pretrained teacher models and student baselines.} Rows indicate training token budget; columns indicate architecture size. In our main experiments, student distillation pretraining (and thus the baselines) is fixed to 50B tokens. We report in-domain perplexity (ppl) on held-out FineWeb-Edu (lower is better), average out-of-distribution perplexity across 11 corpora (lower is better), and average downstream accuracy (acc) across 15 benchmarks (higher is better). Teacher performance consistently improves with both architecture size and training tokens, with minor variance in downstream accuracy. View full results at \secrefc{appendix:eval}.}
\label{tab:teachers_student_baseline}
\vspace{-.5cm}
\end{table*}

\vspace{-.2cm}

\subsection{Training Details} \label{sec:training}
\vspace{-.3cm}

\paragraph{Data and tokenization.} All models are trained on FineWeb-Edu~\citep{fineweb}. Teacher pretraining and student distillation pretraining draw from the same data pool with different random seeds, so the sequences seen may overlap but are not identical. For each run, no data is repeated during training per run. 

\vspace{-.15cm}

\paragraph{Hyperparameters.} We use an 8192-token context length and a 256-sequence batch size. We train with AdamW using $\beta_1=0.9$, $\beta_2=0.95$, and weight decay 0.1. The peak learning rate is $3 \times 10^{-4}$ with a cosine decay to 10\% of the peak, linear warm-up for the first 5\% of steps, gradient clipping of 1.0, and bf16 precision.

\vspace{-.15cm}

\paragraph{Distillation pretraining.} During distillation, the teacher model is kept frozen and provides logits only. We use a temperature of $\tau=1$ for the knowledge distillation loss to apply no scaling to the logits. For each teacher-student configuration, we search over the loss mixing coefficient $\alpha \in \{0.2, 0.4, 0.5, 0.6, 0.8, 1.0\}$.

\vspace{-.15cm}

\paragraph{Baseline.} The standard pretraining baseline for each student corresponds to training with $\alpha = 0$ (language modeling loss only) under the same data ordering and random seed as distillation pretraining. This ensures that any improvement from distillation is attributable to the teacher signal rather than confounding factors.

\subsection{Evaluations}
\label{sec:eval}
We evaluate distillation effectiveness along three dimensions: in-domain perplexity, out-of-distribution perplexity, and downstream accuracy for a broad range of performance (see \secrefc{appendix:eval_imp} for full evaluation metrics).

\paragraph{In-domain perplexity.}
We measure in-domain perplexity (id ppl) on a held-out test split from training dataset FineWeb-Edu, which was not seen during training. Lower perplexity indicates better fit to the training distribution. This evaluates how well the student models the same distribution as the training data.

\paragraph{Out-of-domain perplexity.}
We also measure out-of-distribution perplexity (ood ppl) on 11 high-quality corpora: Wikitext-103~\citep{merity2017pointer}, C4~\citep{C4}, GSM8K~\citep{gsm8k}, DM Mathematics~\citep{dm_math}, HumanEval~\citep{humaneval}, CodeSearchNet~\citep{codesearchnet}, arXiv~\citep{arXivSummerization}, CNN-DailyMail~\citep{CNNDailyMail}, ECHR~\citep{ECHR}, PubMedQA~\citep{PubMedQA}, XQuAD~\citep{XQuAD}.
These corpora span diverse domains, including web, news, scientific, math, code, biomedical, legal, and multilingual documents and QA, providing broad coverage of distribution shifts.
This evaluates how well the model generalizes beyond the training corpus.

\paragraph{Downstream accuracy.}
We evaluate on 15 downstream accuracy-based benchmarks (downstream acc): MMLU~\citep{mmlu}, ARC-Easy~\citep{ARC}, SciQ~\citep{SciQ}, OpenBookQA~\citep{OpenBookQA}, MathQA~\citep{MathQA}, TruthfulQA~\citep{TruthfulQA}, ANLI-R1~\citep{ANLI}, CommonsenseQA~\citep{CommonsenseQA}, HellaSwag~\citep{HellaSwag}, PIQA~\citep{PIQA}, WinoGrande~\citep{WinoGrande},  Social IQa~\citep{SocialIQA}, LogiQA 2.0~\citep{LogiQA}, MedMCQA~\citep{MedMCQA}, and RACE~\citep{RACE}. These benchmarks cover broad knowledge, science, math, and medical QA, commonsense and logical reasoning, and reading comprehension to offer a diverse view of downstream transfer, evaluating transfer for tasks beyond next-token prediction.

\paragraph{Measuring improvement.}
For each evaluation type, we compute improvement as the percentage change relative to the standard pretraining baseline. Due to the high number of benchmarks within each evaluation type, we report the average for brevity, where we average the percentage improvements across benchmarks rather than averaging raw scores, as different benchmarks have different scales.

\section{Distillation Under Different Teacher-Student Relationships}
\label{sec:results}

Before studying distillation, we validate that our teacher models are properly trained and exhibit expected scaling behavior. We train teachers across four architecture sizes (0.7B, 1.7B, 3.8B, 8.0B) and six token budgets (10B, 30B, 50B, 80B, 100B, 300B), yielding 24 teacher models.
\tabref{tab:teachers_student_baseline} reports teacher performance across all configurations. As expected, performance improves with both architecture size and training tokens.
This confirms that our teachers span a meaningful range of capabilities, from undertrained small models (0.7B at 10B tokens) to well-trained large models (8.0B at 300B tokens). We also report the student baselines trained on 50B tokens as references for measuring distillation improvement in subsequent experiments.

To study distillation's maximum strength, we analyze distillation pretraining when selecting the optimal loss mixing coefficient $\alpha$ 
for each teacher-student configuration. 
We fix the student to 1.7B trained on 50B tokens and pair it with all 24 teachers. Training curves are detailed in \secrefc{appendix:train_curve}.
We report the best $\alpha$ per evaluation type to capture differing trends.  Using a single best $\alpha$ conditioned on all evaluation types matches the same findings (see \secrefc{appendix:best_alpha_all}).
Given the high number of evaluations, we report the average improvement. Per-benchmark results follow the trend of the aggregated one (see \secrefc{appendix:eval}). 
Another common practice
is pure distillation ($\alpha=1.0$),
training only on teacher's output distribution.
Generally, we find it worse than mixed loss (see \secrefc{appendix:pure_KD}).

With the best $\alpha$, \tabref{tab:exp1_results} shows the percentage improvement over the standard pretraining baseline, and \figref{fig:exp1_trends} visualizes the trends.
\tabref{tab:exp1_results} shows that
most teacher configurations yield positive improvement across all three evaluation types, with only the weakest teachers (0.7B and 1.7B architecture with 10B tokens) degrading all performance (expected given their limited capacity). \figref{fig:exp1_trends} reveals that improvement generally increases with teacher architecture size (left column) and training tokens (middle column). Teacher's in-domain perplexity (right column), as a proxy for capability, correlates with improvement. From these, several notable deviations from the general expectation emerge.

\subsection{Weak-to-Strong and Same-Level Distillation}

\begin{tcolorbox}[before skip=18pt, after skip=10pt,
  enhanced jigsaw,
  colback=LightGray,
  colframe=finding1border,
  drop fuzzy shadow southeast={fill=finding1shadow},
  boxrule=0.9pt,
  boxsep=0.1pt,
  left=10pt,
  right=10pt,
  top=5pt,
  bottom=4pt
]
{\small\textsf{\textbf{Finding 1:} With proper language modeling and distillation loss mixing, even weak-to-strong and same-level distillation improves over
standard pretraining.}}
\end{tcolorbox}

\begin{figure*}[!t]
  \centering
  \includegraphics[width=\textwidth]{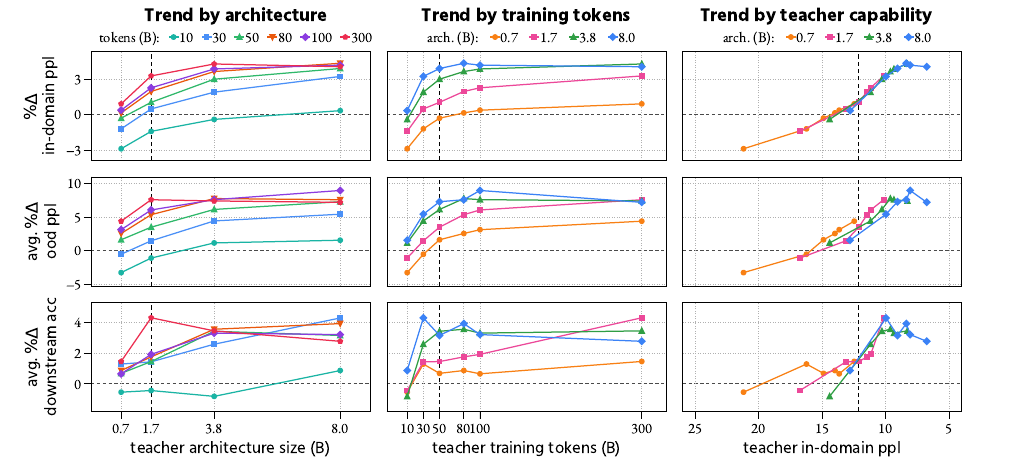}
  \vspace{-.6cm}
  \caption{\textbf{Distillation improvement under best loss mixing coefficient $\bm{\alpha}$.} Rows show evaluation types; columns show improvement along teacher architecture (left), training tokens (middle), and in-domain perplexity (right). Vertical dashed lines indicate the student's configuration (1.7B arch., 50B tokens, 12.2 ppl). Left of the dashed line represents weak-to-strong distillation in that dimension; right represents strong-to-weak. Improvement generally increases with teacher strength but saturates or reverses for strong teachers.}
  \label{fig:exp1_trends}
\end{figure*}

Even teachers weaker than the student can improve distillation. In \figref{fig:exp1_trends}, configurations to the left of the vertical dashed lines represent weak-to-strong distillation in that dimension. Across all three weak-to-strong scenarios (smaller architecture, fewer tokens, or higher train loss), most configurations yield positive improvement. Most strikingly, even teachers weaker in both architecture and tokens (thus weaker in in-domain ppl as well) can help: the 0.7B teacher at 30B tokens still achieves 1.3\% downstream accuracy improvement. This challenges the assumption that distillation requires a strictly stronger teacher. The key is using a lower loss mixing coefficient, which balances the weaker teacher's signal with ground-truth supervision.

An interesting special case is when teacher and student share the same architecture and token budget (1.7B at 50B in our setup), differing only in random seed. This configuration still improves the student across all metrics (+1.0\% in-domain ppl, +3.5\% out-of-distribution ppl, +1.5\% downstream acc), suggesting that distillation transfers knowledge beyond what architecture and data alone capture. This aligns with Born-Again Networks~\citep{BornAgain} in vision, where same-level distillation helps despite no capacity advantage.

\begin{table*}[!t]
\centering
\small
\setlength{\tabcolsep}{0pt}
\setlength{\catskip}{16pt}
\setlength{\abovetopsep}{0pt}
\setlength{\belowbottomsep}{0pt}
\setlength{\aboverulesep}{0pt}
\setlength{\belowrulesep}{0pt}
\begin{tabular*}{\textwidth}{@{\hspace{\catskip}} r @{\hspace{1.5\catskip}} c@{}c@{}c@{}c @{\hspace{1.3\catskip}} c@{}c@{}c@{}c @{\hspace{1.3\catskip}} c@{}c@{}c@{}c @{\hspace{\catskip}\extracolsep{\fill}}}
\toprule
\rule[-3pt]{0pt}{12pt}& \multicolumn{4}{c}{\hspace{-25pt}$\pdelta$~in-domain ppl ($\uparrow$)} & \multicolumn{4}{c}{\hspace{-23pt}avg. $\pdelta$~out-of-distribution ppl ($\uparrow$)} & \multicolumn{4}{c}{\hspace{-18pt}avg. $\pdelta$~downstream acc ($\uparrow$)} \\
\cmidrule(l{0pt}r{23pt}){2-5} \cmidrule(l{0pt}r{23pt}){6-9} \cmidrule(l{0pt}r{18pt}){10-13}
\rule{0pt}{9pt}& 0.7B & 1.7B & 3.8B & 8.0B & 0.7B & 1.7B & 3.8B & 8.0B & 0.7B & 1.7B & 3.8B & 8.0B \\[1pt]
\midrule
10B & \val{-2.9}{-2.9}{0.2}{4.3}{2.9} & \val{-1.4}{-1.4}{0.2}{4.3}{2.9} & \val{-0.4}{-0.4}{0.2}{4.3}{2.9} & \valb{0.3}{0.3}{0.2}{4.3}{2.9} & \val{-3.2}{-3.2}{0.2}{8.9}{3.2} & \val{-1.1}{-1.1}{0.2}{8.9}{3.2} & \val{1.2}{1.2}{0.2}{8.9}{3.2} & \valb{1.6}{1.6}{0.4}{8.9}{3.2} & \val{-0.5}{-0.5}{0.2}{4.3}{0.8} & \val{-0.4}{-0.4}{0.2}{4.3}{0.8} & \val{-0.8}{-0.8}{0.5}{4.3}{0.8} & \valb{0.9}{0.9}{0.4}{4.3}{0.8} \\[0pt]
30B & \val{-1.2}{-1.2}{0.2}{4.3}{2.9} & \val{0.5}{0.5}{0.2}{4.3}{2.9} & \val{1.9}{1.9}{0.5}{4.3}{2.9} & \valb{3.2}{3.2}{0.6}{4.3}{2.9} & \val{-0.5}{-0.5}{0.2}{8.9}{3.2} & \val{1.5}{1.5}{0.2}{8.9}{3.2} & \val{4.4}{4.4}{0.5}{8.9}{3.2} & \valb{5.4}{5.4}{0.5}{8.9}{3.2} & \val{1.3}{1.3}{0.2}{4.3}{0.8} & \val{1.4}{1.4}{0.2}{4.3}{0.8} & \val{2.6}{2.6}{0.5}{4.3}{0.8} & \valbu{4.3}{4.3}{0.6}{4.3}{0.8} \\[0pt]
50B & \val{-0.3}{-0.3}{0.2}{4.3}{2.9} & \val{1.0}{1.0}{0.4}{4.3}{2.9} & \val{3.0}{3.0}{0.6}{4.3}{2.9} & \valb{3.9}{3.9}{0.8}{4.3}{2.9} & \val{1.6}{1.6}{0.2}{8.9}{3.2} & \val{3.5}{3.5}{0.4}{8.9}{3.2} & \val{6.1}{6.1}{0.5}{8.9}{3.2} & \valb{7.3}{7.3}{0.8}{8.9}{3.2} & \val{0.7}{0.7}{0.2}{4.3}{0.8} & \val{1.5}{1.5}{0.2}{4.3}{0.8} & \valb{3.4}{3.4}{0.6}{4.3}{0.8} & \val{3.2}{3.2}{0.8}{4.3}{0.8} \\[0pt]
80B & \val{0.1}{0.1}{0.2}{4.3}{2.9} & \val{2.0}{2.0}{0.4}{4.3}{2.9} & \val{3.6}{3.6}{0.6}{4.3}{2.9} & \valbu{4.3}{4.3}{0.8}{4.3}{2.9} & \val{2.6}{2.6}{0.4}{8.9}{3.2} & \val{5.3}{5.3}{0.6}{8.9}{3.2} & \valbu{7.7}{7.7}{0.8}{8.9}{3.2} & \val{7.6}{7.6}{0.6}{8.9}{3.2} & \val{0.9}{0.9}{0.4}{4.3}{0.8} & \val{1.8}{1.8}{0.2}{4.3}{0.8} & \valu{3.6}{3.6}{0.8}{4.3}{0.8} & \valb{3.9}{3.9}{1.0}{4.3}{0.8} \\[0pt]
100B & \val{0.4}{0.4}{0.2}{4.3}{2.9} & \val{2.3}{2.3}{0.5}{4.3}{2.9} & \val{3.9}{3.9}{0.8}{4.3}{2.9} & \valb{4.2}{4.2}{0.8}{4.3}{2.9} & \val{3.1}{3.1}{0.2}{8.9}{3.2} & \val{6.0}{6.0}{0.6}{8.9}{3.2} & \val{7.6}{7.6}{0.8}{8.9}{3.2} & \valbu{8.9}{8.9}{0.8}{8.9}{3.2} & \val{0.7}{0.7}{0.4}{4.3}{0.8} & \val{1.9}{1.9}{0.5}{4.3}{0.8} & \valb{3.3}{3.3}{0.6}{4.3}{0.8} & \val{3.2}{3.2}{1.0}{4.3}{0.8} \\[0pt]
300B & \valu{0.9}{0.9}{0.2}{4.3}{2.9} & \valu{3.3}{3.3}{0.6}{4.3}{2.9} & \valbu{4.3}{4.3}{0.8}{4.3}{2.9} & \val{4.0}{4.0}{0.8}{4.3}{2.9} & \valu{4.4}{4.4}{0.6}{8.9}{3.2} & \valbu{7.6}{7.6}{0.6}{8.9}{3.2} & \val{7.4}{7.4}{0.8}{8.9}{3.2} & \val{7.2}{7.2}{0.6}{8.9}{3.2} & \valu{1.5}{1.5}{0.6}{4.3}{0.8} & \valbu{4.3}{4.3}{0.8}{4.3}{0.8} & \val{3.5}{3.5}{0.5}{4.3}{0.8} & \val{2.8}{2.8}{0.8}{4.3}{0.8} \\
\bottomrule
\end{tabular*}
\caption{\textbf{Distillation improvement over standard pretraining baseline under the best loss mixing coefficient $\bm{\alpha}$.} The student is 1.7B trained on 50B tokens. Rows indicate teacher token budget; columns indicate teacher architecture size. Each cell shows percentage improvement (positive is better) with the corresponding best $\alpha$ in parentheses. Bold indicates best within row (same tokens, different architectures); underline indicates best within column (same architecture, different tokens). Even weak teachers (smaller architecture or fewer tokens than the student) can improve distillation with proper $\alpha$ selection. However, the strongest teachers do not always yield the best results: at 300B tokens, smaller teacher architectures can outperform larger ones. Complete per-benchmark results are listed in \secrefc{appendix:eval}.}
\label{tab:exp1_results}
\end{table*}

\subsection{Strong-to-Weak Distillation}
\begin{tcolorbox}[before skip=18pt, after skip=10pt,
  enhanced jigsaw,
  colback=LightGray,
  colframe=finding2border,
  drop fuzzy shadow southeast={fill=finding2shadow},
  boxrule=0.9pt,
  boxsep=0.1pt,
  left=10pt,
  right=10pt,
  top=5pt,
  bottom=4pt
]
{\small\textsf{\textbf{Finding 2:} Stronger teachers do not guarantee better distillation and can degrade transfer.}}
\end{tcolorbox}

Surprisingly, the highest downstream accuracy improvement comes from the same-size teacher with more training tokens, instead of a stronger teacher that is both larger and trained on more tokens. In strong-to-weak scenarios, contrary to expectation, we observe saturation and reversal at high teacher strength.

First, at high token budgets, larger architectures can underperform smaller ones. In \tabref{tab:exp1_results}, at 300B tokens, downstream improvement follows the trend of a smaller teacher distilling a better student:  1.7B (+4.3\%) $>$ 3.8B (+3.5\%) $>$ 8.0B (+2.8\%), with the same-architecture teacher outperforming larger ones. A similar pattern holds for out-of-distribution: 1.7B (+7.6\%) $>$ 3.8B (+7.4\%) $>$ 8.0B (+7.2\%).

Second, at fixed architecture, training longer can hurt for large teachers. For the 8.0B teacher, downstream accuracy peaks at 30B tokens (+4.3\%) and declines with further training: +3.9\% at 80B, +3.2\% at 100B, and +2.8\% at 300B. The 3.8B teacher shows a similar pattern, peaking at 80B tokens (+3.6\%) before declining. Out-of-domain perplexity follows the same trend: 8.0B teacher peaks at 100B tokens (+8.9\%) and drops to +7.2\% at 300B, and 3.8B teacher peaks at 80B.  This suggests saturation at high teacher training budgets.

In contrast, when the teacher is same-architecture or smaller, more teacher training tokens monotonically lead to better distillation. These observations demonstrate that architecture mismatch between teacher and student can limit knowledge transfer, and that overtraining large teachers reduces the usefulness of their output distributions for smaller students. Thus, compatibility matters more than absolute teacher scale.

\subsection{Performance by Domains}
\begin{tcolorbox}[before skip=18pt, after skip=10pt,
  enhanced jigsaw,
  colback=LightGray,
  colframe=finding3border,
  drop fuzzy shadow southeast={fill=finding3shadow},
  boxrule=0.9pt,
  boxsep=0.1pt,
  left=10pt,
  right=10pt,
  top=5pt,
  bottom=4pt
]
{\small\textsf{\textbf{Finding 3:} Distillation improves generalization
(out-of-distribution, downstream) more easily than
in-domain fitting.}}
\end{tcolorbox}

We observe different trends across evaluation types.
Out-of-domain perplexity and downstream accuracy are easier to improve than in-domain perplexity. In \tabref{tab:exp1_results}, several configurations that fail to improve in-domain perplexity still yield gains on the other two metrics. For example, the 0.7B teacher at 50B tokens degrades in-domain perplexity (-0.3\%) but improves out-of-distribution perplexity (+1.6\%) and downstream accuracy (+0.7\%). The magnitude of improvement also differs: out-of-distribution perplexity shows the largest gains, reaching up to +8.9\%, compared to +4.3\% for both in-domain perplexity and downstream accuracy. Notably, in-domain and out-of-distribution evaluations are both perplexity tasks, yet distillation yields substantially better improvement on the curated out-of-distribution high-quality corpora. This pattern suggests that improvement on one evaluation type does not guarantee improvement on another. More broadly, distillation can transfer useful knowledge for generalization, even when it cannot improve fit to the training distribution.

\begin{figure*}[!t]
  \centering
  \includegraphics[width=\textwidth]{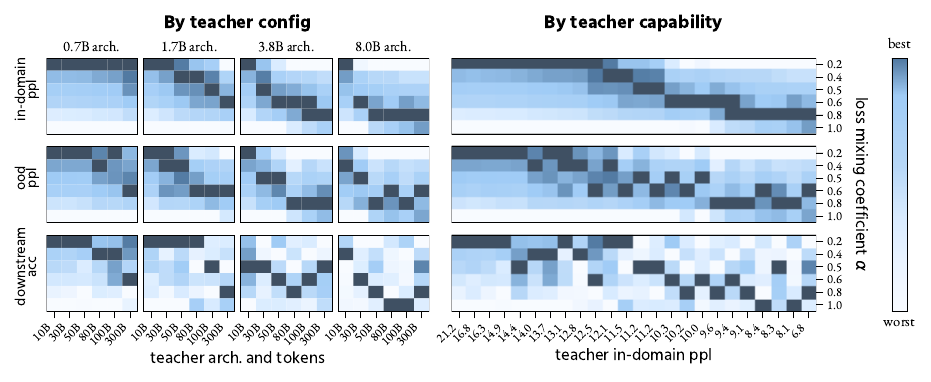}
  \vspace{-.6cm}
  \caption{\textbf{Relative performance of $\alpha$ across teacher configurations.} Column-wise, each cell shows the normalized performance within each teacher's runs, where darker colors indicate better $\alpha$ for that teacher. Left panel groups teachers by architecture and training tokens; right panel orders all teachers by in-domain perplexity. All three evaluation types follow the same pattern, with downstream accuracy being noisier: for each teacher, performance increases gradually toward an optimal $\alpha$ then decreases, and stronger teachers favor higher $\alpha$.}
  \label{fig:alpha_heatmap}
  \vspace{-.3cm}
\end{figure*}

\section{The Role of Loss Mixing}
\label{sec:alpha_role}

\subsection{Selecting the Optimal Mixing Coefficient}

\paragraph{Each teacher has an optimal $\alpha$.}
\figref{fig:alpha_heatmap} (left) visualizes the relative performance of each $\alpha$ value across teacher configurations. For each teacher-evaluation pair, we apply min-max normalization across the six $\alpha$ values on evaluation improvements, where $\text{normalized} = (\text{value} - \min) / (\max - \min)$, so the best $\alpha$ maps to 100\% (darkest) and the worst to 0\% (lightest) for within-teacher comparisons. This normalization allows us to compare $\alpha$'s optimality patterns within each teacher regardless of absolute improvement magnitudes. Across all three evaluation types, the heatmap reveals a consistent pattern: moving from $\alpha=0.2$ to $\alpha=1.0$, performance rises toward an optimal value then falls. For perplexity evaluations, the trend is smooth and consistent, while downstream accuracy shows more variance. This smoothness suggests that optimal $\alpha$ reflects a true optimum rather than noise.

\begin{figure*}[!t]
  \centering
  \includegraphics[width=\textwidth]{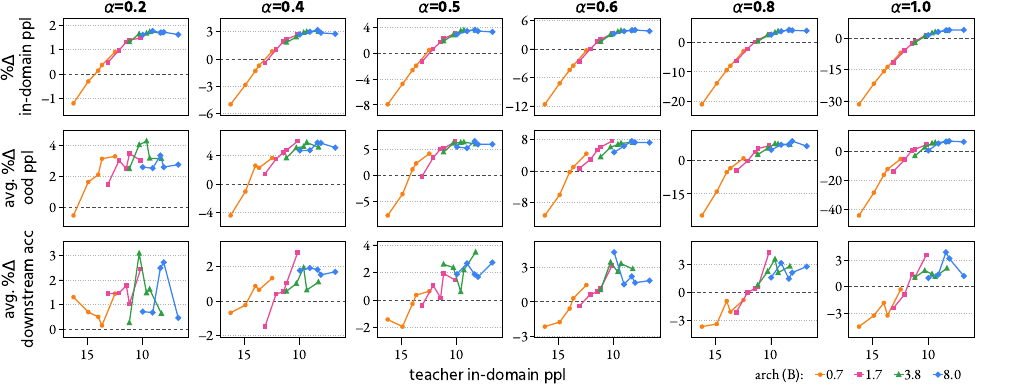}
  \vspace{-.6cm}
  \caption{\textbf{Distillation improvement versus teacher in-domain perplexity under different loss mixing coefficient $\bm{\alpha}$.} Rows show evaluation types; columns show the trend for increasing $\alpha$. Each line represents a teacher architecture, with points corresponding to different token budgets (10B-token teachers excluded for suboptimal and thus noisier performance). As $\alpha$ increases, trends become more obvious. For in-domain perplexity (top row), all architectures converge to a single trend: teacher loss alone predicts improvement regardless of architecture. For out-of-distribution perplexity (middle row), each architecture follows a separate trend, with smaller teachers outperforming larger ones even at higher loss levels. For downstream (bottom row), the upward general trend emerges but with more variance.}
  \label{fig:alpha_trend}
  \vspace{-.4cm}
\end{figure*}

\noindent
\begin{minipage}[t]{0.45\textwidth}
\vspace{0pt}
\paragraph{Optimal $\alpha$ correlates with teacher strength.}
\figref{fig:alpha_heatmap} (right) reorders teachers by their in-domain perplexity, revealing a clear trend: weaker teachers favor lower $\alpha$, while stronger teachers favor higher $\alpha$. \figref{fig:alpha_scatter} directly confirms this pattern. For weaker teachers, low-$\alpha$ points cluster at the top (better improvement), whereas for stronger teachers, high-$\alpha$ points dominate. Intuitively, weaker teachers provide less reliable output distributions, so the student benefits from retaining more ground-truth supervision via lower $\alpha$.
\end{minipage}%
\hfill
\begin{minipage}[t]{0.5\textwidth}
\vspace{-18pt}
\centering
\includegraphics[width=\textwidth]{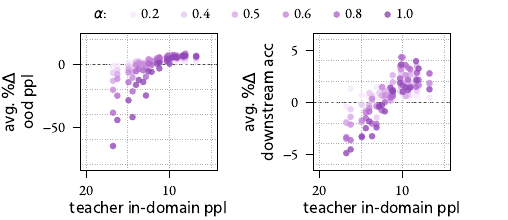}
\captionof{figure}{\textbf{Distillation improvement versus teacher capability by $\bm{\alpha}$.} Each point represents one teacher-alpha combination. For weaker teachers (higher perplexity, left), lower $\alpha$ values yield better improvement. For stronger teachers (lower perplexity, right), higher $\alpha$ values perform better.}
\label{fig:alpha_scatter}
\end{minipage}

\paragraph{Stronger teacher falls short: a loss mixing perspective.} Interestingly, for 3.8B and 8.0B teachers, we observe the optimal $\alpha$ rising then falling along the token dimension. In downstream accuracy, for example, the 8.0B teacher's optimal $\alpha$ across token budgets (10B, 30B, 50B, 80B, 100B, 300B) is 0.4, 0.6, 0.8, 1.0, 1.0, 0.8; for the 3.8B teacher, it is 0.5, 0.6, 0.6, 0.8, 0.6, 0.5. As these large teachers train longer, optimal $\alpha$ first increases, but eventually decreases at the high token budgets. This pattern does not appear for same-architecture (1.7B) or smaller (0.7B) teachers, where optimal $\alpha$ increases monotonically with training tokens. This provides additional evidence for the counterintuitive strong-to-weak observation in \secrefc{sec:results}: when large teachers become overtrained and too strong, the student needs to reduce reliance on their signal for better performance. From a practical standpoint, the decreasing optimal $\alpha$ trend signals that investing in an even stronger teacher may yield diminishing returns or, in some cases, even hurt.

\subsection{Trends when Increasing Distillation Loss Weight}
\label{sec:fixed_alpha}

Previous sections tuned $\alpha$ per teacher configuration. We now examine how distillation behavior changes as $\alpha$ increases. For each $\alpha$ value, \figref{fig:alpha_trend} shows each evaluation type's improvement against teacher's in-domain perplexity to see whether any consistent trend emerges and how it evolves.

\paragraph{Different evaluation types show different trends.}
\figref{fig:alpha_trend} shows improvement versus teacher in-domain perplexity across $\alpha$. As $\alpha$ increases, trends become more pronounced across all evaluation types. For in-domain perplexity, all architectures converge to a single trend: improvement correlates with teacher loss regardless of architecture size. This pattern strengthens as $\alpha$ grows, with the four architecture lines nearly overlapping at high $\alpha$. This shows that given a teacher's loss, the student's improvement is directly predictable.

For out-of-distribution perplexity, each architecture follows a separate trend, with lines more horizontally separated than in the in-domain case, indicating that teacher loss and out-of-distribution improvement do not share the same relationship across architectures. For downstream accuracy, a general upward trend emerges but with considerable variance. These observations demonstrate that in-domain transfer closely follows teacher loss as we focus more on distillation, but out-of-distribution generalization and downstream performance depend on factors beyond teacher loss, such as tokens and architecture match.

\paragraph{Teacher train loss alone is a misleading metric for teacher quality.} The findings suggest that using teacher train loss as the sole indicator of teacher quality is misleading. While teacher train loss reliably indicates in-domain transfer, it fails to capture out-of-distribution and downstream performance. In \figref{fig:alpha_trend} (middle and bottom row), smaller teachers with higher train loss yield better improvement than larger teachers with lower train loss. This arises from the additional tokens the smaller teacher has seen during pretraining, which provides broader coverage of the data distribution despite a higher loss for a subset. Practitioners selecting teachers based on train loss therefore miss configurations that transfer better for generalization.

\begin{table*}[!t]
\centering
\small
\setlength{\tabcolsep}{0pt}
\setlength{\catskip}{16pt}
\setlength{\abovetopsep}{0pt}
\setlength{\belowbottomsep}{0pt}
\setlength{\aboverulesep}{0pt}
\setlength{\belowrulesep}{0pt}
\begin{tabular*}{\textwidth}{@{\hspace{\catskip}} r @{\hspace{1.5\catskip}} c@{}c@{}c@{}c @{\hspace{1.3\catskip}} c@{}c@{}c@{}c @{\hspace{1.3\catskip}} c@{}c@{}c@{}c @{\hspace{\catskip}\extracolsep{\fill}}}
\toprule
\rule[-3pt]{0pt}{12pt}\rlap{\hspace{-20pt}\smash{\raisebox{-10pt}{\fontsize{8}{9}\selectfont\shortstack[l]{\hspace{15pt}teacher\\[-6pt]\hspace{5pt}\textcolor{gray}{\rotatebox{-25}{\rule{18pt}{0.4pt}}}\\[-6.5pt]\hspace{-13.5pt}student}}}}& \multicolumn{4}{c}{\hspace{-25pt}$\pdelta$~in-domain ppl ($\uparrow$)} & \multicolumn{4}{c}{\hspace{-23pt}avg. $\pdelta$~out-of-distribution ppl ($\uparrow$)} & \multicolumn{4}{c}{\hspace{-18pt}avg. $\pdelta$~downstream acc ($\uparrow$)} \\
\cmidrule(l{0pt}r{23pt}){2-5} \cmidrule(l{0pt}r{23pt}){6-9} \cmidrule(l{0pt}r{18pt}){10-13}
\rule{0pt}{9pt}& 0.7B & 1.7B & 3.8B & 8.0B & 0.7B & 1.7B & 3.8B & 8.0B & 0.7B & 1.7B & 3.8B & 8.0B \\[1pt]
\midrule
0.7B & \valu{0.1}{0.1}{0.2}{4.3}{2.9} & \valu{1.8}{1.8}{0.6}{4.3}{2.9} & \val{2.6}{2.6}{0.8}{4.3}{2.9} & \valb{2.8}{2.8}{0.8}{4.3}{2.9} & \val{0.5}{0.5}{0.2}{8.9}{3.2} & \val{3.2}{3.2}{0.6}{8.9}{3.2} & \valb{5.4}{5.4}{0.6}{8.9}{3.2} & \val{5.3}{5.3}{0.6}{8.9}{3.2} & \valu{1.3}{1.3}{0.6}{4.3}{0.8} & \valu{2.1}{2.1}{0.4}{4.3}{0.8} & \valb{2.7}{2.7}{1.0}{4.3}{0.8} & \val{1.9}{1.9}{1.0}{4.3}{0.8} \\[0pt]
1.7B & \val{-0.3}{-0.3}{0.2}{4.3}{2.9} & \val{1.0}{1.0}{0.4}{4.3}{2.9} & \valu{3.0}{3.0}{0.6}{4.3}{2.9} & \valbu{3.9}{3.9}{0.8}{4.3}{2.9} & \valu{1.6}{1.6}{0.2}{8.9}{3.2} & \valu{3.5}{3.5}{0.4}{8.9}{3.2} & \valu{6.1}{6.1}{0.5}{8.9}{3.2} & \valbu{7.3}{7.3}{0.8}{8.9}{3.2} & \val{0.7}{0.7}{0.2}{4.3}{0.8} & \val{1.5}{1.5}{0.2}{4.3}{0.8} & \valbu{3.4}{3.4}{0.6}{4.3}{0.8} & \valu{3.2}{3.2}{0.8}{4.3}{0.8} \\[0pt]
3.8B & \val{-1.8}{-1.8}{0.2}{4.3}{2.9} & \val{-0.3}{-0.3}{0.2}{4.3}{2.9} & \val{1.2}{1.2}{0.4}{4.3}{2.9} & \valb{2.8}{2.8}{0.5}{4.3}{2.9} & \val{0.3}{0.3}{0.2}{8.9}{3.2} & \val{2.7}{2.7}{0.2}{8.9}{3.2} & \val{4.7}{4.7}{0.4}{8.9}{3.2} & \valb{6.2}{6.2}{0.5}{8.9}{3.2} & \val{0.3}{0.3}{0.2}{4.3}{0.8} & \val{0.1}{0.1}{0.5}{4.3}{0.8} & \val{2.8}{2.8}{0.4}{4.3}{0.8} & \valb{3.0}{3.0}{0.5}{4.3}{0.8} \\[0pt]
8.0B & \val{-3.2}{-3.2}{0.2}{4.3}{2.9} & \val{-1.9}{-1.9}{0.2}{4.3}{2.9} & \val{-0.6}{-0.6}{0.2}{4.3}{2.9} & \valb{0.8}{0.8}{0.4}{4.3}{2.9} & \val{1.4}{1.4}{0.2}{8.9}{3.2} & \val{3.4}{3.4}{0.2}{8.9}{3.2} & \val{4.8}{4.8}{0.4}{8.9}{3.2} & \valb{7.1}{7.1}{0.5}{8.9}{3.2} & \val{-0.3}{-0.3}{0.2}{4.3}{0.8} & \val{0.6}{0.6}{0.2}{4.3}{0.8} & \val{1.8}{1.8}{0.2}{4.3}{0.8} & \valb{3.1}{3.1}{0.4}{4.3}{0.8} \\
\bottomrule
\end{tabular*}
\caption{\textbf{Distillation improvement with fixed 50B training tokens for both teacher and student.} Rows indicate student architecture; columns indicate teacher architecture. Diagonal cells represent same-level distillation; above-diagonal represent strong-to-weak; below-diagonal represent weak-to-strong. Strong-to-weak consistently improves all metrics. Weak-to-strong is effective out-of-distribution and downstream but struggles in-domain, especially with larger students.}
\vspace{-.3cm}
\label{tab:exp2_results}
\end{table*}

\section{Ablation: Generalization Across Student Sizes}
\label{sec:ablations}

\vspace{-.15cm}

We conduct additional experiments with varying student sizes. 
Our main experiments fix the student to 1.7B. We now vary both teacher and student architecture sizes while keeping training tokens fixed at 50B for both. This creates a symmetric setting where the only difference between teacher and student is the architecture size.
\tabref{tab:exp2_results} shows the results. The findings from our main experiments generalize across student sizes. Strong-to-weak distillation (above-diagonal) consistently improves all metrics across all student sizes, but the greatest improvement does not necessarily come with the maximum teacher-student size gap. Weak-to-strong distillation (below-diagonal) remains effective for out-of-distribution perplexity and downstream accuracy, but struggles for in-domain perplexity, particularly for larger students; the 8.0B student shows negative in-domain improvement with all smaller teachers. Same-level distillation (diagonal) yields positive improvement across all metrics and student sizes, confirming distillation transfers useful knowledge even without a teacher advantage. Experiments in \secrefc{app:additional_ablations} ablate distillation tokens and architecture families to further support the conclusions.

\vspace{-.15cm}

\section{Related Work}
\label{sec:related_work}
\vspace{-.15cm}

\paragraph{Knowledge distillation for language models.}
  Knowledge distillation (KD)~\citep{hintonKD} transfers knowledge from a teacher to a student by training the student to match the teacher's output distribution. The teacher's soft predictions
  contain dark knowledge to guide the student, or relative similarities between classes absent in hard labels~\citep{KDtransfer,UndImpKD}. Originally proposed for model compression in vision~\citep{ModelCompression}, KD has been widely adopted in LLMs, from masked LMs~\citep{PKD,distillbert,tinyBERT} to autoregressive LMs~\citep{MiniLM}. While recent industrial models primarily use teacher-generated synthetic
  data~\citep{llama3,qwen2.5,qwen3,deepseek-R1}, we revisit logit-level distillation in pretraining. Unlike prior work that focuses on compression with a fixed strong teacher, we systematically
  study how teacher size and compute affect distillation outcomes.

\paragraph{Factors governing distillation success.}
  Distillation outcomes depend on design choices beyond teacher strength, including loss design~\citep{DAKD,RLKD}, logits processing~\citep{AdaKD}, target generation~\citep{MiniPLM,PengKD}, and
  intermediate signal matching~\citep{FitNets,PKD}. In this work, we focus on the regime of teacher-student relations. In practice, industrial models increasingly rely on both weaker and stronger prior-generation or cross-family models as teachers~\citep{qwen2.5,qwen3,deepseek-R1}, yet systematic studies on their relations in LLM pretraining remain limited. This work fills this gap and challenges common assumptions about strong-to-weak distillation. 

\vspace{-.15cm}

\section{Conclusion}
\label{sec:conclusion}
\vspace{-.15cm}
We systematically study knowledge distillation in LLM pretraining across a range of teacher-student relationships. Our findings challenge common assumptions by showing weak-to-strong distillation can help and stronger teachers do not always yield better students. Different evaluation types also show different patterns. Together, these results challenge the common belief that distillation pretraining requires a strong teacher.

\subsection*{Acknowledgments}
This work is supported by the computational resources generously provided by Google’s TPU Research Cloud program.  
This work was also performed using Princeton Research Computing resources, a consortium led by the Princeton Institute for Computational Science and Engineering (PICSciE) and Research Computing at Princeton University. 
We also gratefully acknowledge the use of the Neuronic GPU computing cluster maintained by the Department of Computer Science at Princeton University. 
We thank Sachin Konan, Yida Yin, and Boya Zeng for helpful discussions.

\bibliographystyle{plainnat}
\bibliography{main}

\newpage
\section*{\Large Appendix}
\vspace{.2cm}
\appendix

\section{Detailed Related Works}
\label{appendix:related_works}

\paragraph{Knowledge distillation.}
Knowledge distillation (KD) \citep{hintonKD} is a widely used technique to transfer knowledge from a teacher model to a student model by training the student to match the teacher's predictive distribution. The key insight is that the teacher's soft predictions contain ``dark knowledge,'' information about the relative similarities between classes that is not present in hard labels~\citep{KDtransfer,UndImpKD}. Originally proposed for model compression~\citep{ModelCompression} in vision, KD has since been extended in many directions, including distilling intermediate representations \citep{FitNets}, attention maps \citep{Att2Att}, and relational information between samples \citep{RDK}. Various works have also studied the theoretical foundations of why distillation works, attributing its success to label smoothing effects \citep{WhenSmoothing,KDSmoothing}, regularization \citep{KDRegularizatoin,ClassRegKD}, and improved optimization landscapes \citep{KD_Efficacy,UndKD,KDTeacherPatient}. In this work, we focus on logit-level distillation for language models, where the student directly matches the teacher's output distribution.

\paragraph{Distillation in language models.}
 While distillation was introduced in vision-based models, it already proves effective in large language models (LLMs).
 In language modeling, logit-level distillation is widely used to compress pretrained Transformers, from masked LMs \citep{distillbert,tinyBERT}, to autoregressive LMs where distillation can be sensitive to the teacher and token-level behaviors \citep{gen_vs_fidelity}. Beyond matching output logits, prior work also distills intermediate signals to improve transfer \citep{PKD,MiniLM}. Distillation can further be realized through teacher-generated synthetic data, where the student is trained on sequences produced by the teacher \citep{seq2seq,Self-Instruct,LaMiniLM}. While logit-level distillation is a common technique, most recent LLM distillation proceeds via teacher-generated synthetic data \citep{llama3,qwen2.5,qwen3,deepseek-R1}; in this work, we revisit logit-level distillation in pretraining and study when it improves performance.

\paragraph{Design choices in LLM distillation.}
Recent LLM distillation studies suggest that distillation success is governed by a small set of controllable factors rather than teacher strength alone. In pretraining, choices such as logits processing \citep{AdaKD}, loss design \citep{DAKD,RLKD}, and generating teacher targets offline versus online \citep{MiniPLM,PengKD,anil2020largescaledistributedneural} can qualitatively change distillation outcomes. Moreover, KD is not monotonic in teacher strength: for autoregressive LMs, stronger teachers can sometimes degrade student performance \citep{RevisitKD,KDscalinglaw}.
Prior work has shown that stronger teachers can sometimes degrade performance, but has not systematically characterized when this occurs in large-scale pretraining, where multiple components vary across a wide range. In this work, we focus on design choices related to the teacher side and conduct large-scale experiments varying all key components, such as architecture, tokens, and loss selection.

\paragraph{Teacher-student relationships in distillation.}
While knowledge distillation has long been viewed as a compression technique \citep{ModelCompression,hintonKD}, a growing body of work shows that this strong-to-weak framing is not necessary. In vision, Born-Again Networks \citep{BornAgain} first showed that same-architecture distillation can still improve the student, with subsequent work clarifying why this effect holds \citep{CNN_selfKD,KD_Efficacy,KDRegularizatoin}.
 In LLMs, while distillation remains a standard compression approach, strict strong-to-weak supervision may be unrealistic in the long run \citep{w2s_generalization}. Consistent with this trend, industrial models could rely on prior-generation or cross-family models to supervise newer ones, exemplified in Qwen models \citep{qwen2.5,qwen3} and DeepSeek-R1 \citep{deepseek-R1}. Despite these developments, systematic studies of how the scale of different components governs logit-level distillation in LLM pretraining remain limited. \citet{KDscalinglaw} studies distillation scaling laws, but constrains teachers to a fixed budget distribution, while \citet{PengKD} studies how student size affects outcome but assumes access to a stronger pretrained teacher. To address these gaps, in this work we vary token budget and model size to study across diverse settings that result in all strong-to-weak, same-architecture, and weak-to-strong regimes.

\section{Architecture Details}
\label{appendix:archs_details}

We use four model sizes based on the Llama3 architecture family. All models share the same architectural design choices: Grouped Query Attention (GQA) with a 4:1 ratio (32 query heads, 8 KV heads), SwiGLU activation (silu + linear gating), RMSNorm with $\epsilon = 10^{-5}$, RoPE positional encoding with max timescale 500,000, vocabulary size of 128,256, and no embedding tying. \tabref{tab:arch_configs} details the configuration for each model size.

\begin{table*}[h]
\centering
\small
\begin{minipage}[t]{0.24\textwidth}
\centering
\textbf{0.7B}\\[3pt]
\begin{tabular}{@{\hspace{6pt}}l@{\hspace{8pt}}r@{\hspace{6pt}}}
\toprule
Parameter & Value \\
\midrule
Hidden dim & 1024 \\
Num layers & 12 \\
MLP dim & 8192 \\
Query heads & 32 \\
KV heads & 8 \\
Head dim & 128 \\
Vocab size & 128,256 \\
RoPE base & 500,000 \\
Norm $\epsilon$ & $10^{-5}$ \\
\bottomrule
\end{tabular}
\end{minipage}%
\hfill
\begin{minipage}[t]{0.24\textwidth}
\centering
\textbf{1.7B}\\[3pt]
\begin{tabular}{@{\hspace{6pt}}l@{\hspace{8pt}}r@{\hspace{6pt}}}
\toprule
Parameter & Value \\
\midrule
Hidden dim & 2048 \\
Num layers & 16 \\
MLP dim & 8192 \\
Query heads & 32 \\
KV heads & 8 \\
Head dim & 128 \\
Vocab size & 128,256 \\
RoPE base & 500,000 \\
Norm $\epsilon$ & $10^{-5}$ \\
\bottomrule
\end{tabular}
\end{minipage}%
\hfill
\begin{minipage}[t]{0.24\textwidth}
\centering
\textbf{3.8B}\\[3pt]
\begin{tabular}{@{\hspace{6pt}}l@{\hspace{8pt}}r@{\hspace{6pt}}}
\toprule
Parameter & Value \\
\midrule
Hidden dim & 3072 \\
Num layers & 28 \\
MLP dim & 8192 \\
Query heads & 32 \\
KV heads & 8 \\
Head dim & 128 \\
Vocab size & 128,256 \\
RoPE base & 500,000 \\
Norm $\epsilon$ & $10^{-5}$ \\
\bottomrule
\end{tabular}
\end{minipage}%
\hfill
\begin{minipage}[t]{0.24\textwidth}
\centering
\textbf{8.0B}\\[3pt]
\begin{tabular}{@{\hspace{6pt}}l@{\hspace{8pt}}r@{\hspace{6pt}}}
\toprule
Parameter & Value \\
\midrule
Hidden dim & 4096 \\
Num layers & 32 \\
MLP dim & 14336 \\
Query heads & 32 \\
KV heads & 8 \\
Head dim & 128 \\
Vocab size & 128,256 \\
RoPE base & 500,000 \\
Norm $\epsilon$ & $10^{-5}$ \\
\bottomrule
\end{tabular}
\end{minipage}
\caption{\textbf{Architecture configurations for each model size.} All models share the same design choices (GQA, SwiGLU, RMSNorm, RoPE) but differ in hidden dimension, number of layers, and MLP dimension.}
\label{tab:arch_configs}
\end{table*}

\section{Evaluation Implementations}
\label{appendix:eval_imp}

We evaluate models using three categories of metrics: in-domain perplexity, out-of-distribution perplexity, and downstream accuracy. This section details the implementation of each.

\subsection{In-Domain Perplexity}
\label{appendix:eval_indomain_ppl}

In-domain perplexity measures how well the model predicts text from the same distribution as the training data. We compute perplexity on a held-out test split of FineWeb-Edu~\citep{fineweb} containing 1 billion tokens that are not seen during any model training (neither teacher pretraining nor student baseline or distillation).

Perplexity is defined as the exponentiated average cross-entropy loss (see Equation~\ref{eq:ppl}), where $N$ is the total number of tokens and $P(x_i \mid x_{<i})$ is the model's predicted probability of token $x_i$ given the preceding context. To compute perplexity, we tokenize the test set into a single long sequence, randomly split it into fixed-length chunks matching the model's context length (8192 tokens), run a forward pass to obtain logits, compute the cross-entropy loss between predicted logits and target tokens (shifted by one position), and average the loss across all tokens before exponentiating.
\begin{equation}
\text{PPL} = \exp\left(-\frac{1}{N}\sum_{i=1}^{N}\log P(x_i \mid x_{<i})\right)
\label{eq:ppl}
\end{equation}

\subsection{Out-of-Domain Perplexity}
\label{appendix:eval_ood_ppl}

Out-of-domain perplexity evaluates generalization to distributions beyond the training corpus. We measure perplexity on 11 curated corpora spanning diverse domains:

\begin{itemize}[leftmargin=*, topsep=2pt, itemsep=1pt]
    \item \textbf{Web and general text:} Wikitext-103~\citep{merity2017pointer} (Wikipedia articles), C4~\citep{C4} (Common Crawl web text)
    \item \textbf{Mathematics:} GSM8K~\citep{gsm8k} (grade school math problems), DM Mathematics~\citep{dm_math} (mathematical text)
    \item \textbf{Code:} HumanEval~\citep{humaneval} (code generation problems), CodeSearchNet~\citep{codesearchnet} (code with documentation)
    \item \textbf{Scientific:} arXiv~\citep{arXivSummerization} (scientific paper abstracts), PubMedQA~\citep{PubMedQA} (biomedical question-answer pairs)
    \item \textbf{News:} CNN-DailyMail~\citep{CNNDailyMail} (news articles)
    \item \textbf{Legal:} ECHR~\citep{ECHR} (European Court of Human Rights cases)
    \item \textbf{Multilingual:} XQuAD~\citep{XQuAD} (multilingual question-answering across 11 languages)
\end{itemize}

Each corpus is formatted according to its structure. For plain text corpora (Wikitext, arXiv), documents are joined with double newlines. For QA datasets (GSM8K, PubMedQA), question-answer pairs are concatenated. For code (HumanEval), prompts and canonical solutions are combined. Perplexity is computed using the same procedure as in-domain evaluation.

\subsection{Downstream Accuracy}
\label{appendix:eval_acc}

Downstream accuracy evaluates transfer to tasks requiring capabilities beyond next-token prediction. We use 15 multiple-choice benchmarks covering broad knowledge, reasoning, and comprehension.

\paragraph{Scoring mechanism.}
For multiple-choice tasks, the model does not generate text. Instead, we use log-likelihood scoring: for each question, we construct prompt-answer pairs for each possible choice, compute the log-likelihood $\log P(\text{answer} \mid \text{question})$ for each choice, and select the answer with the highest log-likelihood as the prediction.

For a context-continuation pair, we tokenize both, concatenate them, run a forward pass to obtain logits, apply log-softmax, and sum the log probabilities of the continuation tokens (see Equation~\ref{eq:acc_continuation}). Some benchmarks use length-normalized accuracy to prevent bias toward shorter answers. In this case, the log-likelihood is divided by the number of characters in the continuation before comparison.
\begin{equation}
\log P(\text{continuation} \mid \text{context}) = \sum_{j=1}^{|\text{cont}|} \log P(t_j \mid \text{context}, t_{<j})
\label{eq:acc_continuation}
\end{equation}

\paragraph{Benchmarks.}
We evaluate on 15 benchmarks with the following configurations:

\begin{itemize}[leftmargin=*, topsep=2pt, itemsep=1pt]
    \item \textbf{Broad knowledge:} MMLU~\citep{mmlu} (57 academic subjects, 5-shot, 4 choices)
    \item \textbf{Science and math:} ARC-Easy~\citep{ARC} (elementary science, 0-shot, 3-5 choices), SciQ~\citep{SciQ} (science QA, 0-shot, 4 choices), OpenBookQA~\citep{OpenBookQA} (science facts, 5-shot, 4 choices), MathQA~\citep{MathQA} (math word problems, 5-shot, 5 choices)
    \item \textbf{Commonsense reasoning:} HellaSwag~\citep{HellaSwag} (sentence completion, 0-shot, 4 choices), PIQA~\citep{PIQA} (physical intuition, 0-shot, 2 choices), WinoGrande~\citep{WinoGrande} (coreference, 5-shot, 2 choices), Social IQa~\citep{SocialIQA} (social commonsense, 0-shot, 3 choices), CommonsenseQA~\citep{CommonsenseQA} (general commonsense, 10-shot, 5 choices)
    \item \textbf{Reading comprehension:} RACE~\citep{RACE} (English exams, 0-shot, 4 choices)
    \item \textbf{Logical reasoning:} LogiQA 2.0~\citep{LogiQA} (logical reasoning, 10-shot, 4 choices)
    \item \textbf{Truthfulness:} TruthfulQA MC1~\citep{TruthfulQA} (identifying truthful answers among misconceptions, 0-shot with 10 in-prompt examples, variable choices)
    \item \textbf{Natural language inference:} ANLI R1~\citep{ANLI} (adversarial entailment, 10-shot, 3 choices)
    \item \textbf{Medical:} MedMCQA~\citep{MedMCQA} (medical entrance exams, 10-shot, 4 choices)

\paragraph{Length normalization and few-shot selection.} Four benchmarks use length-normalized accuracy to prevent bias toward shorter answers: HellaSwag, ARC-Easy, PIQA, and OpenBookQA. These benchmarks score the full text of answer choices, which vary significantly in length. The remaining benchmarks use raw accuracy, as their answer choices are either single letters or similar-length phrases. For few-shot evaluation, we use deterministic selection: examples are drawn from the first $N$ samples of the training split (or dev split for MMLU). This ensures reproducibility across runs.
\end{itemize}

\section{Pure Distillation}
\label{appendix:pure_KD}

\begin{table*}[!h]
\centering
\small
\setlength{\valboxwidth}{30pt}
\setlength{\tabcolsep}{0pt}
\setlength{\catskip}{16pt}
\setlength{\abovetopsep}{0pt}
\setlength{\belowbottomsep}{0pt}
\setlength{\aboverulesep}{0pt}
\setlength{\belowrulesep}{0pt}
\begin{tabular*}{\textwidth}{@{\hspace{\catskip}} r @{\hspace{1.1\catskip}} c@{}c@{}c@{}c @{\hspace{0.8\catskip}} c@{}c@{}c@{}c @{\hspace{0.8\catskip}} c@{}c@{}c@{}c @{\hspace{\catskip}\extracolsep{\fill}}}
\toprule
\rule[-3pt]{0pt}{12pt}& \multicolumn{4}{c}{\hspace{-25pt}$\pdelta$~in-domain ppl ($\uparrow$)} & \multicolumn{4}{c}{\hspace{-23pt}avg. $\pdelta$~out-of-distribution ppl ($\uparrow$)} & \multicolumn{4}{c}{\hspace{-18pt}avg. $\pdelta$~downstream acc ($\uparrow$)} \\
\cmidrule(l{0pt}r{23pt}){2-5} \cmidrule(l{0pt}r{23pt}){6-9} \cmidrule(l{0pt}r{18pt}){10-13}
\rule{0pt}{9pt}& 0.7B & 1.7B & 3.8B & 8.0B & 0.7B & 1.7B & 3.8B & 8.0B & 0.7B & 1.7B & 3.8B & 8.0B \\[1pt]
\midrule
10B & \valna{-68.7}{-68.7}{3.9}{68.7} & \valna{-36.1}{-36.1}{3.9}{68.7} & \valna{-20.6}{-20.6}{3.9}{68.7} & \valnab{-11.2}{-11.2}{3.9}{68.7} & \valna{-146.5}{-146.5}{7.3}{146.5} & \valna{-65.1}{-65.1}{7.3}{146.5} & \valna{-42.0}{-42.0}{7.3}{146.5} & \valnab{-24.6}{-24.6}{7.3}{146.5} & \valna{-8.6}{-8.6}{3.9}{8.6} & \valna{-4.9}{-4.9}{3.9}{8.6} & \valna{-2.5}{-2.5}{3.9}{8.6} & \valnab{-0.4}{-0.4}{3.9}{8.6} \\
30B & \valna{-31.6}{-31.6}{3.9}{68.7} & \valna{-11.5}{-11.5}{3.9}{68.7} & \valna{-2.2}{-2.2}{3.9}{68.7} & \valnab{1.4}{1.4}{3.9}{68.7} & \valna{-44.3}{-44.3}{7.3}{146.5} & \valna{-13.8}{-13.8}{7.3}{146.5} & \valna{-3.0}{-3.0}{7.3}{146.5} & \valnab{0.8}{0.8}{7.3}{146.5} & \valna{-4.5}{-4.5}{3.9}{8.6} & \valna{-2.4}{-2.4}{3.9}{8.6} & \valnab{+1.1}{1.1}{3.9}{8.6} & \valna{+1.0}{1.0}{3.9}{8.6} \\
50B & \valna{-21.8}{-21.8}{3.9}{68.7} & \valna{-5.9}{-5.9}{3.9}{68.7} & \valna{0.9}{0.9}{3.9}{68.7} & \valnab{3.0}{3.0}{3.9}{68.7} & \valna{-28.5}{-28.5}{7.3}{146.5} & \valna{-5.6}{-5.6}{7.3}{146.5} & \valna{2.9}{2.9}{7.3}{146.5} & \valnab{5.7}{5.7}{7.3}{146.5} & \valna{-3.3}{-3.3}{3.9}{8.6} & \valna{-0.9}{-0.9}{3.9}{8.6} & \valnab{+1.9}{1.9}{3.9}{8.6} & \valna{+1.4}{1.4}{3.9}{8.6} \\
80B & \valna{-15.8}{-15.8}{3.9}{68.7} & \valna{-2.6}{-2.6}{3.9}{68.7} & \valna{2.4}{2.4}{3.9}{68.7} & \valnab{3.7}{3.7}{3.9}{68.7} & \valna{-16.1}{-16.1}{7.3}{146.5} & \valna{0.9}{0.9}{7.3}{146.5} & \valna{5.8}{5.8}{7.3}{146.5} & \valnab{7.0}{7.0}{7.3}{146.5} & \valna{-1.8}{-1.8}{3.9}{8.6} & \valna{+1.4}{1.4}{3.9}{8.6} & \valna{+1.4}{1.4}{3.9}{8.6} & \valnabu{+3.9}{3.9}{3.9}{8.6} \\
100B & \valna{-13.8}{-13.8}{3.9}{68.7} & \valna{-1.5}{-1.5}{3.9}{68.7} & \valna{2.8}{2.8}{3.9}{68.7} & \valnab{3.8}{3.8}{3.9}{68.7} & \valna{-12.2}{-12.2}{7.3}{146.5} & \valna{1.6}{1.6}{7.3}{146.5} & \valna{6.3}{6.3}{7.3}{146.5} & \valnabu{7.3}{7.3}{7.3}{146.5} & \valna{-3.3}{-3.3}{3.9}{8.6} & \valna{+1.1}{1.1}{3.9}{8.6} & \valna{+1.2}{1.2}{3.9}{8.6} & \valnab{+3.2}{3.2}{3.9}{8.6} \\
300B & \valnau{-7.0}{-7.0}{3.9}{68.7} & \valnau{1.6}{1.6}{3.9}{68.7} & \valnau{3.6}{3.6}{3.9}{68.7} & \valnabu{3.9}{3.9}{3.9}{68.7} & \valnau{-5.2}{-5.2}{7.3}{146.5} & \valnau{4.9}{4.9}{7.3}{146.5} & \valnabu{6.9}{6.9}{7.3}{146.5} & \valna{6.8}{6.8}{7.3}{146.5} & \valnau{-0.3}{-0.3}{3.9}{8.6} & \valnabu{+3.6}{3.6}{3.9}{8.6} & \valnau{+2.1}{2.1}{3.9}{8.6} & \valna{+1.2}{1.2}{3.9}{8.6} \\
\bottomrule
\end{tabular*}
\caption{\textbf{Distillation improvement under pure distillation ($\bm{\alpha=1.0}$).} Rows indicate teacher token budget; columns indicate teacher architecture size. Pure distillation works for strong teachers (large architecture, high token budget) but degrades significantly for weak teachers, particularly smaller architectures with fewer training tokens.}
\label{tab:pure_kd}
\end{table*}

\begin{table*}[!h]
\centering
\small
\setlength{\valboxwidth}{30pt}
\setlength{\tabcolsep}{0pt}
\setlength{\catskip}{16pt}
\setlength{\abovetopsep}{0pt}
\setlength{\belowbottomsep}{0pt}
\setlength{\aboverulesep}{0pt}
\setlength{\belowrulesep}{0pt}
\begin{tabular*}{\textwidth}{@{\hspace{\catskip}} r @{\hspace{1.1\catskip}} c@{}c@{}c@{}c @{\hspace{0.8\catskip}} c@{}c@{}c@{}c @{\hspace{0.8\catskip}} c@{}c@{}c@{}c @{\hspace{\catskip}\extracolsep{\fill}}}
\toprule
\rule[-3pt]{0pt}{12pt}& \multicolumn{4}{c}{\hspace{-25pt}$\pdelta$~in-domain ppl ($\uparrow$)} & \multicolumn{4}{c}{\hspace{-23pt}avg. $\pdelta$~out-of-distribution ppl ($\uparrow$)} & \multicolumn{4}{c}{\hspace{-18pt}avg. $\pdelta$~downstream acc ($\uparrow$)} \\
\cmidrule(l{0pt}r{23pt}){2-5} \cmidrule(l{0pt}r{23pt}){6-9} \cmidrule(l{0pt}r{18pt}){10-13}
\rule{0pt}{9pt}& 0.7B & 1.7B & 3.8B & 8.0B & 0.7B & 1.7B & 3.8B & 8.0B & 0.7B & 1.7B & 3.8B & 8.0B \\[1pt]
\midrule
10B & \valna{-65.8}{-65.8}{0.1}{65.8} & \valna{-34.7}{-34.7}{0.1}{65.8} & \valna{-20.2}{-20.2}{0.1}{65.8} & \valnab{-11.5}{-11.5}{0.1}{65.8} & \valna{-143.3}{-143.3}{0.1}{143.3} & \valna{-64.0}{-64.0}{0.1}{143.3} & \valna{-43.2}{-43.2}{0.1}{143.3} & \valnab{-26.2}{-26.2}{0.1}{143.3} & \valna{-8.1}{-8.1}{0.1}{8.1} & \valna{-4.5}{-4.5}{0.1}{8.1} & \valna{-1.7}{-1.7}{0.1}{8.1} & \valnab{-1.3}{-1.3}{0.1}{8.1} \\
30B & \valna{-30.4}{-30.4}{0.1}{65.8} & \valna{-12.0}{-12.0}{0.1}{65.8} & \valna{-4.1}{-4.1}{0.1}{65.8} & \valnab{-1.8}{-1.8}{0.1}{65.8} & \valna{-43.8}{-43.8}{0.1}{143.3} & \valna{-15.3}{-15.3}{0.1}{143.3} & \valna{-7.4}{-7.4}{0.1}{143.3} & \valnab{-4.6}{-4.6}{0.1}{143.3} & \valna{-5.8}{-5.8}{0.1}{8.1} & \valna{-3.8}{-3.8}{0.1}{8.1} & \valnab{-1.5}{-1.5}{0.1}{8.1} & \valna{-3.3}{-3.3}{0.1}{8.1} \\
50B & \valna{-21.5}{-21.5}{0.1}{65.8} & \valna{-6.9}{-6.9}{0.1}{65.8} & \valna{-2.1}{-2.1}{0.1}{65.8} & \valnab{-0.9}{-0.9}{0.1}{65.8} & \valna{-30.1}{-30.1}{0.1}{143.3} & \valna{-9.1}{-9.1}{0.1}{143.3} & \valna{-3.2}{-3.2}{0.1}{143.3} & \valnab{-1.6}{-1.6}{0.1}{143.3} & \valna{-4.0}{-4.0}{0.1}{8.1} & \valna{-2.4}{-2.4}{0.1}{8.1} & \valnab{-1.5}{-1.5}{0.1}{8.1} & \valna{-1.8}{-1.8}{0.1}{8.1} \\
80B & \valna{-15.9}{-15.9}{0.1}{65.8} & \valna{-4.6}{-4.6}{0.1}{65.8} & \valna{-1.2}{-1.2}{0.1}{65.8} & \valnab{-0.6}{-0.6}{0.1}{65.8} & \valna{-18.7}{-18.7}{0.1}{143.3} & \valna{-4.4}{-4.4}{0.1}{143.3} & \valna{-1.9}{-1.9}{0.1}{143.3} & \valnab{-0.6}{-0.6}{0.1}{143.3} & \valna{-2.7}{-2.7}{0.1}{8.1} & \valnau{-0.4}{-0.4}{0.1}{8.1} & \valna{-2.2}{-2.2}{0.1}{8.1} & \valnabu{0.0}{0.0}{0.1}{8.1} \\
100B & \valna{-14.2}{-14.2}{0.1}{65.8} & \valna{-3.8}{-3.8}{0.1}{65.8} & \valna{-1.1}{-1.1}{0.1}{65.8} & \valnab{-0.4}{-0.4}{0.1}{65.8} & \valna{-15.3}{-15.3}{0.1}{143.3} & \valna{-4.4}{-4.4}{0.1}{143.3} & \valnab{-1.3}{-1.3}{0.1}{143.3} & \valna{-1.6}{-1.6}{0.1}{143.3} & \valna{-4.0}{-4.0}{0.1}{8.1} & \valna{-0.8}{-0.8}{0.1}{8.1} & \valna{-2.1}{-2.1}{0.1}{8.1} & \valnab{0.0}{0.0}{0.1}{8.1} \\
300B & \valnau{-7.9}{-7.9}{0.1}{65.8} & \valnau{-1.7}{-1.7}{0.1}{65.8} & \valnau{-0.7}{-0.7}{0.1}{65.8} & \valnabu{-0.1}{-0.1}{0.1}{65.8} & \valnau{-9.6}{-9.6}{0.1}{143.3} & \valnau{-2.7}{-2.7}{0.1}{143.3} & \valna{-0.5}{-0.5}{0.1}{143.3} & \valnabu{-0.4}{-0.4}{0.1}{143.3} & \valnau{-1.8}{-1.8}{0.1}{8.1} & \valnab{-0.7}{-0.7}{0.1}{8.1} & \valnau{-1.4}{-1.4}{0.1}{8.1} & \valna{-1.6}{-1.6}{0.1}{8.1} \\
\bottomrule
\end{tabular*}
\caption{\textbf{Difference of effectiveness between pure distillation and optimal $\bm{\alpha}$ distillation.} Negative values indicate pure distillation underperforms optimal $\alpha$ selection. The gap is largest for weak teachers, where pure distillation can be over 60\% worse than optimal $\alpha$ on in-domain perplexity. For strong teachers (8.0B at high token budgets), the gap narrows, but pure distillation is still generally worse, indicating that pure distillation approaches optimal performance when the teacher is sufficiently strong.}
\label{tab:pure_kd_diff}
\end{table*}

Pure distillation sets the loss mixing coefficient $\alpha=1.0$, using only the knowledge distillation loss without the language modeling loss. This is a common practice in many implementations, as it simplifies training by removing the need to tune $\alpha$.

\tabref{tab:pure_kd} shows distillation improvement under pure distillation. The results reveal a clear divide based on teacher strength. For strong teachers (large architecture with high token budgets), pure distillation yields substantial improvement across all evaluation types. For example, the 8.0B teacher at 80B tokens achieves +3.7\% in-domain perplexity, +7.0\% out-of-distribution perplexity, and +3.9\% downstream accuracy. However, for weak teachers, pure distillation degrades performance significantly. The 0.7B teacher at 10B tokens yields -68.7\% in-domain perplexity and -146.5\% out-of-distribution perplexity (number can exceed 100\% because ${ppl} \in [1,\infty)$), indicating that relying entirely on a weak teacher's signal without ground-truth supervision is harmful. Additionally, we also observe the same trend from our mixed loss experiments: stronger teachers do not always yield better students. For downstream accuracy, the 8.0B teacher peaks at 80B tokens (+3.9\%) and declines with further training (+3.2\% at 100B, +1.2\% at 300B).

\tabref{tab:pure_kd_diff} compares pure distillation to optimal $\alpha$ selection. In almost all configurations, there exists a better loss mixing than pure distillation. The gap is substantial for weak teachers: pure distillation underperforms optimal $\alpha$ by over 60\% on in-domain perplexity for small teachers with few training tokens. Even for strong teachers, pure distillation rarely matches optimal $\alpha$. Only a few configurations with the strongest teachers (8.0B at high token budgets) show near-zero difference, where pure distillation happens to be optimal. These findings suggest that tuning $\alpha$ is worthwhile across most teacher configurations.

\section{Choosing Best $\alpha$ Conditioning on All Evaluation Types}
\label{appendix:best_alpha_all}

\begin{table*}[!h]
\centering
\small
\setlength{\tabcolsep}{0pt}
\setlength{\catskip}{16pt}
\setlength{\abovetopsep}{0pt}
\setlength{\belowbottomsep}{0pt}
\setlength{\aboverulesep}{0pt}
\setlength{\belowrulesep}{0pt}
\begin{tabular*}{\textwidth}{@{\hspace{\catskip}} r @{\hspace{1.5\catskip}} c@{}c@{}c@{}c @{\hspace{1.3\catskip}} c@{}c@{}c@{}c @{\hspace{1.3\catskip}} c@{}c@{}c@{}c @{\hspace{\catskip}\extracolsep{\fill}}}
\toprule
\rule[-3pt]{0pt}{12pt}& \multicolumn{4}{c}{\hspace{-25pt}$\pdelta$~in-domain ppl ($\uparrow$)} & \multicolumn{4}{c}{\hspace{-23pt}avg. $\pdelta$~out-of-distribution ppl ($\uparrow$)} & \multicolumn{4}{c}{\hspace{-18pt}avg. $\pdelta$~downstream acc ($\uparrow$)} \\
\cmidrule(l{0pt}r{23pt}){2-5} \cmidrule(l{0pt}r{23pt}){6-9} \cmidrule(l{0pt}r{18pt}){10-13}
\rule{0pt}{9pt}& 0.7B & 1.7B & 3.8B & 8.0B & 0.7B & 1.7B & 3.8B & 8.0B & 0.7B & 1.7B & 3.8B & 8.0B \\[1pt]
\midrule
10B & \val{-2.9}{-2.9}{0.2}{4.3}{2.9} & \val{-1.4}{-1.4}{0.2}{4.3}{2.9} & \valb{-0.4}{-0.4}{0.2}{4.3}{2.9} & \val{-0.5}{-0.5}{0.4}{4.3}{2.9} & \val{-3.2}{-3.2}{0.2}{7.7}{3.2} & \val{-1.1}{-1.1}{0.2}{7.7}{3.2} & \val{1.2}{1.2}{0.2}{7.7}{3.2} & \valb{1.6}{1.6}{0.4}{7.7}{3.2} & \val{-0.5}{-0.5}{0.2}{4.3}{1.0} & \val{-0.4}{-0.4}{0.2}{4.3}{1.0} & \val{-1.0}{-1.0}{0.2}{4.3}{1.0} & \valb{0.9}{0.9}{0.4}{4.3}{1.0} \\[0pt]
30B & \val{-1.2}{-1.2}{0.2}{4.3}{2.9} & \val{0.5}{0.5}{0.2}{4.3}{2.9} & \val{1.9}{1.9}{0.5}{4.3}{2.9} & \valb{3.2}{3.2}{0.6}{4.3}{2.9} & \val{-0.5}{-0.5}{0.2}{7.7}{3.2} & \val{1.5}{1.5}{0.2}{7.7}{3.2} & \val{4.4}{4.4}{0.5}{7.7}{3.2} & \valb{4.7}{4.7}{0.6}{7.7}{3.2} & \val{1.3}{1.3}{0.2}{4.3}{1.0} & \val{1.4}{1.4}{0.2}{4.3}{1.0} & \val{2.6}{2.6}{0.5}{4.3}{1.0} & \valbu{4.3}{4.3}{0.6}{4.3}{1.0} \\[0pt]
50B & \val{-0.3}{-0.3}{0.2}{4.3}{2.9} & \val{1.0}{1.0}{0.2}{4.3}{2.9} & \val{3.0}{3.0}{0.6}{4.3}{2.9} & \valb{3.9}{3.9}{0.8}{4.3}{2.9} & \val{1.6}{1.6}{0.2}{7.7}{3.2} & \val{3.0}{3.0}{0.2}{7.7}{3.2} & \val{6.0}{6.0}{0.6}{7.7}{3.2} & \valb{7.3}{7.3}{0.8}{7.7}{3.2} & \val{0.7}{0.7}{0.2}{4.3}{1.0} & \val{1.5}{1.5}{0.2}{4.3}{1.0} & \valb{3.4}{3.4}{0.6}{4.3}{1.0} & \val{3.2}{3.2}{0.8}{4.3}{1.0} \\[0pt]
80B & \val{-1.3}{-1.3}{0.4}{4.3}{2.9} & \val{1.6}{1.6}{0.6}{4.3}{2.9} & \val{3.5}{3.5}{0.8}{4.3}{2.9} & \valb{3.7}{3.7}{1.0}{4.3}{2.9} & \val{2.6}{2.6}{0.4}{7.7}{3.2} & \val{5.3}{5.3}{0.6}{7.7}{3.2} & \valbu{7.7}{7.7}{0.8}{7.7}{3.2} & \val{7.0}{7.0}{1.0}{7.7}{3.2} & \val{0.9}{0.9}{0.4}{4.3}{1.0} & \val{0.9}{0.9}{0.6}{4.3}{1.0} & \valu{3.6}{3.6}{0.8}{4.3}{1.0} & \valb{3.9}{3.9}{1.0}{4.3}{1.0} \\[0pt]
100B & \valu{0.4}{0.4}{0.2}{4.3}{2.9} & \val{2.3}{2.3}{0.5}{4.3}{2.9} & \valb{3.8}{3.8}{0.6}{4.3}{2.9} & \val{3.8}{3.8}{1.0}{4.3}{2.9} & \val{3.1}{3.1}{0.2}{7.7}{3.2} & \val{5.1}{5.1}{0.5}{7.7}{3.2} & \val{7.0}{7.0}{0.6}{7.7}{3.2} & \valbu{7.3}{7.3}{1.0}{7.7}{3.2} & \val{0.1}{0.1}{0.2}{4.3}{1.0} & \val{1.9}{1.9}{0.5}{4.3}{1.0} & \valb{3.3}{3.3}{0.6}{4.3}{1.0} & \val{3.2}{3.2}{1.0}{4.3}{1.0} \\[0pt]
300B & \val{-0.3}{-0.3}{0.6}{4.3}{2.9} & \valu{3.3}{3.3}{0.6}{4.3}{2.9} & \valbu{4.3}{4.3}{0.8}{4.3}{2.9} & \valu{4.0}{4.0}{0.8}{4.3}{2.9} & \valu{4.4}{4.4}{0.6}{7.7}{3.2} & \valbu{7.6}{7.6}{0.6}{7.7}{3.2} & \val{7.4}{7.4}{0.8}{7.7}{3.2} & \val{6.6}{6.6}{0.8}{7.7}{3.2} & \valu{1.5}{1.5}{0.6}{4.3}{1.0} & \valb{3.2}{3.2}{0.6}{4.3}{1.0} & \val{2.8}{2.8}{0.8}{4.3}{1.0} & \val{2.8}{2.8}{0.8}{4.3}{1.0} \\
\bottomrule
\end{tabular*}
\caption{\textbf{Distillation improvement under jointly optimized $\bm{\alpha}$ across all benchmarks.} The student is 1.7B trained on 50B tokens. For each teacher configuration, a single $\alpha$ is selected to maximize average improvement across all 27 benchmarks. Results closely match those from per-metric optimization (\tabref{tab:exp1_results}), confirming that our findings are robust to $\alpha$ selection methodology.}
\label{tab:best_alpha_all}
\end{table*}

In the main experiments (\secrefc{sec:results}), we select the optimal loss mixing coefficient $\alpha$ independently for each evaluation type, allowing different $\alpha$ values for in-domain perplexity, out-of-distribution perplexity, and downstream accuracy. Here we present results using a single $\alpha$ per teacher configuration, selected to maximize average improvement across all benchmarks.

\paragraph{Methodology.} For each teacher configuration, we compute the average percentage improvement across all 27 benchmarks (1 in-domain perplexity + 11 out-of-distribution perplexity corpora + 15 downstream accuracy tasks), weighting each benchmark equally. We then select the $\alpha$ that maximizes this average. This approach naturally weights toward out-of-distribution and downstream metrics, which comprise 26 of the 27 benchmarks.

\paragraph{Results.} \tabref{tab:best_alpha_all} shows the results under jointly optimized $\alpha$. The patterns closely match those from per-metric optimization in \tabref{tab:exp1_results}: weak-to-strong distillation remains effective, same-level distillation yields improvement, and stronger teachers do not always produce better students. The jointly optimal $\alpha$ values follow the same trend as per-metric values, where weaker teachers favor lower $\alpha$, while stronger teachers favor higher $\alpha$. The slight differences in improvement magnitude (compared to \tabref{tab:exp1_results}) reflect the trade-off from using a single $\alpha$ across metrics with potentially different optima. These results confirm that our findings are robust to the $\alpha$ selection methodology.

\section{Training Curves}
\label{appendix:train_curve}

\begin{figure*}[!t]
  \centering
  \includegraphics[width=\textwidth]{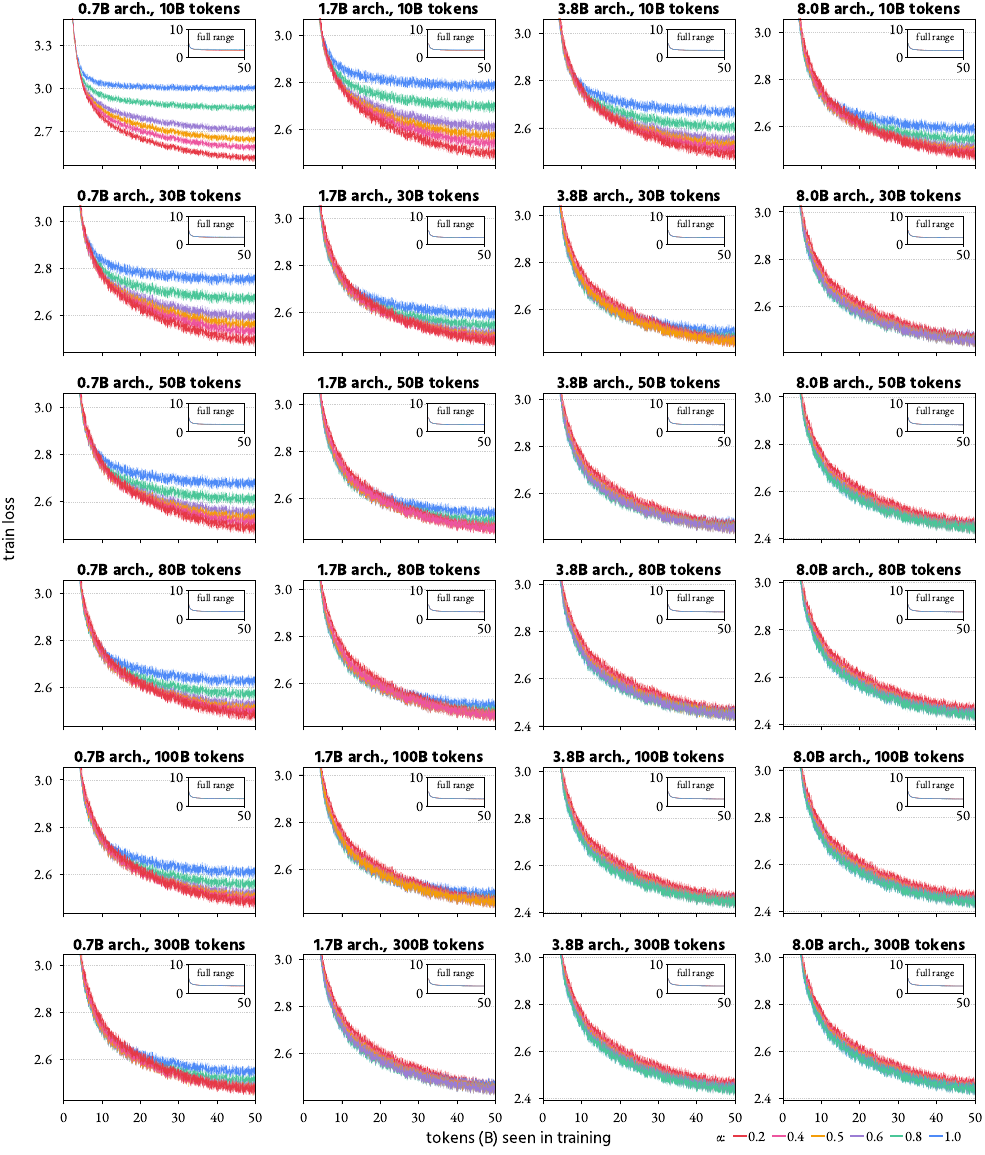}
  \caption{Training loss curves for distillation pretraining across all teacher configurations. Columns are ordered by teacher architecture size and rows by teacher token budget. Each subplot shows training loss versus tokens seen by the 1.7B student, zoomed in for clarity, with different colors indicating loss mixing coefficient $\alpha$. Insets show the full y-axis range for reference. Across all configurations, higher $\alpha$ leads to faster loss reduction in the initial training stage, regardless of whether it eventually converges to the lowest or highest final loss among the $\alpha$ values.}
  \label{fig:train_curves}
\end{figure*}

\figref{fig:train_curves} shows training loss curves for distillation pretraining across all teacher configurations. Each subplot corresponds to a teacher configuration, with different colors indicating the loss mixing coefficient $\alpha$. The distillation training runs converge steadily. Across all configurations, higher $\alpha$ leads to faster loss reduction in the initial training stage, as the student receives stronger supervision from the teacher's distribution. However, this initial advantage does not always translate to lower final loss: for weak teachers, high $\alpha$ eventually converges to higher loss than low $\alpha$, while for strong teachers, high $\alpha$ maintains its advantage throughout.

\newpage

\section{Additional Ablations}
\label{app:additional_ablations}

\begin{table}[h]
\newsavebox{\abtablebox}
\savebox{\abtablebox}{%
\begin{minipage}[t]{0.48\textwidth}
  \centering
  \small
  \setlength{\tabcolsep}{3pt}
  \begin{tabular}{l cccc}
  \toprule
  & 0.4B & 0.5B & 0.8B & 1.2B \\
  \midrule
  in-domain ppl & \valpha{\textcolor{tabred}{-2.5\%}}{0.2} & \valpha{\textcolor{tabred}{-0.1\%}}{0.2} & \valpha{\textcolor{tabgreen}{+2.1\%}}{0.5} & \valpha{\textcolor{tabgreen}{+3.2\%}}{0.8} \\[10pt]
  ood ppl & \valpha{\textcolor{tabgreen}{+0.2\%}}{0.2} & \valpha{\textcolor{tabgreen}{+1.6\%}}{0.4} & \valpha{\textcolor{tabgreen}{+6.5\%}}{0.5} & \valpha{\textcolor{tabgreen}{+6.1\%}}{0.6} \\[10pt]
  downstream acc & \valpha{\textcolor{tabgreen}{+0.1\%}}{0.2} & \valpha{\textcolor{tabgreen}{+0.2\%}}{0.2} & \valpha{\textcolor{tabgreen}{+1.0\%}}{0.2} & \valpha{\textcolor{tabgreen}{+0.7\%}}{0.4} \\[4pt]
  \bottomrule
  \end{tabular}
  \caption{\textbf{Distillation improvement with Qwen3 architecture.} Student is 0.5B trained on 10B tokens. Results show percentage improvement over standard baseline.}
  \label{tab:ablation_arch}
\end{minipage}%
}%
\usebox{\abtablebox}%
\hfill
\raisebox{6pt}{\begin{minipage}[t][\dimexpr\ht\abtablebox+\dp\abtablebox\relax][s]{0.48\textwidth}
  \centering
  \small
  \begin{tabular*}{\textwidth}{@{\hspace{2pt}\extracolsep{\fill}} l cccc @{\hspace{2pt}}}
  \toprule
  & 0.7B & 1.7B & 3.8B & 8.0B \\
  \midrule
  in-domain ppl & \textcolor{tabred}{-0.8\%} & \textcolor{tabgreen}{+0.2\%} & \textcolor{tabgreen}{+0.9\%} & \textcolor{tabgreen}{+0.9\%} \\[6pt]
  ood ppl & \textcolor{tabgreen}{+2.4\%} & \textcolor{tabgreen}{+2.8\%} & \textcolor{tabgreen}{+2.4\%} & \textcolor{tabgreen}{+3.4\%} \\[6pt]
  downstream acc & \textcolor{tabgreen}{+0.3\%} & \textcolor{tabgreen}{+1.0\%} & \textcolor{tabgreen}{+3.4\%} & \textcolor{tabgreen}{+3.1\%} \\
  \bottomrule
  \end{tabular*}
  \vfill
  \caption{\textbf{Distillation improvement at 300B training tokens with $\bm{\alpha=0.2}$.} Student is 1.7B trained on 300B tokens. Results show percentage improvement over standard pretraining 300B baseline.}
  \label{tab:ablation_tokens}
\end{minipage}}
\end{table}

\subsection{Generalization to Different Architecture Families}
\label{sec:ablation_arch}

Our main experiments use the Llama3 architecture family. We now test whether our findings transfer to a different architecture by using Qwen3~\citep{qwen3}, with variants of 0.4B, 0.5B, 0.8B, and 1.2B parameters. We use the 0.5B model as the student and train teachers, student, and baseline for 10B tokens, sweeping $\alpha$ values. The results in \tabref{tab:ablation_arch} again confirm our main findings: weak-to-strong distillation works,
and larger teachers do not always yield better distillation.

\subsection{Generalization to More Training Tokens}
\label{sec:ablation_tokens}

Our main experiments train students on 50B tokens. We now test whether our findings hold at larger scale by training with 300B tokens for teacher, student, and baseline. We evaluate only $\alpha=0.2$ as it is easiest for improvement. The results in \tabref{tab:ablation_tokens} confirm our main findings: weak-to-strong distillation remains effective, and at this $\alpha$, the 3.8B teacher outperforms the 8.0B teacher in downstream performance.

\section{Full Evaluation Results}
\label{appendix:eval}

We report full per-benchmark evaluation results for all teacher-student configurations. \tabref{tab:full_eval_ppl} presents perplexity results across 12 benchmarks, including in-domain and out-of-distribution evaluations. \tabref{tab:full_eval_acc} presents downstream accuracy results across 15 benchmarks. For each teacher configuration, we report the best-performing loss mixing coefficient $\alpha$ along with the corresponding improvement over the baseline student.

\section{Mechanism Study: Per-Token Analysis of Knowledge Distillation}
\label{appendix:mechanism}

To understand \textit{how} distillation transfers knowledge, we conduct a per-token mechanism study. We collect per-token logit data ($\sim$500K tokens per model) on three datasets: FineWeb-Edu (in-domain), C4 (out-of-distribution), and WikiText (out-of-distribution). We analyze 29 models: a 1.7B baseline (standard pretraining), 4 teachers (0.5B, 1.7B, 3.8B, 8.0B, all trained on 50B tokens), and 24 distilled 1.7B students (4 teachers $\times$ 6 alpha values: 0.2, 0.4, 0.5, 0.6, 0.8, 1.0).

\subsection{Overview: PPL Improvement Across Alpha and Datasets}

\begin{figure*}[!t]
  \centering
  \hspace*{-0.3cm}
  \includegraphics[width=\textwidth, trim=0 0 0 0.5cm, clip]{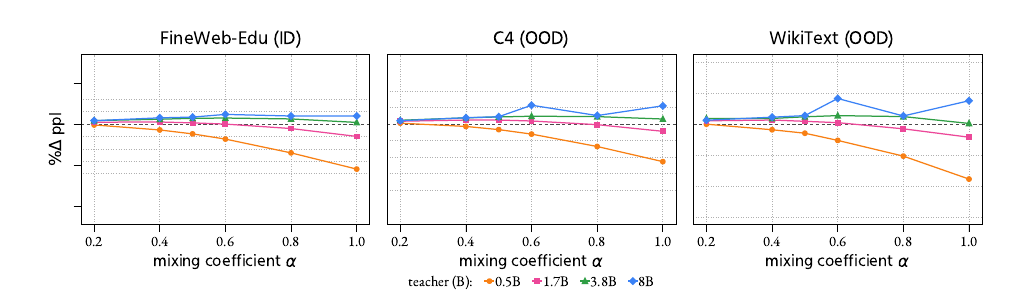}
  \vspace{-.4cm}
  \caption{\textbf{Percentage PPL improvement over baseline across alpha values and datasets.} Each panel is a dataset (FineWeb-Edu in-domain, C4 and WikiText out-of-distribution). Lines represent teacher sizes; the dashed line at $y$=0 is the baseline. Strong teachers improve the student across all alpha values; weak teachers degrade performance at high alpha. Optimal alpha increases with teacher strength.}
  \label{fig:exp5_ppl_vs_alpha}
  \vspace{-.1cm}
\end{figure*}

\begin{figure*}[!t]
  \centering
  \includegraphics[width=\textwidth, trim=0 0 0 0.5cm, clip]{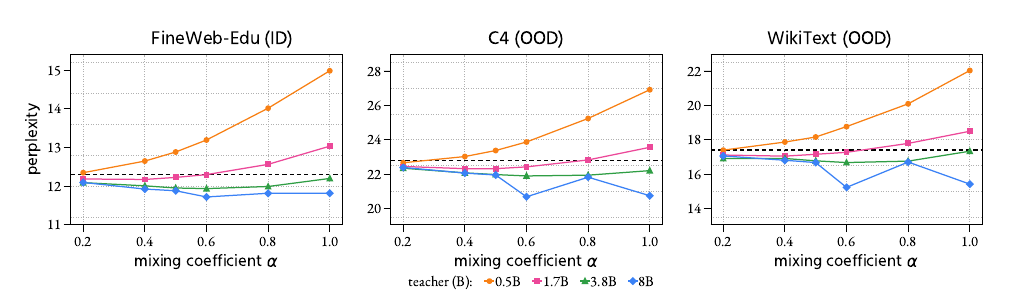}
  \vspace{-.8cm}
  \caption{\textbf{Raw perplexity values across alpha values and datasets.} The absolute perplexity instead of relative improvement. Dashed lines indicate the baseline. Lines below the dashed line indicate better-than-baseline performance.}
  \label{fig:exp5_ppl_raw}
\end{figure*}

\figref{fig:exp5_ppl_vs_alpha} shows percentage PPL improvement over the baseline as alpha varies. Strong teachers (8.0B, 3.8B) consistently improve the student across all datasets, with 8.0B achieving up to $\sim$9\% improvement on C4. The 1.7B same-size teacher provides modest improvement ($\sim$2\%), peaking around $\alpha$=0.4. The 0.5B teacher hurts at all $\alpha > 0.2$, with degradation increasing sharply at higher alpha (up to $-$18\% on C4 at $\alpha$=1.0). Optimal alpha increases with teacher quality: $\sim$0.2 for 0.5B, $\sim$0.4 for 1.7B, and $\sim$0.6 for 3.8B/8.0B. These patterns are consistent across in-domain and out-of-distribution datasets. \figref{fig:exp5_ppl_raw} shows the corresponding raw perplexity.

\subsection{Distillation Improves Hard Tokens}
\label{appendix:mechanism_difficulty}

A central question is whether distillation acts as a form of regularization (benefiting tokens uniformly) or as genuine knowledge transfer (benefiting tokens where the baseline is most uncertain). We test this by binning tokens into four difficulty levels based on the baseline model's entropy $H$: easy ($H<2$), moderate ($2 \leq H < 5$), hard ($5 \leq H < 8$), and difficult ($H \geq 8$).

\begin{figure*}[!t]
  \centering
  \vspace{-.8cm}
  \includegraphics[width=\textwidth, trim=0 0 0 0.5cm, clip]{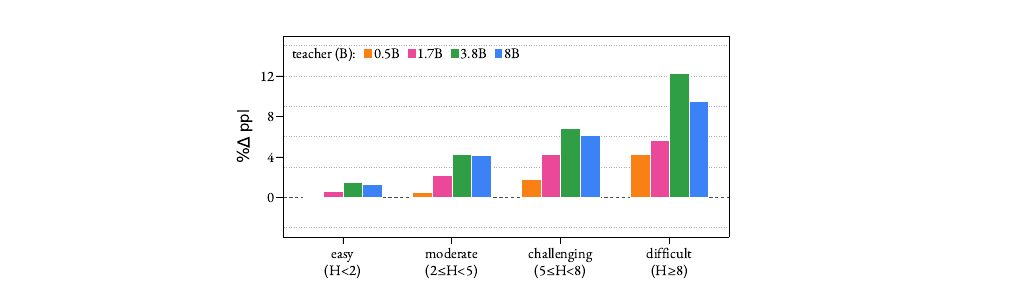}
  \vspace{-.8cm}
  \caption{\textbf{Improvement by token difficulty at each teacher's best alpha.} Token difficulty bins are defined by the baseline model's entropy. Improvement is concentrated on harder tokens for all, refuting the regularization hypothesis. Stronger teachers show larger improvement at every level, but the monotonic concentration on hard tokens is universal.}
  \label{fig:exp5_token_difficulty}
  \vspace{-.4cm}
\end{figure*}

\figref{fig:exp5_token_difficulty} shows PPL improvement by difficulty bin at each teacher's best alpha. Improvement is monotonically concentrated on harder tokens for \textit{all} teachers. Easy tokens see $\sim$0--1\% improvement while difficult tokens see $\sim$4--12\% improvement. This pattern directly refutes the regularization hypothesis: regularization such as label smoothing would produce uniform improvement across difficulty levels. Instead, distillation specifically helps where the baseline is most uncertain. Stronger teachers provide more improvement at every difficulty level, but the concentration on hard tokens is universal, even the 0.5B teacher at its best $\alpha$=0.2 shows the same pattern at smaller magnitude.

\begin{figure*}[!t]
  \centering
  \includegraphics[width=\textwidth]{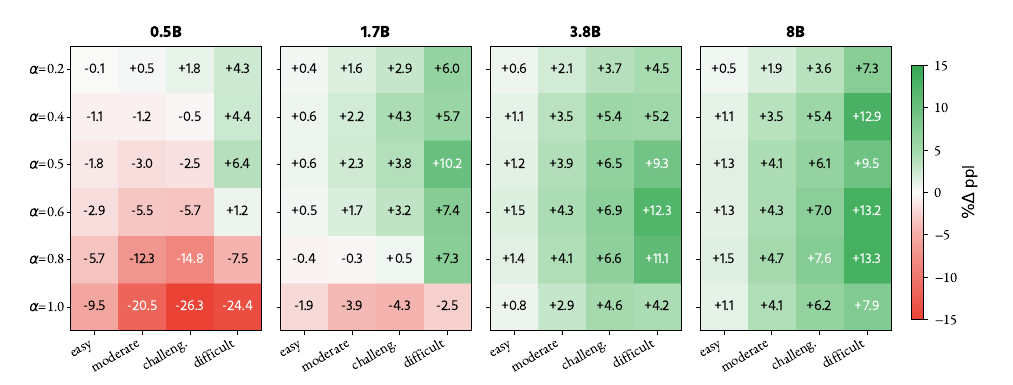}
  \vspace{-.7cm}
  \caption{\textbf{PPL improvement by token difficulty across all alpha values.} Each panel corresponds to a teacher size. Rows are alpha values (0.2 to 1.0); columns are difficulty bins. Green indicates improvement; red indicates degradation. For weak teachers, degradation spreads from easy to hard tokens as alpha increases. For strong teachers, improvement is robust across all alpha values, with the strongest gains consistently on difficult tokens.}
  \label{fig:exp5_difficulty_heatmap}
  \vspace{-.4cm}
\end{figure*}

\figref{fig:exp5_difficulty_heatmap} provides the complete picture across all 24 students, not just best alpha. For the 0.5B teacher, low alpha yields mild improvement on hard tokens, but as alpha increases, easy and moderate tokens degrade first, followed by hard tokens. At $\alpha$=1.0, all bins are deeply negative ($-$9.5\% to $-$26.3\%). For the 3.8B and 8.0B teachers, virtually all cells are positive, with the difficult-token column consistently showing the strongest improvement (up to +13.3\% for 8.0B at $\alpha$=0.4--0.8).

\begin{figure*}[!t]
  \centering
  \includegraphics[width=\textwidth]{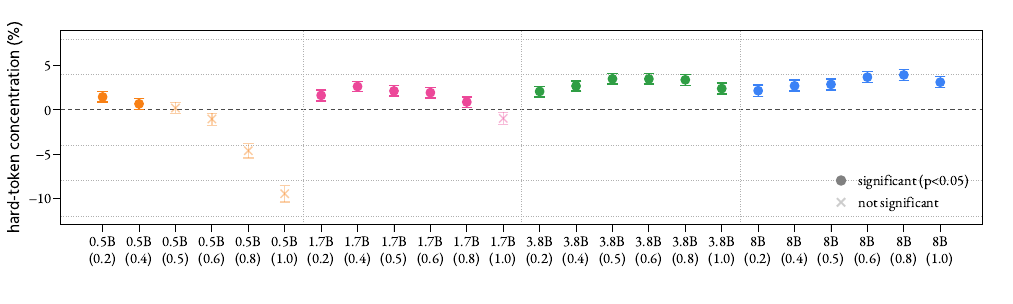}
  \vspace{-1.0cm}
  \caption{\textbf{Statistical significance of hard-token concentration.} Hard-token concentration is the improvement gap between hard and easy tokens. Filled circles indicate statistical significance ($p<0.05$, bootstrap with 1000 resamples); faded crosses indicate non-significance. Error bars show 95\% confidence intervals. All students with positive overall improvement show significant hard-token concentration. When distillation hurts (0.5B at high alpha), concentration reverses, where hard tokens are damaged more than easy tokens.}
  \label{fig:exp5_bootstrap_ci}
\end{figure*}

\figref{fig:exp5_bootstrap_ci} confirms statistical significance via bootstrap confidence intervals. We define hard-token concentration as the improvement gap between hard and easy tokens. For all students with positive overall improvement, 100\% show statistically significant hard-token concentration ($p<0.001$). Non-significant cases are confined to edge cases where overall improvement is near zero or negative (0.5B at $\alpha \geq 0.5$, 1.7B at $\alpha$=1.0). When distillation hurts (0.5B at $\alpha$=0.8, 1.0), the concentration goes strongly negative: hard tokens are damaged \textit{more} than easy tokens, consistent with a weak teacher actively misleading student's most uncertain predictions.

\vspace{-.5cm}
\subsection{Novel Information from the Teacher}
\label{appendix:mechanism_info_gain}

To identify the causal mechanism behind distillation's improvement, we classify each token into four categories based on whether the ground-truth token appears in the top-10 predictions of the baseline and teacher: (1) ``both correct'' ($\sim$68\% of tokens), (2) ``teacher only (novel info)'' ($\sim$2--5\%), (3) ``baseline only (teacher wrong)'' ($\sim$2--4\%), and (4) ``neither correct'' ($\sim$23\%).

\begin{figure*}[!t]
  \centering
  \includegraphics[width=\textwidth]{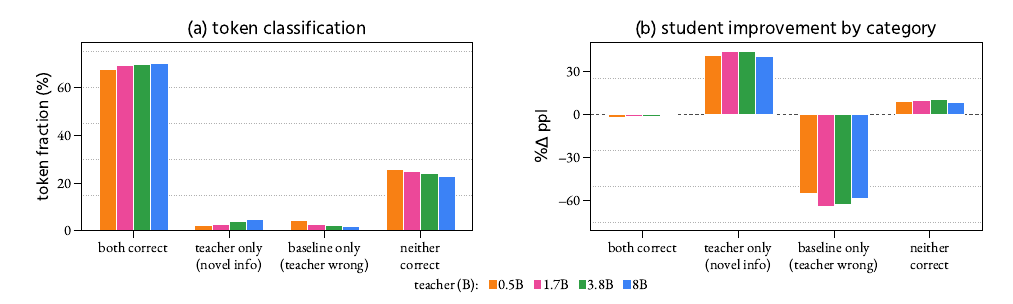}
  \vspace{-.6cm}
  \caption{\textbf{Causal analysis of distillation by token category.} Tokens are classified by whether the ground-truth appears in the top-10 predictions of the baseline and/or teacher. Panel (a): fraction of tokens in each category. Panel (b): student PPL improvement per category. Improvement comes from ``teacher only'' tokens (+30--40\%) where the teacher provides novel correct predictions. ``Baseline only'' tokens ($-$55\% to $-$65\%) represent the cost of distillation. This explains why $\alpha<1$ is optimal: it limits damage from tokens where the teacher is wrong but the baseline is right.}
  \label{fig:exp5_teacher_info_gain}
  \vspace{-.6cm}
\end{figure*}

\figref{fig:exp5_teacher_info_gain} reveals the mechanism. On ``both correct'' tokens, the student shows $\sim$0\% improvement --- when both models already know the answer, the teacher adds nothing. On ``teacher only'' tokens, improvement is massive (+30--40\%) --- the teacher provides novel correct information that the baseline lacks, and the student absorbs it. On ``baseline only'' tokens, the teacher actively misleads the student ($-$55\% to $-$65\%) --- this is the cost of distillation. On ``neither correct'' tokens, the student still improves ($\sim$+10\%), suggesting the teacher's soft distribution is informative even when neither model's top predictions are correct.

These results explain two key observations. First, stronger teachers help more because they have more ``teacher only'' tokens (8.0B: 4.9\% vs.\ 0.5B: 2.3\%). Second, $\alpha < 1$ is optimal because it limits the damage from the ``baseline only'' category where the teacher's signal is harmful.

\subsection{Distillation Is Not Label Smoothing}
\label{appendix:mechanism_ls}

\begin{figure*}[h]
  \centering
  \includegraphics[width=\textwidth, trim=0 0 0 0.5cm, clip]{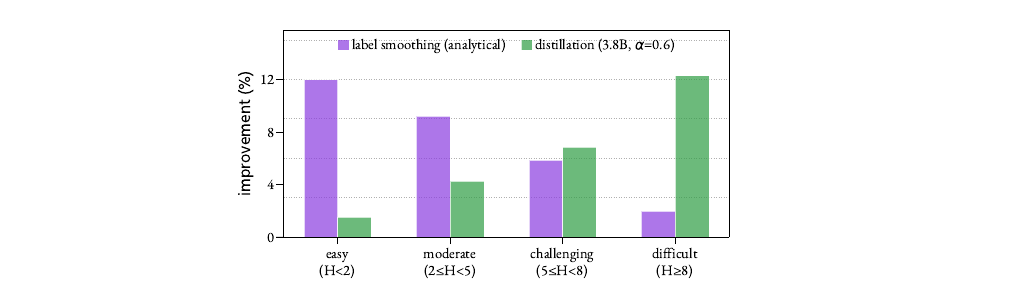}
  \vspace{-.9cm}
  \caption{\textbf{Distillation vs.\ label smoothing: opposite benefit profiles.} Purple bars show the analytical label smoothing benefit profile (proportional to $\log V - H$); green bars show the measured distillation improvement (3.8B teacher, $\alpha$=0.6). Label smoothing helps most on easy tokens; distillation helps most on hard tokens. The crossing pattern refutes the hypothesis that distillation is a form of regularization.}
  \label{fig:exp5_kd_vs_label_smoothing}
\end{figure*}

A common hypothesis is that distillation acts similarly to label smoothing, a uniform regularizer. \figref{fig:exp5_kd_vs_label_smoothing} directly tests this by comparing the token-level benefit profiles of label smoothing and distillation. Label smoothing's benefit is analytically proportional to $\log V - H$, where $V$ is the vocabulary size and $H$ is the token entropy, meaning it helps most on easy tokens where the model is overconfident. Distillation's measured benefit shows the opposite pattern: it helps most on hard tokens.

The two profiles cross: label smoothing benefit decreases with token difficulty while distillation benefit increases. This definitively refutes the hypothesis that distillation is merely a form of regularization similar to label smoothing, they operate on complementary parts of the token difficulty spectrum.

\subsection{Distribution Convergence: Weak Teachers Distort, Strong Teachers Transfer}
\label{appendix:mechanism_convergence}

We measure whether the distilled student's output distribution moves toward the teacher or stays close to the baseline by computing a convergence ratio: the top-10 token overlap between student and teacher, divided by the overlap between the distillation pretraining student and standard pretraining baseline. A ratio of 1 means equidistant; ratio $>$ 1 means the student is closer to the teacher.

\begin{figure*}[h]
  \centering
  \includegraphics[width=\textwidth, trim=0 0 0 0.5cm, clip]{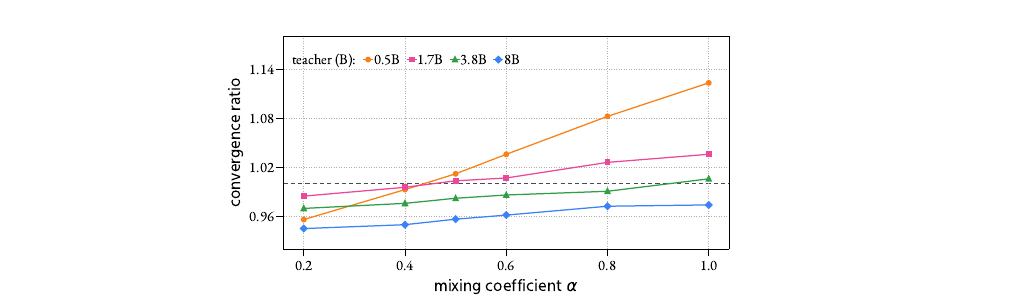}
  \vspace{-.65cm}
  \caption{\textbf{Distribution convergence ratio across alpha.} The ratio measures top-10 token overlap (student--teacher) divided by overlap (student--baseline). Ratio $>$ 1 means the student is closer to the teacher; ratio $<$ 1 means closer to the baseline. The 0.5B teacher distorts the student distribution at high alpha (ratio up to 1.12), while the 3.8B and 8.0B teachers transfer knowledge without distortion (ratio $\leq$ 1.01). This explains why weak teachers require low alpha.}
  \label{fig:exp5_convergence_ratio}
  \vspace{-.5cm}
\end{figure*}

\figref{fig:exp5_convergence_ratio} shows that the 0.5B teacher distorts the student distribution: the ratio rises from 0.96 to 1.12 as alpha increases, meaning the student's predictions increasingly resemble the weaker teacher's. In contrast, the 3.8B teacher ($\sim$0.97--1.01) and 8.0B teacher ($\sim$0.95--0.97) transfer knowledge without distorting the student's distribution. This mechanistically explains why optimal alpha is lower for weak teachers: at high alpha, weak teachers pull the student away from its natural distribution, causing the performance degradation.

\subsection{Probability Redistribution, Not Sharpening}
\label{appendix:mechanism_entropy}
\vspace{-.2cm}

\begin{figure*}[h]
  \centering
  \includegraphics[width=\textwidth, trim=0 0 0 0.5cm, clip]{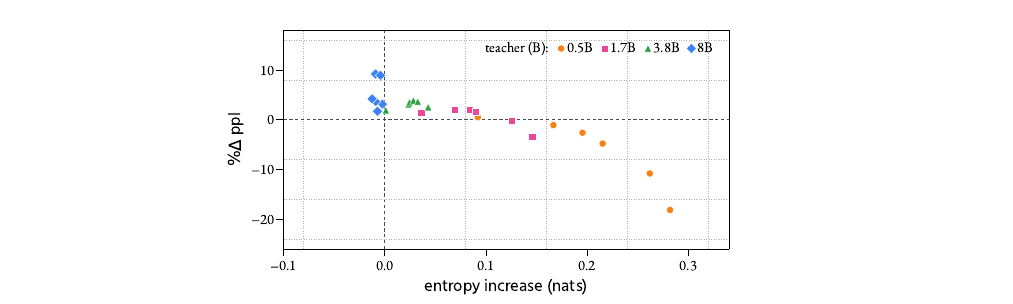}
  \vspace{-.6cm}
  \caption{\textbf{Entropy change vs.\ PPL improvement.} Each point is one distilled student. Most successful students show increased entropy (less confident) yet improved PPL (more accurate), indicating distillation redistributes probability rather than sharpening predictions. Strong teachers achieve high improvement with modest entropy change; the 0.5B teacher at high alpha increases entropy substantially while degrading PPL.}
  \label{fig:exp5_entropy_vs_ppl}
\end{figure*}

A natural expectation is that distillation sharpens predictions (lowering entropy) to improve perplexity. \figref{fig:exp5_entropy_vs_ppl} tests this by plotting each student's entropy change against its PPL improvement. Most successful students have \textit{positive} entropy increase (less confident) yet \textit{positive} PPL improvement (better predictions). This counterintuitive result indicates that distillation redistributes probability mass from wrong confident predictions to better-calibrated ones, rather than simply sharpening.

Strong teachers (8.0B, 3.8B) cluster in the upper region, modest entropy change with strong PPL improvement, indicating efficient knowledge transfer. The 0.5B teacher at high alpha occupies the lower-right region, large entropy increase \textit{and} PPL degradation, the weak teacher makes the student both less confident and accurate.

\subsection{Position Independence}
\label{appendix:mechanism_position}

\begin{figure*}[h]
  \centering
  \includegraphics[width=\textwidth, trim=0 0 0 0.5cm, clip]{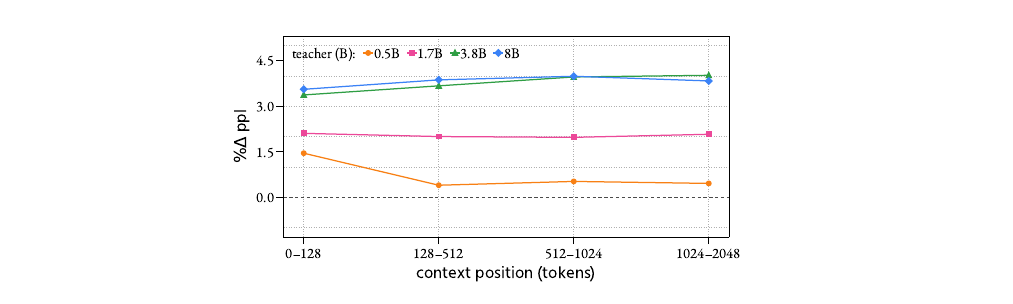}
  \vspace{-.8cm}
  \caption{\textbf{PPL improvement by context position.} Each line represents a teacher at its best alpha, with improvement measured in four position bins across the context window. Improvement is uniform across positions for all teachers, ruling out position-dependent effects and supporting zthat distillation transfers token-level predictive knowledge.}
  \label{fig:exp5_position_dependence}
\end{figure*}

\figref{fig:exp5_position_dependence} tests whether distillation improvement varies with position in the context window. PPL improvement is computed in four position bins (0--128, 128--512, 512--1024, 1024--2048 tokens). All teachers show essentially flat improvement across positions, there is no early-token vs.\ late-token effect. This rules out the hypothesis that distillation primarily helps with ``cold start'' predictions (early tokens with little context) or with long-range dependencies (late tokens). The uniformity supports that distillation transfers token-level knowledge about \textit{what} to predict given the context, rather than position-dependent structural knowledge.

\newpage
\mbox{}

\footnotesize
\setlength{\tabcolsep}{0pt}
\setlength{\aboverulesep}{0pt}
\setlength{\belowrulesep}{0pt}

\begin{xltabular}{\textwidth}{@{\hspace{12pt}} l *{12}{>{\centering\arraybackslash}X} @{\hspace{12pt}}}
\caption{\textbf{Full perplexity evaluation results.} We include (1) in-domain perplexity evaluation on FineWeb-Edu (FW); (2) out-of-distribution perplexity evaluation on Wikitext-103 (Wiki), C4 (C4), GSM8K (GSM), DM Mathematics (DMM), HumanEval (HEval), CodeSearchNet (CSN), arXiv (AX), CNN-DailyMail (CNN), ECHR (ECHR), PubMedQA (PM), and XQuAD (XQA).} \label{tab:full_eval_ppl} \\
\toprule
\rule[-3pt]{0pt}{12pt}perplexity ($\downarrow$) & FW & Wiki & C4 & GSM & DMM & HEval & CSN & AX & CNN & ECHR & PM & XQA \\
\midrule
\endfirsthead
\multicolumn{13}{l}{\small\textit{Table~\ref{tab:full_eval_ppl} continued from previous page}} \\[4pt]
\toprule
\rule[-3pt]{0pt}{12pt}perplexity ($\downarrow$) & FW & Wiki & C4 & GSM & DMM & HEval & CSN & AX & CNN & ECHR & PM & XQA \\
\midrule
\noalign{\vskip 2pt}
\endhead
\midrule
\noalign{\vskip 4pt}
\multicolumn{13}{r}{\small\textit{Continued on next page}} \\
\endfoot
\bottomrule
\endlastfoot

\rowcolor{rowpurple}
\multicolumn{13}{l}{\hspace{6pt}\rule[-3.5pt]{0pt}{12pt}$\rightarrow$ \textit{standard pretrained student baseline}} \\
\rule{0pt}{10pt} & 12.17 & 16.01 & 21.53 & 6.61 & 9.33 & 13.54 & 20.79 & 24.48 & 14.27 & 9.24 & 24.77 & 12.13 \\
\specialrule{\lightrulewidth}{0pt}{0pt}
\rowcolor{rowgreen}
\multicolumn{13}{l}{\hspace{6pt}\rule[-3.5pt]{0pt}{12pt}$\rightarrow$ \textit{0.7B arch. 10B tokens teacher config ($\alpha$=0.2)}} \\
\rule{0pt}{10pt}pretrained teacher & 21.21 & 32.96 & 36.50 & 16.94 & 14.44 & 44.51 & 79.14 & 47.53 & 25.72 & 15.10 & 50.07 & 74.66 \\
distilled student & 12.52 & 16.49 & 21.92 & 6.91 & 9.15 & 14.06 & 20.70 & 25.33 & 14.70 & 9.44 & 24.88 & 14.02 \\
improvement (\%) & \textcolor{tabred}{-2.9} & \textcolor{tabred}{-3.0} & \textcolor{tabred}{-1.8} & \textcolor{tabred}{-4.5} & \textcolor{tabgreen}{+1.9} & \textcolor{tabred}{-3.8} & \textcolor{tabgreen}{+0.4} & \textcolor{tabred}{-3.5} & \textcolor{tabred}{-3.1} & \textcolor{tabred}{-2.1} & \textcolor{tabred}{-0.5} & \textcolor{tabred}{-15.6} \\
\specialrule{\lightrulewidth}{0pt}{0pt}
\rowcolor{rowblue}
\multicolumn{13}{l}{\hspace{6pt}\rule[-3.5pt]{0pt}{12pt}$\rightarrow$ \textit{0.7B arch. 30B tokens teacher config ($\alpha$=0.2)}} \\
\rule{0pt}{10pt}pretrained teacher & 16.25 & 23.39 & 28.30 & 10.48 & 11.05 & 24.79 & 36.41 & 35.11 & 19.31 & 11.89 & 37.08 & 21.47 \\
distilled student & 12.32 & 16.03 & 21.55 & 6.74 & 9.10 & 14.06 & 20.91 & 24.23 & 14.35 & 9.30 & 24.83 & 12.29 \\
improvement (\%) & \textcolor{tabred}{-1.2} & \textcolor{tabred}{-0.1} & \textcolor{tabred}{-0.1} & \textcolor{tabred}{-1.9} & \textcolor{tabgreen}{+2.4} & \textcolor{tabred}{-3.8} & \textcolor{tabred}{-0.6} & \textcolor{tabgreen}{+1.0} & \textcolor{tabred}{-0.6} & \textcolor{tabred}{-0.6} & \textcolor{tabred}{-0.2} & \textcolor{tabred}{-1.3} \\
\specialrule{\lightrulewidth}{0pt}{0pt}
\rowcolor{rowyellow}
\multicolumn{13}{l}{\hspace{6pt}\rule[-3.5pt]{0pt}{12pt}$\rightarrow$ \textit{0.7B arch. 50B tokens teacher config ($\alpha$=0.2)}} \\
\rule{0pt}{10pt}pretrained teacher & 14.91 & 20.62 & 26.06 & 8.65 & 10.54 & 19.86 & 30.64 & 31.23 & 17.66 & 10.98 & 32.49 & 17.37 \\
distilled student & 12.21 & 15.98 & 21.43 & 6.53 & 9.01 & 13.02 & 19.37 & 24.13 & 14.30 & 9.23 & 24.49 & 12.18 \\
improvement (\%) & \textcolor{tabred}{-0.3} & \textcolor{tabgreen}{+0.2} & \textcolor{tabgreen}{+0.5} & \textcolor{tabgreen}{+1.2} & \textcolor{tabgreen}{+3.3} & \textcolor{tabgreen}{+3.9} & \textcolor{tabgreen}{+6.8} & \textcolor{tabgreen}{+1.4} & \textcolor{tabred}{-0.3} & \textcolor{tabgreen}{+0.1} & \textcolor{tabgreen}{+1.1} & \textcolor{tabred}{-0.4} \\
\specialrule{\lightrulewidth}{0pt}{0pt}
\rowcolor{roworange}
\multicolumn{13}{l}{\hspace{6pt}\rule[-3.5pt]{0pt}{12pt}$\rightarrow$ \textit{0.7B arch. 80B tokens teacher config ($\alpha$=0.4)}} \\
\rule{0pt}{10pt}pretrained teacher & 14.01 & 19.35 & 24.54 & 8.02 & 10.40 & 15.52 & 23.99 & 29.33 & 16.49 & 10.40 & 31.18 & 12.83 \\
distilled student & 12.33 & 16.15 & 21.47 & 6.49 & 8.88 & 12.74 & 18.40 & 24.60 & 14.36 & 9.27 & 24.48 & 11.52 \\
improvement (\%) & \textcolor{tabred}{-1.3} & \textcolor{tabred}{-0.9} & \textcolor{tabgreen}{+0.3} & \textcolor{tabgreen}{+1.8} & \textcolor{tabgreen}{+4.7} & \textcolor{tabgreen}{+5.9} & \textcolor{tabgreen}{+11.5} & \textcolor{tabred}{-0.5} & \textcolor{tabred}{-0.7} & \textcolor{tabred}{-0.3} & \textcolor{tabgreen}{+1.2} & \textcolor{tabgreen}{+5.0} \\
\specialrule{\lightrulewidth}{0pt}{0pt}
\rowcolor{rowpink}
\multicolumn{13}{l}{\hspace{6pt}\rule[-3.5pt]{0pt}{12pt}$\rightarrow$ \textit{0.7B arch. 100B tokens teacher config ($\alpha$=0.2)}} \\
\rule{0pt}{10pt}pretrained teacher & 13.67 & 18.87 & 23.98 & 7.95 & 10.10 & 15.30 & 22.76 & 28.56 & 16.03 & 10.19 & 30.77 & 11.44 \\
distilled student & 12.13 & 15.74 & 21.26 & 6.53 & 8.87 & 12.79 & 19.18 & 24.27 & 14.17 & 9.15 & 24.10 & 11.31 \\
improvement (\%) & \textcolor{tabgreen}{+0.4} & \textcolor{tabgreen}{+1.7} & \textcolor{tabgreen}{+1.3} & \textcolor{tabgreen}{+1.2} & \textcolor{tabgreen}{+4.9} & \textcolor{tabgreen}{+5.6} & \textcolor{tabgreen}{+7.8} & \textcolor{tabgreen}{+0.9} & \textcolor{tabgreen}{+0.7} & \textcolor{tabgreen}{+1.0} & \textcolor{tabgreen}{+2.7} & \textcolor{tabgreen}{+6.8} \\
\specialrule{\lightrulewidth}{0pt}{0pt}
\rowcolor{rowcyan}
\multicolumn{13}{l}{\hspace{6pt}\rule[-3.5pt]{0pt}{12pt}$\rightarrow$ \textit{0.7B arch. 300B tokens teacher config ($\alpha$=0.6)}} \\
\rule{0pt}{10pt}pretrained teacher & 12.49 & 17.05 & 22.08 & 7.17 & 10.44 & 14.15 & 19.84 & 25.70 & 14.71 & 9.43 & 28.50 & 8.39 \\
distilled student & 12.21 & 15.94 & 21.28 & 6.46 & 8.77 & 12.76 & 18.65 & 24.09 & 14.22 & 9.16 & 24.04 & 10.15 \\
improvement (\%) & \textcolor{tabred}{-0.3} & \textcolor{tabgreen}{+0.5} & \textcolor{tabgreen}{+1.2} & \textcolor{tabgreen}{+2.4} & \textcolor{tabgreen}{+5.9} & \textcolor{tabgreen}{+5.8} & \textcolor{tabgreen}{+10.3} & \textcolor{tabgreen}{+1.6} & \textcolor{tabgreen}{+0.3} & \textcolor{tabgreen}{+0.9} & \textcolor{tabgreen}{+2.9} & \textcolor{tabgreen}{+16.3} \\
\specialrule{\lightrulewidth}{0pt}{0pt}
\rowcolor{rowlime}
\multicolumn{13}{l}{\hspace{6pt}\rule[-3.5pt]{0pt}{12pt}$\rightarrow$ \textit{1.7B arch. 10B tokens teacher config ($\alpha$=0.2)}} \\
\rule{0pt}{10pt}pretrained teacher & 16.77 & 23.56 & 29.12 & 11.10 & 11.43 & 26.62 & 43.59 & 35.87 & 19.87 & 12.18 & 35.05 & 41.09 \\
distilled student & 12.34 & 16.12 & 21.62 & 6.83 & 9.18 & 13.58 & 20.46 & 24.65 & 14.44 & 9.32 & 24.45 & 13.23 \\
improvement (\%) & \textcolor{tabred}{-1.4} & \textcolor{tabred}{-0.7} & \textcolor{tabred}{-0.4} & \textcolor{tabred}{-3.3} & \textcolor{tabgreen}{+1.6} & \textcolor{tabred}{-0.3} & \textcolor{tabgreen}{+1.6} & \textcolor{tabred}{-0.7} & \textcolor{tabred}{-1.2} & \textcolor{tabred}{-0.8} & \textcolor{tabgreen}{+1.3} & \textcolor{tabred}{-9.1} \\
\specialrule{\lightrulewidth}{0pt}{0pt}
\rowcolor{rowviolet}
\multicolumn{13}{l}{\hspace{6pt}\rule[-3.5pt]{0pt}{12pt}$\rightarrow$ \textit{1.7B arch. 30B tokens teacher config ($\alpha$=0.2)}} \\
\rule{0pt}{10pt}pretrained teacher & 13.14 & 17.76 & 23.13 & 7.39 & 9.85 & 16.29 & 23.70 & 26.95 & 15.47 & 9.88 & 27.04 & 14.69 \\
distilled student & 12.11 & 15.79 & 21.28 & 6.48 & 9.08 & 13.53 & 20.51 & 24.27 & 14.16 & 9.16 & 23.95 & 11.92 \\
improvement (\%) & \textcolor{tabgreen}{+0.5} & \textcolor{tabgreen}{+1.4} & \textcolor{tabgreen}{+1.2} & \textcolor{tabgreen}{+2.1} & \textcolor{tabgreen}{+2.6} & \textcolor{tabgreen}{+0.1} & \textcolor{tabgreen}{+1.4} & \textcolor{tabgreen}{+0.9} & \textcolor{tabgreen}{+0.7} & \textcolor{tabgreen}{+0.9} & \textcolor{tabgreen}{+3.3} & \textcolor{tabgreen}{+1.7} \\
\specialrule{\lightrulewidth}{0pt}{0pt}
\rowcolor{rowrose}
\multicolumn{13}{l}{\hspace{6pt}\rule[-3.5pt]{0pt}{12pt}$\rightarrow$ \textit{1.7B arch. 50B tokens teacher config ($\alpha$=0.4)}} \\
\rule{0pt}{10pt}pretrained teacher & 12.13 & 16.08 & 21.50 & 6.92 & 9.45 & 13.60 & 19.64 & 24.21 & 14.25 & 9.27 & 25.08 & 11.74 \\
distilled student & 12.05 & 15.66 & 21.08 & 6.41 & 9.07 & 13.24 & 18.96 & 23.93 & 14.06 & 9.09 & 23.78 & 11.16 \\
improvement (\%) & \textcolor{tabgreen}{+1.0} & \textcolor{tabgreen}{+2.2} & \textcolor{tabgreen}{+2.1} & \textcolor{tabgreen}{+3.0} & \textcolor{tabgreen}{+2.7} & \textcolor{tabgreen}{+2.2} & \textcolor{tabgreen}{+8.8} & \textcolor{tabgreen}{+2.3} & \textcolor{tabgreen}{+1.5} & \textcolor{tabgreen}{+1.7} & \textcolor{tabgreen}{+4.0} & \textcolor{tabgreen}{+8.0} \\
\specialrule{\lightrulewidth}{0pt}{0pt}
\rowcolor{rowteal}
\multicolumn{13}{l}{\hspace{6pt}\rule[-3.5pt]{0pt}{12pt}$\rightarrow$ \textit{1.7B arch. 80B tokens teacher config ($\alpha$=0.6)}} \\
\rule{0pt}{10pt}pretrained teacher & 11.46 & 15.01 & 20.42 & 6.17 & 9.31 & 12.31 & 18.19 & 22.96 & 13.48 & 8.82 & 23.53 & 8.83 \\
distilled student & 11.98 & 15.54 & 20.91 & 6.41 & 8.89 & 12.50 & 18.36 & 23.74 & 13.93 & 9.04 & 23.91 & 10.34 \\
improvement (\%) & \textcolor{tabgreen}{+1.6} & \textcolor{tabgreen}{+2.9} & \textcolor{tabgreen}{+2.9} & \textcolor{tabgreen}{+3.0} & \textcolor{tabgreen}{+4.7} & \textcolor{tabgreen}{+7.7} & \textcolor{tabgreen}{+11.7} & \textcolor{tabgreen}{+3.0} & \textcolor{tabgreen}{+2.3} & \textcolor{tabgreen}{+2.2} & \textcolor{tabgreen}{+3.5} & \textcolor{tabgreen}{+14.8} \\
\specialrule{\lightrulewidth}{0pt}{0pt}
\rowcolor{rowamber}
\multicolumn{13}{l}{\hspace{6pt}\rule[-3.5pt]{0pt}{12pt}$\rightarrow$ \textit{1.7B arch. 100B tokens teacher config ($\alpha$=0.6)}} \\
\rule{0pt}{10pt}pretrained teacher & 11.19 & 14.64 & 19.97 & 6.03 & 9.65 & 11.49 & 15.44 & 22.55 & 13.20 & 8.65 & 23.27 & 8.18 \\
distilled student & 11.93 & 15.60 & 20.83 & 6.29 & 9.22 & 12.22 & 17.74 & 23.54 & 13.87 & 8.98 & 23.85 & 10.07 \\
improvement (\%) & \textcolor{tabgreen}{+2.0} & \textcolor{tabgreen}{+2.6} & \textcolor{tabgreen}{+3.3} & \textcolor{tabgreen}{+4.9} & \textcolor{tabgreen}{+1.1} & \textcolor{tabgreen}{+9.8} & \textcolor{tabgreen}{+14.7} & \textcolor{tabgreen}{+3.8} & \textcolor{tabgreen}{+2.8} & \textcolor{tabgreen}{+2.9} & \textcolor{tabgreen}{+3.7} & \textcolor{tabgreen}{+17.0} \\
\specialrule{\lightrulewidth}{0pt}{0pt}
\rowcolor{rowsky}
\multicolumn{13}{l}{\hspace{6pt}\rule[-3.5pt]{0pt}{12pt}$\rightarrow$ \textit{1.7B arch. 300B tokens teacher config ($\alpha$=0.6)}} \\
\rule{0pt}{10pt}pretrained teacher & 10.17 & 13.17 & 18.50 & 5.40 & 10.38 & 10.17 & 14.82 & 20.48 & 12.16 & 8.07 & 21.46 & 6.11 \\
distilled student & 11.77 & 15.34 & 20.65 & 6.25 & 9.19 & 11.89 & 17.40 & 23.32 & 13.71 & 8.91 & 23.56 & 9.44 \\
improvement (\%) & \textcolor{tabgreen}{+3.3} & \textcolor{tabgreen}{+4.2} & \textcolor{tabgreen}{+4.1} & \textcolor{tabgreen}{+5.5} & \textcolor{tabgreen}{+1.5} & \textcolor{tabgreen}{+12.2} & \textcolor{tabgreen}{+16.3} & \textcolor{tabgreen}{+4.8} & \textcolor{tabgreen}{+3.9} & \textcolor{tabgreen}{+3.7} & \textcolor{tabgreen}{+4.9} & \textcolor{tabgreen}{+22.2} \\
\specialrule{\lightrulewidth}{0pt}{0pt}
\rowcolor{rowfuchsia}
\multicolumn{13}{l}{\hspace{6pt}\rule[-3.5pt]{0pt}{12pt}$\rightarrow$ \textit{3.8B arch. 10B tokens teacher config ($\alpha$=0.2)}} \\
\rule{0pt}{10pt}pretrained teacher & 14.43 & 19.48 & 25.18 & 8.70 & 10.31 & 22.34 & 34.10 & 29.85 & 16.99 & 10.67 & 29.79 & 34.46 \\
distilled student & 12.22 & 15.91 & 21.43 & 6.57 & 8.91 & 12.94 & 20.24 & 24.45 & 14.30 & 9.23 & 23.96 & 12.60 \\
improvement (\%) & \textcolor{tabred}{-0.4} & \textcolor{tabgreen}{+0.6} & \textcolor{tabgreen}{+0.5} & \textcolor{tabgreen}{+0.7} & \textcolor{tabgreen}{+4.5} & \textcolor{tabgreen}{+4.5} & \textcolor{tabgreen}{+2.6} & \textcolor{tabgreen}{+0.2} & \textcolor{tabred}{-0.2} & \textcolor{tabgreen}{+0.1} & \textcolor{tabgreen}{+3.3} & \textcolor{tabred}{-3.9} \\
\specialrule{\lightrulewidth}{0pt}{0pt}
\rowcolor{rowemerald}
\multicolumn{13}{l}{\hspace{6pt}\rule[-3.5pt]{0pt}{12pt}$\rightarrow$ \textit{3.8B arch. 30B tokens teacher config ($\alpha$=0.5)}} \\
\rule{0pt}{10pt}pretrained teacher & 11.20 & 14.53 & 20.04 & 6.19 & 8.63 & 12.32 & 18.08 & 22.28 & 13.25 & 8.71 & 22.98 & 11.27 \\
distilled student & 11.94 & 15.59 & 20.90 & 6.33 & 8.85 & 12.59 & 18.94 & 23.80 & 13.89 & 9.03 & 23.72 & 11.43 \\
improvement (\%) & \textcolor{tabgreen}{+1.9} & \textcolor{tabgreen}{+2.6} & \textcolor{tabgreen}{+2.9} & \textcolor{tabgreen}{+4.2} & \textcolor{tabgreen}{+5.1} & \textcolor{tabgreen}{+7.1} & \textcolor{tabgreen}{+8.9} & \textcolor{tabgreen}{+2.8} & \textcolor{tabgreen}{+2.6} & \textcolor{tabgreen}{+2.3} & \textcolor{tabgreen}{+4.2} & \textcolor{tabgreen}{+5.8} \\
\specialrule{\lightrulewidth}{0pt}{0pt}
\rowcolor{rowindigo}
\multicolumn{13}{l}{\hspace{6pt}\rule[-3.5pt]{0pt}{12pt}$\rightarrow$ \textit{3.8B arch. 50B tokens teacher config ($\alpha$=0.5)}} \\
\rule{0pt}{10pt}pretrained teacher & 10.30 & 13.22 & 18.71 & 5.64 & 9.03 & 11.08 & 15.08 & 20.33 & 12.28 & 8.16 & 20.90 & 8.38 \\
distilled student & 11.83 & 15.44 & 20.78 & 6.40 & 9.14 & 12.02 & 18.01 & 23.66 & 13.77 & 8.97 & 23.51 & 10.24 \\
improvement (\%) & \textcolor{tabgreen}{+2.8} & \textcolor{tabgreen}{+3.6} & \textcolor{tabgreen}{+3.5} & \textcolor{tabgreen}{+3.3} & \textcolor{tabgreen}{+2.0} & \textcolor{tabgreen}{+11.2} & \textcolor{tabgreen}{+13.4} & \textcolor{tabgreen}{+3.4} & \textcolor{tabgreen}{+3.5} & \textcolor{tabgreen}{+2.9} & \textcolor{tabgreen}{+5.1} & \textcolor{tabgreen}{+15.6} \\
\specialrule{\lightrulewidth}{0pt}{0pt}
\rowcolor{rowred}
\multicolumn{13}{l}{\hspace{6pt}\rule[-3.5pt]{0pt}{12pt}$\rightarrow$ \textit{3.8B arch. 80B tokens teacher config ($\alpha$=0.8)}} \\
\rule{0pt}{10pt}pretrained teacher & 9.64 & 12.18 & 17.77 & 5.23 & 8.65 & 9.10 & 12.72 & 18.89 & 11.60 & 7.81 & 20.07 & 6.92 \\
distilled student & 11.75 & 15.36 & 20.58 & 6.21 & 8.80 & 12.03 & 17.32 & 23.84 & 13.64 & 8.89 & 23.54 & 9.55 \\
improvement (\%) & \textcolor{tabgreen}{+3.5} & \textcolor{tabgreen}{+4.0} & \textcolor{tabgreen}{+4.4} & \textcolor{tabgreen}{+6.0} & \textcolor{tabgreen}{+5.6} & \textcolor{tabgreen}{+11.2} & \textcolor{tabgreen}{+16.7} & \textcolor{tabgreen}{+2.6} & \textcolor{tabgreen}{+4.4} & \textcolor{tabgreen}{+3.9} & \textcolor{tabgreen}{+4.9} & \textcolor{tabgreen}{+21.2} \\
\specialrule{\lightrulewidth}{0pt}{0pt}
\rowcolor{rowmint}
\multicolumn{13}{l}{\hspace{6pt}\rule[-3.5pt]{0pt}{12pt}$\rightarrow$ \textit{3.8B arch. 100B tokens teacher config ($\alpha$=0.8)}} \\
\rule{0pt}{10pt}pretrained teacher & 9.36 & 11.86 & 17.39 & 5.06 & 8.79 & 9.05 & 12.07 & 18.33 & 11.35 & 7.66 & 19.93 & 6.52 \\
distilled student & 11.70 & 15.27 & 20.55 & 6.42 & 9.05 & 12.07 & 17.17 & 23.25 & 13.62 & 8.88 & 23.37 & 9.61 \\
improvement (\%) & \textcolor{tabgreen}{+3.9} & \textcolor{tabgreen}{+4.7} & \textcolor{tabgreen}{+4.6} & \textcolor{tabgreen}{+2.9} & \textcolor{tabgreen}{+2.9} & \textcolor{tabgreen}{+10.9} & \textcolor{tabgreen}{+17.4} & \textcolor{tabgreen}{+5.1} & \textcolor{tabgreen}{+4.5} & \textcolor{tabgreen}{+3.9} & \textcolor{tabgreen}{+5.6} & \textcolor{tabgreen}{+20.8} \\
\specialrule{\lightrulewidth}{0pt}{0pt}
\rowcolor{rowpeach}
\multicolumn{13}{l}{\hspace{6pt}\rule[-3.5pt]{0pt}{12pt}$\rightarrow$ \textit{3.8B arch. 300B tokens teacher config ($\alpha$=0.8)}} \\
\rule{0pt}{10pt}pretrained teacher & 8.29 & 10.61 & 16.12 & 4.71 & 8.83 & 8.28 & 11.56 & 16.88 & 10.53 & 7.17 & 20.29 & 5.16 \\
distilled student & 11.65 & 15.24 & 20.54 & 6.24 & 9.08 & 12.25 & 17.88 & 23.47 & 13.60 & 8.89 & 23.45 & 9.48 \\
improvement (\%) & \textcolor{tabgreen}{+4.3} & \textcolor{tabgreen}{+4.8} & \textcolor{tabgreen}{+4.6} & \textcolor{tabgreen}{+5.6} & \textcolor{tabgreen}{+2.7} & \textcolor{tabgreen}{+9.6} & \textcolor{tabgreen}{+14.0} & \textcolor{tabgreen}{+4.2} & \textcolor{tabgreen}{+4.7} & \textcolor{tabgreen}{+3.9} & \textcolor{tabgreen}{+5.3} & \textcolor{tabgreen}{+21.9} \\
\specialrule{\lightrulewidth}{0pt}{0pt}
\rowcolor{rowlavender}
\multicolumn{13}{l}{\hspace{6pt}\rule[-3.5pt]{0pt}{12pt}$\rightarrow$ \textit{8.0B arch. 10B tokens teacher config ($\alpha$=0.4)}} \\
\rule{0pt}{10pt}pretrained teacher & 12.81 & 16.80 & 22.57 & 7.17 & 9.69 & 17.80 & 27.11 & 25.91 & 15.04 & 9.68 & 25.21 & 26.55 \\
distilled student & 12.23 & 15.95 & 21.33 & 6.50 & 8.97 & 13.01 & 18.43 & 24.20 & 14.23 & 9.19 & 23.88 & 13.40 \\
improvement (\%) & \textcolor{tabred}{-0.5} & \textcolor{tabgreen}{+0.4} & \textcolor{tabgreen}{+0.9} & \textcolor{tabgreen}{+1.8} & \textcolor{tabgreen}{+3.8} & \textcolor{tabgreen}{+3.9} & \textcolor{tabgreen}{+11.3} & \textcolor{tabgreen}{+1.2} & \textcolor{tabgreen}{+0.3} & \textcolor{tabgreen}{+0.5} & \textcolor{tabgreen}{+3.6} & \textcolor{tabred}{-10.5} \\
\specialrule{\lightrulewidth}{0pt}{0pt}
\rowcolor{rowapricot}
\multicolumn{13}{l}{\hspace{6pt}\rule[-3.5pt]{0pt}{12pt}$\rightarrow$ \textit{8.0B arch. 30B tokens teacher config ($\alpha$=0.5)}} \\
\rule{0pt}{10pt}pretrained teacher & 9.98 & 12.83 & 18.31 & 5.53 & 8.60 & 10.21 & 15.07 & 19.79 & 11.98 & 8.04 & 20.32 & 9.57 \\
distilled student & 11.82 & 15.46 & 20.77 & 6.29 & 8.99 & 12.50 & 18.50 & 23.87 & 13.80 & 8.98 & 23.24 & 10.83 \\
improvement (\%) & \textcolor{tabgreen}{+2.9} & \textcolor{tabgreen}{+3.5} & \textcolor{tabgreen}{+3.5} & \textcolor{tabgreen}{+4.8} & \textcolor{tabgreen}{+3.6} & \textcolor{tabgreen}{+7.7} & \textcolor{tabgreen}{+11.0} & \textcolor{tabgreen}{+2.5} & \textcolor{tabgreen}{+3.3} & \textcolor{tabgreen}{+2.8} & \textcolor{tabgreen}{+6.2} & \textcolor{tabgreen}{+10.7} \\
\specialrule{\lightrulewidth}{0pt}{0pt}
\rowcolor{rowaqua}
\multicolumn{13}{l}{\hspace{6pt}\rule[-3.5pt]{0pt}{12pt}$\rightarrow$ \textit{8.0B arch. 50B tokens teacher config ($\alpha$=0.8)}} \\
\rule{0pt}{10pt}pretrained teacher & 9.08 & 11.48 & 17.18 & 5.25 & 9.71 & 8.99 & 13.26 & 17.90 & 11.19 & 7.60 & 19.15 & 7.42 \\
distilled student & 11.70 & 15.36 & 20.61 & 6.36 & 9.18 & 11.89 & 17.42 & 23.52 & 13.62 & 8.90 & 23.32 & 9.74 \\
improvement (\%) & \textcolor{tabgreen}{+3.9} & \textcolor{tabgreen}{+4.0} & \textcolor{tabgreen}{+4.3} & \textcolor{tabgreen}{+3.8} & \textcolor{tabgreen}{+1.6} & \textcolor{tabgreen}{+12.2} & \textcolor{tabgreen}{+16.2} & \textcolor{tabgreen}{+3.9} & \textcolor{tabgreen}{+4.5} & \textcolor{tabgreen}{+3.8} & \textcolor{tabgreen}{+5.8} & \textcolor{tabgreen}{+19.7} \\
\specialrule{\lightrulewidth}{0pt}{0pt}
\rowcolor{rowcoral}
\multicolumn{13}{l}{\hspace{6pt}\rule[-3.5pt]{0pt}{12pt}$\rightarrow$ \textit{8.0B arch. 80B tokens teacher config ($\alpha$=0.6)}} \\
\rule{0pt}{10pt}pretrained teacher & 8.38 & 10.67 & 16.38 & 4.82 & 9.74 & 8.24 & 11.66 & 16.98 & 10.64 & 7.28 & 18.28 & 6.00 \\
distilled student & 11.70 & 15.30 & 20.66 & 6.28 & 8.87 & 11.77 & 17.93 & 23.36 & 13.64 & 8.90 & 23.72 & 9.58 \\
improvement (\%) & \textcolor{tabgreen}{+3.9} & \textcolor{tabgreen}{+4.4} & \textcolor{tabgreen}{+4.0} & \textcolor{tabgreen}{+5.1} & \textcolor{tabgreen}{+4.8} & \textcolor{tabgreen}{+13.1} & \textcolor{tabgreen}{+13.8} & \textcolor{tabgreen}{+4.6} & \textcolor{tabgreen}{+4.4} & \textcolor{tabgreen}{+3.7} & \textcolor{tabgreen}{+4.2} & \textcolor{tabgreen}{+21.0} \\
\specialrule{\lightrulewidth}{0pt}{0pt}
\rowcolor{rowperiwinkle}
\multicolumn{13}{l}{\hspace{6pt}\rule[-3.5pt]{0pt}{12pt}$\rightarrow$ \textit{8.0B arch. 100B tokens teacher config ($\alpha$=0.8)}} \\
\rule{0pt}{10pt}pretrained teacher & 8.08 & 10.31 & 15.99 & 4.71 & 9.34 & 7.95 & 10.58 & 16.38 & 10.42 & 7.15 & 17.84 & 5.67 \\
distilled student & 11.66 & 15.27 & 20.60 & 6.24 & 8.97 & 11.28 & 16.29 & 23.32 & 13.58 & 8.89 & 23.72 & 9.25 \\
improvement (\%) & \textcolor{tabgreen}{+4.2} & \textcolor{tabgreen}{+4.6} & \textcolor{tabgreen}{+4.3} & \textcolor{tabgreen}{+5.7} & \textcolor{tabgreen}{+3.8} & \textcolor{tabgreen}{+16.7} & \textcolor{tabgreen}{+21.7} & \textcolor{tabgreen}{+4.8} & \textcolor{tabgreen}{+4.8} & \textcolor{tabgreen}{+3.9} & \textcolor{tabgreen}{+4.2} & \textcolor{tabgreen}{+23.8} \\
\specialrule{\lightrulewidth}{0pt}{0pt}
\rowcolor{rowchampagne}
\multicolumn{13}{l}{\hspace{6pt}\rule[-3.5pt]{0pt}{12pt}$\rightarrow$ \textit{8.0B arch. 300B tokens teacher config ($\alpha$=0.6)}} \\
\rule{0pt}{10pt}pretrained teacher & 6.77 & 8.85 & 14.89 & 4.36 & 9.43 & 8.21 & 11.17 & 15.25 & 9.74 & 6.73 & 17.42 & 4.62 \\
distilled student & 11.72 & 15.28 & 20.69 & 6.27 & 9.12 & 11.90 & 16.94 & 23.45 & 13.68 & 8.92 & 24.18 & 9.91 \\
improvement (\%) & \textcolor{tabgreen}{+3.8} & \textcolor{tabgreen}{+4.5} & \textcolor{tabgreen}{+3.9} & \textcolor{tabgreen}{+5.1} & \textcolor{tabgreen}{+2.2} & \textcolor{tabgreen}{+12.2} & \textcolor{tabgreen}{+18.5} & \textcolor{tabgreen}{+4.2} & \textcolor{tabgreen}{+4.1} & \textcolor{tabgreen}{+3.5} & \textcolor{tabgreen}{+2.3} & \textcolor{tabgreen}{+18.3} \\

\end{xltabular}

\vspace{1cm}

\begin{xltabular}{\textwidth}{@{\hspace{12pt}} l *{15}{>{\centering\arraybackslash}X} @{\hspace{12pt}}}
\caption{\textbf{Full downstream accuracy evaluation results.} We include MMLU (MMLU), ARC-Easy (ARC), SciQ (SQ), OpenBookQA (OBA), MathQA (MQA), TruthfulQA (TQA), ANLI-R1 (ANLI), CommonsenseQA (CSQA), HellaSwag (HS), PIQA (PIQA), WinoGrande (WG), Social IQa (SIQA), LogiQA 2.0 (LQ), MedMCQA (MCQA), and RACE (RACE).} \label{tab:full_eval_acc} \\
\toprule
\rule[-3pt]{0pt}{12pt}accuracy ($\uparrow$) & MMLU & ARC & SQ & OBA & MQA & TQA & ANLI & CSQA & HS & PIQA & WG & SIQA & LQ & MCQA & RACE \\
\midrule
\endfirsthead
\multicolumn{16}{l}{\small\textit{Table~\ref{tab:full_eval_acc} continued from previous page}} \\[4pt]
\toprule
\rule[-3pt]{0pt}{12pt}accuracy ($\uparrow$) & MMLU & ARC & SQ & OBA & MQA & TQA & ANLI & CSQA & HS & PIQA & WG & SIQA & LQ & MCQA & RACE \\
\midrule
\noalign{\vskip 2pt}
\endhead
\midrule
\noalign{\vskip 4pt}
\multicolumn{16}{r}{\small\textit{Continued on next page}} \\
\endfoot
\bottomrule
\endlastfoot

\rowcolor{rowgray}
\multicolumn{16}{l}{\hspace{6pt}\rule[-3.5pt]{0pt}{12pt}$\rightarrow$ \textit{random chance}} \\
\rule{0pt}{10pt} & 25.0 & 25.0 & 25.0 & 25.0 & 20.0 & -- & 33.3 & 20.0 & 25.0 & 50.0 & 50.0 & 33.3 & 25.0 & 25.0 & 25.0 \\
\specialrule{\lightrulewidth}{0pt}{0pt}
\rowcolor{rowpurple}
\multicolumn{16}{l}{\hspace{6pt}\rule[-3.5pt]{0pt}{12pt}$\rightarrow$ \textit{standard pretrained student baseline}} \\
\rule{0pt}{10pt} & 25.4 & 62.8 & 87.2 & 33.4 & 23.7 & 21.4 & 32.9 & 18.9 & 51.1 & 71.1 & 54.0 & 38.7 & 23.6 & 24.5 & 32.7 \\
\specialrule{\lightrulewidth}{0pt}{0pt}
\rowcolor{rowgreen}
\multicolumn{16}{l}{\hspace{6pt}\rule[-3.5pt]{0pt}{12pt}$\rightarrow$ \textit{0.7B arch. 10B tokens teacher config ($\alpha$=0.2)}} \\
\rule{0pt}{10pt}pretrained teacher & 25.8 & 47.3 & 73.1 & 29.6 & 21.5 & 17.9 & 33.3 & 19.2 & 32.4 & 62.5 & 50.7 & 35.7 & 22.6 & 23.0 & 28.4 \\
distilled student & 25.8 & 60.3 & 86.1 & 32.2 & 22.7 & 20.6 & 33.7 & 19.5 & 47.5 & 70.6 & 53.7 & 40.1 & 22.2 & 27.8 & 32.4 \\
improvement (\%) & \textcolor{tabgreen}{+1.7} & \textcolor{tabred}{-4.0} & \textcolor{tabred}{-1.3} & \textcolor{tabred}{-3.6} & \textcolor{tabred}{-4.1} & \textcolor{tabred}{-4.0} & \textcolor{tabgreen}{+2.4} & \textcolor{tabgreen}{+3.0} & \textcolor{tabred}{-7.2} & \textcolor{tabred}{-0.7} & \textcolor{tabred}{-0.4} & \textcolor{tabgreen}{+3.6} & \textcolor{tabred}{-5.9} & \textcolor{tabgreen}{+13.2} & \textcolor{tabred}{-0.9} \\
\specialrule{\lightrulewidth}{0pt}{0pt}
\rowcolor{rowblue}
\multicolumn{16}{l}{\hspace{6pt}\rule[-3.5pt]{0pt}{12pt}$\rightarrow$ \textit{0.7B arch. 30B tokens teacher config ($\alpha$=0.2)}} \\
\rule{0pt}{10pt}pretrained teacher & 26.6 & 53.0 & 80.5 & 31.2 & 22.1 & 21.8 & 33.0 & 20.6 & 39.2 & 66.2 & 52.6 & 37.6 & 24.1 & 27.4 & 30.3 \\
distilled student & 26.5 & 60.7 & 87.1 & 34.2 & 23.6 & 21.9 & 34.9 & 20.6 & 49.2 & 70.9 & 54.2 & 40.3 & 22.8 & 24.8 & 32.9 \\
improvement (\%) & \textcolor{tabgreen}{+4.2} & \textcolor{tabred}{-3.4} & \textcolor{tabred}{-0.1} & \textcolor{tabgreen}{+2.4} & \textcolor{tabred}{-0.1} & \textcolor{tabgreen}{+2.3} & \textcolor{tabgreen}{+6.1} & \textcolor{tabgreen}{+9.1} & \textcolor{tabred}{-3.8} & \textcolor{tabred}{-0.2} & \textcolor{tabgreen}{+0.4} & \textcolor{tabgreen}{+4.1} & \textcolor{tabred}{-3.2} & \textcolor{tabgreen}{+1.1} & \textcolor{tabgreen}{+0.6} \\
\specialrule{\lightrulewidth}{0pt}{0pt}
\rowcolor{rowyellow}
\multicolumn{16}{l}{\hspace{6pt}\rule[-3.5pt]{0pt}{12pt}$\rightarrow$ \textit{0.7B arch. 50B tokens teacher config ($\alpha$=0.2)}} \\
\rule{0pt}{10pt}pretrained teacher & 26.1 & 55.6 & 81.6 & 33.0 & 22.4 & 18.4 & 33.2 & 20.3 & 43.0 & 67.8 & 52.1 & 38.6 & 23.5 & 24.5 & 31.3 \\
distilled student & 25.7 & 60.7 & 86.5 & 32.6 & 23.1 & 19.1 & 34.4 & 20.2 & 49.9 & 71.2 & 55.6 & 40.8 & 23.2 & 27.7 & 32.7 \\
improvement (\%) & \textcolor{tabgreen}{+1.1} & \textcolor{tabred}{-3.3} & \textcolor{tabred}{-0.8} & \textcolor{tabred}{-2.4} & \textcolor{tabred}{-2.3} & \textcolor{tabred}{-10.9} & \textcolor{tabgreen}{+4.6} & \textcolor{tabgreen}{+6.9} & \textcolor{tabred}{-2.3} & \textcolor{tabgreen}{+0.2} & \textcolor{tabgreen}{+2.9} & \textcolor{tabgreen}{+5.4} & \textcolor{tabred}{-1.6} & \textcolor{tabgreen}{+12.8} & 0.0 \\
\specialrule{\lightrulewidth}{0pt}{0pt}
\rowcolor{roworange}
\multicolumn{16}{l}{\hspace{6pt}\rule[-3.5pt]{0pt}{12pt}$\rightarrow$ \textit{0.7B arch. 80B tokens teacher config ($\alpha$=0.4)}} \\
\rule{0pt}{10pt}pretrained teacher & 26.5 & 58.8 & 84.4 & 29.8 & 22.7 & 18.1 & 33.9 & 19.2 & 45.5 & 70.1 & 51.1 & 37.0 & 23.7 & 25.3 & 31.8 \\
distilled student & 25.8 & 60.6 & 85.9 & 33.4 & 22.7 & 20.3 & 33.8 & 19.7 & 49.2 & 69.9 & 55.2 & 40.3 & 23.8 & 27.9 & 33.9 \\
improvement (\%) & \textcolor{tabgreen}{+1.4} & \textcolor{tabred}{-3.5} & \textcolor{tabred}{-1.5} & 0.0 & \textcolor{tabred}{-3.8} & \textcolor{tabred}{-5.1} & \textcolor{tabgreen}{+2.7} & \textcolor{tabgreen}{+3.9} & \textcolor{tabred}{-3.8} & \textcolor{tabred}{-1.7} & \textcolor{tabgreen}{+2.3} & \textcolor{tabgreen}{+4.2} & \textcolor{tabgreen}{+0.8} & \textcolor{tabgreen}{+13.6} & \textcolor{tabgreen}{+3.5} \\
\specialrule{\lightrulewidth}{0pt}{0pt}
\rowcolor{rowpink}
\multicolumn{16}{l}{\hspace{6pt}\rule[-3.5pt]{0pt}{12pt}$\rightarrow$ \textit{0.7B arch. 100B tokens teacher config ($\alpha$=0.4)}} \\
\rule{0pt}{10pt}pretrained teacher & 26.0 & 59.6 & 83.3 & 30.6 & 23.5 & 17.9 & 32.8 & 19.3 & 46.7 & 69.5 & 53.1 & 37.5 & 24.2 & 26.5 & 32.2 \\
distilled student & 27.1 & 61.2 & 85.9 & 32.4 & 23.8 & 21.8 & 33.7 & 19.8 & 49.6 & 70.7 & 53.7 & 40.2 & 24.2 & 24.0 & 32.8 \\
improvement (\%) & \textcolor{tabgreen}{+6.7} & \textcolor{tabred}{-2.5} & \textcolor{tabred}{-1.5} & \textcolor{tabred}{-3.0} & \textcolor{tabgreen}{+0.4} & \textcolor{tabgreen}{+1.7} & \textcolor{tabgreen}{+2.4} & \textcolor{tabgreen}{+4.8} & \textcolor{tabred}{-2.9} & \textcolor{tabred}{-0.5} & \textcolor{tabred}{-0.6} & \textcolor{tabgreen}{+3.8} & \textcolor{tabgreen}{+2.7} & \textcolor{tabred}{-1.9} & \textcolor{tabgreen}{+0.3} \\
\specialrule{\lightrulewidth}{0pt}{0pt}
\rowcolor{rowcyan}
\multicolumn{16}{l}{\hspace{6pt}\rule[-3.5pt]{0pt}{12pt}$\rightarrow$ \textit{0.7B arch. 300B tokens teacher config ($\alpha$=0.6)}} \\
\rule{0pt}{10pt}pretrained teacher & 24.9 & 62.7 & 86.7 & 33.2 & 25.3 & 17.0 & 33.5 & 17.7 & 50.9 & 70.6 & 55.2 & 39.7 & 23.3 & 28.0 & 33.1 \\
distilled student & 26.3 & 61.1 & 85.5 & 34.4 & 23.7 & 21.8 & 33.9 & 20.1 & 50.1 & 70.9 & 57.1 & 40.2 & 24.7 & 23.4 & 33.2 \\
improvement (\%) & \textcolor{tabgreen}{+3.6} & \textcolor{tabred}{-2.7} & \textcolor{tabred}{-1.9} & \textcolor{tabgreen}{+3.0} & \textcolor{tabgreen}{+0.1} & \textcolor{tabgreen}{+1.7} & \textcolor{tabgreen}{+3.0} & \textcolor{tabgreen}{+6.1} & \textcolor{tabred}{-2.0} & \textcolor{tabred}{-0.2} & \textcolor{tabgreen}{+5.7} & \textcolor{tabgreen}{+3.8} & \textcolor{tabgreen}{+4.9} & \textcolor{tabred}{-4.5} & \textcolor{tabgreen}{+1.5} \\
\specialrule{\lightrulewidth}{0pt}{0pt}
\rowcolor{rowlime}
\multicolumn{16}{l}{\hspace{6pt}\rule[-3.5pt]{0pt}{12pt}$\rightarrow$ \textit{1.7B arch. 10B tokens teacher config ($\alpha$=0.2)}} \\
\rule{0pt}{10pt}pretrained teacher & 26.0 & 50.5 & 79.5 & 29.4 & 22.2 & 22.3 & 33.8 & 20.6 & 38.0 & 65.5 & 51.5 & 36.9 & 22.4 & 26.5 & 30.3 \\
distilled student & 26.6 & 59.8 & 85.0 & 31.8 & 23.9 & 21.4 & 31.3 & 19.2 & 48.8 & 70.9 & 52.7 & 39.9 & 23.0 & 27.0 & 32.3 \\
improvement (\%) & \textcolor{tabgreen}{+4.7} & \textcolor{tabred}{-4.8} & \textcolor{tabred}{-2.5} & \textcolor{tabred}{-4.8} & \textcolor{tabgreen}{+1.1} & 0.0 & \textcolor{tabred}{-4.9} & \textcolor{tabgreen}{+1.7} & \textcolor{tabred}{-4.4} & \textcolor{tabred}{-0.2} & \textcolor{tabred}{-2.3} & \textcolor{tabgreen}{+3.2} & \textcolor{tabred}{-2.4} & \textcolor{tabgreen}{+10.1} & \textcolor{tabred}{-1.2} \\
\specialrule{\lightrulewidth}{0pt}{0pt}
\rowcolor{rowviolet}
\multicolumn{16}{l}{\hspace{6pt}\rule[-3.5pt]{0pt}{12pt}$\rightarrow$ \textit{1.7B arch. 30B tokens teacher config ($\alpha$=0.2)}} \\
\rule{0pt}{10pt}pretrained teacher & 25.0 & 60.0 & 85.5 & 34.2 & 23.8 & 20.6 & 32.7 & 20.1 & 47.4 & 70.0 & 52.8 & 39.8 & 23.4 & 23.7 & 32.0 \\
distilled student & 26.3 & 61.7 & 85.2 & 33.2 & 23.3 & 19.5 & 32.9 & 22.0 & 50.4 & 71.2 & 54.5 & 40.3 & 23.7 & 27.1 & 33.5 \\
improvement (\%) & \textcolor{tabgreen}{+3.8} & \textcolor{tabred}{-1.8} & \textcolor{tabred}{-2.3} & \textcolor{tabred}{-0.6} & \textcolor{tabred}{-1.4} & \textcolor{tabred}{-9.1} & 0.0 & \textcolor{tabgreen}{+16.5} & \textcolor{tabred}{-1.4} & \textcolor{tabgreen}{+0.2} & \textcolor{tabgreen}{+0.9} & \textcolor{tabgreen}{+4.1} & \textcolor{tabgreen}{+0.3} & \textcolor{tabgreen}{+10.3} & \textcolor{tabgreen}{+2.3} \\
\specialrule{\lightrulewidth}{0pt}{0pt}
\rowcolor{rowrose}
\multicolumn{16}{l}{\hspace{6pt}\rule[-3.5pt]{0pt}{12pt}$\rightarrow$ \textit{1.7B arch. 50B tokens teacher config ($\alpha$=0.2)}} \\
\rule{0pt}{10pt}pretrained teacher & 27.0 & 60.2 & 86.8 & 33.6 & 22.0 & 19.3 & 33.9 & 19.7 & 51.2 & 71.3 & 55.0 & 40.8 & 22.2 & 22.3 & 32.7 \\
distilled student & 26.8 & 62.3 & 86.9 & 34.8 & 23.2 & 19.7 & 34.6 & 19.3 & 51.1 & 70.9 & 55.5 & 41.1 & 24.7 & 25.4 & 32.3 \\
improvement (\%) & \textcolor{tabgreen}{+5.4} & \textcolor{tabred}{-0.8} & \textcolor{tabred}{-0.3} & \textcolor{tabgreen}{+4.2} & \textcolor{tabred}{-1.7} & \textcolor{tabred}{-8.0} & \textcolor{tabgreen}{+5.2} & \textcolor{tabgreen}{+2.2} & \textcolor{tabgreen}{+0.0} & \textcolor{tabred}{-0.2} & \textcolor{tabgreen}{+2.8} & \textcolor{tabgreen}{+6.2} & \textcolor{tabgreen}{+4.9} & \textcolor{tabgreen}{+3.4} & \textcolor{tabred}{-1.2} \\
\specialrule{\lightrulewidth}{0pt}{0pt}
\rowcolor{rowteal}
\multicolumn{16}{l}{\hspace{6pt}\rule[-3.5pt]{0pt}{12pt}$\rightarrow$ \textit{1.7B arch. 80B tokens teacher config ($\alpha$=0.2)}} \\
\rule{0pt}{10pt}pretrained teacher & 25.4 & 64.1 & 89.5 & 37.2 & 24.5 & 23.5 & 34.5 & 20.5 & 54.2 & 72.4 & 54.9 & 41.1 & 24.4 & 27.4 & 34.7 \\
distilled student & 26.3 & 63.3 & 87.5 & 35.8 & 24.1 & 22.5 & 34.8 & 19.2 & 51.5 & 71.9 & 55.7 & 41.4 & 22.1 & 22.7 & 33.4 \\
improvement (\%) & \textcolor{tabgreen}{+3.5} & \textcolor{tabgreen}{+0.8} & \textcolor{tabgreen}{+0.3} & \textcolor{tabgreen}{+7.2} & \textcolor{tabgreen}{+2.0} & \textcolor{tabgreen}{+5.1} & \textcolor{tabgreen}{+5.8} & \textcolor{tabgreen}{+1.3} & \textcolor{tabgreen}{+0.7} & \textcolor{tabgreen}{+1.1} & \textcolor{tabgreen}{+3.2} & \textcolor{tabgreen}{+7.0} & \textcolor{tabred}{-6.2} & \textcolor{tabred}{-7.4} & \textcolor{tabgreen}{+2.0} \\
\specialrule{\lightrulewidth}{0pt}{0pt}
\rowcolor{rowamber}
\multicolumn{16}{l}{\hspace{6pt}\rule[-3.5pt]{0pt}{12pt}$\rightarrow$ \textit{1.7B arch. 100B tokens teacher config ($\alpha$=0.5)}} \\
\rule{0pt}{10pt}pretrained teacher & 25.7 & 66.5 & 89.7 & 37.6 & 22.7 & 22.8 & 35.9 & 21.2 & 55.7 & 72.0 & 56.5 & 39.9 & 23.8 & 24.0 & 34.5 \\
distilled student & 25.5 & 62.5 & 87.7 & 36.4 & 23.4 & 19.0 & 33.6 & 20.5 & 51.6 & 72.4 & 54.9 & 40.5 & 23.7 & 26.8 & 33.8 \\
improvement (\%) & \textcolor{tabgreen}{+0.3} & \textcolor{tabred}{-0.4} & \textcolor{tabgreen}{+0.6} & \textcolor{tabgreen}{+9.0} & \textcolor{tabred}{-1.1} & \textcolor{tabred}{-11.4} & \textcolor{tabgreen}{+2.1} & \textcolor{tabgreen}{+8.2} & \textcolor{tabgreen}{+0.9} & \textcolor{tabgreen}{+1.8} & \textcolor{tabgreen}{+1.8} & \textcolor{tabgreen}{+4.6} & \textcolor{tabgreen}{+0.3} & \textcolor{tabgreen}{+9.2} & \textcolor{tabgreen}{+3.2} \\
\specialrule{\lightrulewidth}{0pt}{0pt}
\rowcolor{rowsky}
\multicolumn{16}{l}{\hspace{6pt}\rule[-3.5pt]{0pt}{12pt}$\rightarrow$ \textit{1.7B arch. 300B tokens teacher config ($\alpha$=0.8)}} \\
\rule{0pt}{10pt}pretrained teacher & 26.1 & 67.5 & 93.1 & 40.0 & 23.8 & 20.7 & 35.0 & 21.9 & 59.4 & 73.3 & 58.7 & 41.0 & 26.3 & 26.8 & 34.8 \\
distilled student & 26.6 & 64.7 & 88.3 & 36.8 & 25.0 & 21.8 & 35.9 & 19.2 & 52.6 & 71.7 & 57.8 & 39.6 & 24.2 & 26.3 & 34.3 \\
improvement (\%) & \textcolor{tabgreen}{+4.8} & \textcolor{tabgreen}{+3.1} & \textcolor{tabgreen}{+1.3} & \textcolor{tabgreen}{+10.2} & \textcolor{tabgreen}{+5.7} & \textcolor{tabgreen}{+1.7} & \textcolor{tabgreen}{+9.1} & \textcolor{tabgreen}{+1.3} & \textcolor{tabgreen}{+2.9} & \textcolor{tabgreen}{+0.8} & \textcolor{tabgreen}{+7.0} & \textcolor{tabgreen}{+2.2} & \textcolor{tabgreen}{+2.7} & \textcolor{tabgreen}{+7.4} & \textcolor{tabgreen}{+4.7} \\
\specialrule{\lightrulewidth}{0pt}{0pt}
\rowcolor{rowfuchsia}
\multicolumn{16}{l}{\hspace{6pt}\rule[-3.5pt]{0pt}{12pt}$\rightarrow$ \textit{3.8B arch. 10B tokens teacher config ($\alpha$=0.5)}} \\
\rule{0pt}{10pt}pretrained teacher & 25.6 & 57.1 & 82.5 & 32.4 & 22.7 & 22.9 & 33.1 & 19.7 & 43.4 & 68.0 & 51.9 & 38.3 & 22.3 & 25.4 & 32.5 \\
distilled student & 26.5 & 59.5 & 85.9 & 31.2 & 23.5 & 23.1 & 34.4 & 18.7 & 47.1 & 70.7 & 52.3 & 39.7 & 23.0 & 24.2 & 32.5 \\
improvement (\%) & \textcolor{tabgreen}{+4.2} & \textcolor{tabred}{-5.2} & \textcolor{tabred}{-1.5} & \textcolor{tabred}{-6.6} & \textcolor{tabred}{-0.8} & \textcolor{tabgreen}{+8.0} & \textcolor{tabgreen}{+4.6} & \textcolor{tabred}{-1.3} & \textcolor{tabred}{-7.9} & \textcolor{tabred}{-0.5} & \textcolor{tabred}{-3.1} & \textcolor{tabgreen}{+2.5} & \textcolor{tabred}{-2.7} & \textcolor{tabred}{-1.4} & \textcolor{tabred}{-0.6} \\
\specialrule{\lightrulewidth}{0pt}{0pt}
\rowcolor{rowemerald}
\multicolumn{16}{l}{\hspace{6pt}\rule[-3.5pt]{0pt}{12pt}$\rightarrow$ \textit{3.8B arch. 30B tokens teacher config ($\alpha$=0.5)}} \\
\rule{0pt}{10pt}pretrained teacher & 24.8 & 66.8 & 89.4 & 38.6 & 22.5 & 18.5 & 35.5 & 19.9 & 54.8 & 72.3 & 55.7 & 40.8 & 22.7 & 24.3 & 35.2 \\
distilled student & 26.7 & 63.0 & 88.0 & 35.4 & 22.7 & 21.9 & 34.0 & 19.9 & 51.5 & 71.8 & 55.7 & 41.6 & 23.5 & 26.8 & 32.1 \\
improvement (\%) & \textcolor{tabgreen}{+5.2} & \textcolor{tabgreen}{+0.3} & \textcolor{tabgreen}{+0.9} & \textcolor{tabgreen}{+6.0} & \textcolor{tabred}{-4.0} & \textcolor{tabgreen}{+2.3} & \textcolor{tabgreen}{+3.3} & \textcolor{tabgreen}{+5.2} & \textcolor{tabgreen}{+0.7} & \textcolor{tabgreen}{+1.1} & \textcolor{tabgreen}{+3.2} & \textcolor{tabgreen}{+7.5} & \textcolor{tabred}{-0.3} & \textcolor{tabgreen}{+9.5} & \textcolor{tabred}{-2.0} \\
\specialrule{\lightrulewidth}{0pt}{0pt}
\rowcolor{rowindigo}
\multicolumn{16}{l}{\hspace{6pt}\rule[-3.5pt]{0pt}{12pt}$\rightarrow$ \textit{3.8B arch. 50B tokens teacher config ($\alpha$=0.6)}} \\
\rule{0pt}{10pt}pretrained teacher & 26.3 & 67.3 & 89.4 & 39.4 & 25.0 & 19.7 & 35.1 & 19.0 & 58.9 & 74.2 & 58.2 & 42.4 & 23.5 & 30.2 & 35.4 \\
distilled student & 25.9 & 65.5 & 88.9 & 35.6 & 23.7 & 22.4 & 34.9 & 18.9 & 52.6 & 72.4 & 55.9 & 41.2 & 23.5 & 26.8 & 33.6 \\
improvement (\%) & \textcolor{tabgreen}{+1.9} & \textcolor{tabgreen}{+4.4} & \textcolor{tabgreen}{+1.9} & \textcolor{tabgreen}{+6.6} & \textcolor{tabgreen}{+0.1} & \textcolor{tabgreen}{+4.6} & \textcolor{tabgreen}{+6.1} & 0.0 & \textcolor{tabgreen}{+3.0} & \textcolor{tabgreen}{+1.9} & \textcolor{tabgreen}{+3.5} & \textcolor{tabgreen}{+6.5} & \textcolor{tabred}{-0.5} & \textcolor{tabgreen}{+9.2} & \textcolor{tabgreen}{+2.6} \\
\specialrule{\lightrulewidth}{0pt}{0pt}
\rowcolor{rowred}
\multicolumn{16}{l}{\hspace{6pt}\rule[-3.5pt]{0pt}{12pt}$\rightarrow$ \textit{3.8B arch. 80B tokens teacher config ($\alpha$=0.8)}} \\
\rule{0pt}{10pt}pretrained teacher & 25.3 & 70.3 & 92.1 & 41.4 & 25.4 & 21.8 & 33.2 & 20.1 & 61.9 & 74.6 & 59.9 & 42.0 & 23.9 & 26.3 & 35.8 \\
distilled student & 25.8 & 64.6 & 88.5 & 35.2 & 24.2 & 20.3 & 34.0 & 21.5 & 52.9 & 72.3 & 56.3 & 40.1 & 23.9 & 27.4 & 33.6 \\
improvement (\%) & \textcolor{tabgreen}{+1.5} & \textcolor{tabgreen}{+2.8} & \textcolor{tabgreen}{+1.5} & \textcolor{tabgreen}{+5.4} & \textcolor{tabgreen}{+2.3} & \textcolor{tabred}{-5.1} & \textcolor{tabgreen}{+3.3} & \textcolor{tabgreen}{+13.4} & \textcolor{tabgreen}{+3.5} & \textcolor{tabgreen}{+1.7} & \textcolor{tabgreen}{+4.2} & \textcolor{tabgreen}{+3.6} & \textcolor{tabgreen}{+1.1} & \textcolor{tabgreen}{+11.9} & \textcolor{tabgreen}{+2.6} \\
\specialrule{\lightrulewidth}{0pt}{0pt}
\rowcolor{rowmint}
\multicolumn{16}{l}{\hspace{6pt}\rule[-3.5pt]{0pt}{12pt}$\rightarrow$ \textit{3.8B arch. 100B tokens teacher config ($\alpha$=0.6)}} \\
\rule{0pt}{10pt}pretrained teacher & 25.6 & 70.3 & 92.4 & 40.0 & 25.2 & 21.3 & 33.9 & 17.9 & 63.2 & 75.2 & 62.4 & 42.7 & 23.2 & 26.4 & 36.4 \\
distilled student & 25.1 & 65.1 & 88.8 & 37.4 & 23.7 & 22.6 & 32.0 & 19.7 & 52.9 & 72.1 & 57.5 & 41.2 & 23.7 & 25.7 & 33.8 \\
improvement (\%) & \textcolor{tabred}{-1.3} & \textcolor{tabgreen}{+3.6} & \textcolor{tabgreen}{+1.8} & \textcolor{tabgreen}{+12.0} & \textcolor{tabgreen}{+0.3} & \textcolor{tabgreen}{+5.7} & \textcolor{tabred}{-2.7} & \textcolor{tabgreen}{+3.9} & \textcolor{tabgreen}{+3.5} & \textcolor{tabgreen}{+1.5} & \textcolor{tabgreen}{+6.6} & \textcolor{tabgreen}{+6.5} & \textcolor{tabgreen}{+0.3} & \textcolor{tabgreen}{+5.0} & \textcolor{tabgreen}{+3.2} \\
\specialrule{\lightrulewidth}{0pt}{0pt}
\rowcolor{rowpeach}
\multicolumn{16}{l}{\hspace{6pt}\rule[-3.5pt]{0pt}{12pt}$\rightarrow$ \textit{3.8B arch. 300B tokens teacher config ($\alpha$=0.5)}} \\
\rule{0pt}{10pt}pretrained teacher & 25.6 & 72.1 & 94.1 & 42.8 & 26.9 & 25.3 & 34.0 & 19.5 & 66.8 & 76.9 & 66.1 & 42.8 & 23.0 & 29.0 & 36.4 \\
distilled student & 25.1 & 65.7 & 88.6 & 37.4 & 23.6 & 21.4 & 32.6 & 19.4 & 53.1 & 71.9 & 56.6 & 41.2 & 23.5 & 28.3 & 33.5 \\
improvement (\%) & \textcolor{tabred}{-1.0} & \textcolor{tabgreen}{+4.6} & \textcolor{tabgreen}{+1.6} & \textcolor{tabgreen}{+12.0} & \textcolor{tabred}{-0.4} & 0.0 & \textcolor{tabred}{-0.9} & \textcolor{tabgreen}{+2.6} & \textcolor{tabgreen}{+3.8} & \textcolor{tabgreen}{+1.2} & \textcolor{tabgreen}{+4.8} & \textcolor{tabgreen}{+6.5} & \textcolor{tabred}{-0.5} & \textcolor{tabgreen}{+15.5} & \textcolor{tabgreen}{+2.3} \\
\specialrule{\lightrulewidth}{0pt}{0pt}
\rowcolor{rowlavender}
\multicolumn{16}{l}{\hspace{6pt}\rule[-3.5pt]{0pt}{12pt}$\rightarrow$ \textit{8.0B arch. 10B tokens teacher config ($\alpha$=0.4)}} \\
\rule{0pt}{10pt}pretrained teacher & 25.2 & 59.6 & 85.5 & 33.6 & 23.7 & 20.1 & 31.9 & 20.1 & 48.4 & 70.5 & 53.1 & 38.7 & 22.3 & 25.8 & 32.3 \\
distilled student & 26.7 & 60.9 & 85.9 & 34.4 & 23.4 & 22.5 & 33.4 & 18.4 & 49.8 & 70.8 & 53.4 & 40.8 & 22.3 & 26.7 & 33.3 \\
improvement (\%) & \textcolor{tabgreen}{+5.1} & \textcolor{tabred}{-3.1} & \textcolor{tabred}{-1.5} & \textcolor{tabgreen}{+3.0} & \textcolor{tabred}{-1.1} & \textcolor{tabgreen}{+5.1} & \textcolor{tabgreen}{+1.5} & \textcolor{tabred}{-2.6} & \textcolor{tabred}{-2.5} & \textcolor{tabred}{-0.4} & \textcolor{tabred}{-1.2} & \textcolor{tabgreen}{+5.4} & \textcolor{tabred}{-5.4} & \textcolor{tabgreen}{+9.0} & \textcolor{tabgreen}{+1.8} \\
\specialrule{\lightrulewidth}{0pt}{0pt}
\rowcolor{rowapricot}
\multicolumn{16}{l}{\hspace{6pt}\rule[-3.5pt]{0pt}{12pt}$\rightarrow$ \textit{8.0B arch. 30B tokens teacher config ($\alpha$=0.6)}} \\
\rule{0pt}{10pt}pretrained teacher & 25.4 & 66.8 & 90.5 & 38.2 & 24.5 & 22.4 & 35.1 & 19.3 & 59.6 & 75.0 & 59.7 & 40.7 & 24.2 & 25.1 & 34.7 \\
distilled student & 26.0 & 64.1 & 87.1 & 33.2 & 24.5 & 23.6 & 34.4 & 22.3 & 52.7 & 72.0 & 56.0 & 40.7 & 24.0 & 26.4 & 33.4 \\
improvement (\%) & \textcolor{tabgreen}{+2.2} & \textcolor{tabgreen}{+2.1} & \textcolor{tabred}{-0.1} & \textcolor{tabred}{-0.6} & \textcolor{tabgreen}{+3.4} & \textcolor{tabgreen}{+10.3} & \textcolor{tabgreen}{+4.6} & \textcolor{tabgreen}{+17.7} & \textcolor{tabgreen}{+3.2} & \textcolor{tabgreen}{+1.3} & \textcolor{tabgreen}{+3.8} & \textcolor{tabgreen}{+5.2} & \textcolor{tabgreen}{+1.9} & \textcolor{tabgreen}{+7.7} & \textcolor{tabgreen}{+2.0} \\
\specialrule{\lightrulewidth}{0pt}{0pt}
\rowcolor{rowaqua}
\multicolumn{16}{l}{\hspace{6pt}\rule[-3.5pt]{0pt}{12pt}$\rightarrow$ \textit{8.0B arch. 50B tokens teacher config ($\alpha$=0.8)}} \\
\rule{0pt}{10pt}pretrained teacher & 25.1 & 68.8 & 91.6 & 40.8 & 24.4 & 22.2 & 34.9 & 19.7 & 63.1 & 75.9 & 59.4 & 42.3 & 23.8 & 26.7 & 37.1 \\
distilled student & 26.6 & 63.2 & 87.1 & 34.8 & 23.2 & 20.8 & 34.0 & 20.3 & 53.2 & 72.4 & 57.2 & 42.6 & 24.0 & 26.0 & 33.5 \\
improvement (\%) & \textcolor{tabgreen}{+4.8} & \textcolor{tabgreen}{+0.6} & \textcolor{tabred}{-0.1} & \textcolor{tabgreen}{+4.2} & \textcolor{tabred}{-1.7} & \textcolor{tabred}{-2.9} & \textcolor{tabgreen}{+3.3} & \textcolor{tabgreen}{+7.4} & \textcolor{tabgreen}{+4.0} & \textcolor{tabgreen}{+1.9} & \textcolor{tabgreen}{+6.0} & \textcolor{tabgreen}{+10.1} & \textcolor{tabgreen}{+1.6} & \textcolor{tabgreen}{+6.0} & \textcolor{tabgreen}{+2.3} \\
\specialrule{\lightrulewidth}{0pt}{0pt}
\rowcolor{rowcoral}
\multicolumn{16}{l}{\hspace{6pt}\rule[-3.5pt]{0pt}{12pt}$\rightarrow$ \textit{8.0B arch. 80B tokens teacher config ($\alpha$=1.0)}} \\
\rule{0pt}{10pt}pretrained teacher & 24.8 & 72.7 & 93.1 & 42.4 & 25.8 & 22.6 & 33.3 & 19.8 & 65.7 & 77.1 & 62.7 & 42.4 & 25.3 & 30.9 & 34.7 \\
distilled student & 26.2 & 64.2 & 88.6 & 37.2 & 23.9 & 20.4 & 34.4 & 20.9 & 53.2 & 71.9 & 54.6 & 40.9 & 23.6 & 28.0 & 33.7 \\
improvement (\%) & \textcolor{tabgreen}{+3.3} & \textcolor{tabgreen}{+2.2} & \textcolor{tabgreen}{+1.6} & \textcolor{tabgreen}{+11.4} & \textcolor{tabgreen}{+1.0} & \textcolor{tabred}{-4.6} & \textcolor{tabgreen}{+4.6} & \textcolor{tabgreen}{+10.4} & \textcolor{tabgreen}{+4.1} & \textcolor{tabgreen}{+1.1} & \textcolor{tabgreen}{+1.2} & \textcolor{tabgreen}{+5.7} & 0.0 & \textcolor{tabgreen}{+14.3} & \textcolor{tabgreen}{+2.9} \\
\specialrule{\lightrulewidth}{0pt}{0pt}
\rowcolor{rowperiwinkle}
\multicolumn{16}{l}{\hspace{6pt}\rule[-3.5pt]{0pt}{12pt}$\rightarrow$ \textit{8.0B arch. 100B tokens teacher config ($\alpha$=1.0)}} \\
\rule{0pt}{10pt}pretrained teacher & 26.8 & 73.5 & 94.4 & 42.6 & 25.9 & 22.2 & 33.7 & 20.6 & 67.0 & 76.6 & 64.9 & 43.5 & 23.9 & 26.6 & 35.9 \\
distilled student & 26.3 & 62.2 & 88.3 & 37.2 & 24.8 & 22.2 & 34.3 & 18.5 & 53.7 & 72.0 & 55.2 & 41.4 & 23.0 & 26.2 & 33.8 \\
improvement (\%) & \textcolor{tabgreen}{+3.5} & \textcolor{tabred}{-0.9} & \textcolor{tabgreen}{+1.3} & \textcolor{tabgreen}{+11.4} & \textcolor{tabgreen}{+4.8} & \textcolor{tabgreen}{+3.4} & \textcolor{tabgreen}{+4.3} & \textcolor{tabred}{-2.2} & \textcolor{tabgreen}{+5.2} & \textcolor{tabgreen}{+1.4} & \textcolor{tabgreen}{+2.3} & \textcolor{tabgreen}{+6.9} & \textcolor{tabred}{-2.7} & \textcolor{tabgreen}{+6.6} & \textcolor{tabgreen}{+3.2} \\
\specialrule{\lightrulewidth}{0pt}{0pt}
\rowcolor{rowchampagne}
\multicolumn{16}{l}{\hspace{6pt}\rule[-3.5pt]{0pt}{12pt}$\rightarrow$ \textit{8.0B arch. 300B tokens teacher config ($\alpha$=0.8)}} \\
\rule{0pt}{10pt}pretrained teacher & 27.2 & 76.5 & 95.1 & 43.8 & 27.5 & 27.3 & 33.9 & 20.8 & 70.6 & 77.4 & 65.4 & 43.8 & 24.2 & 28.1 & 37.9 \\
distilled student & 26.4 & 63.3 & 86.8 & 36.6 & 23.2 & 23.0 & 30.3 & 20.1 & 53.6 & 71.5 & 56.4 & 40.6 & 24.4 & 25.1 & 33.9 \\
improvement (\%) & \textcolor{tabgreen}{+3.8} & \textcolor{tabgreen}{+0.9} & \textcolor{tabred}{-0.5} & \textcolor{tabgreen}{+9.6} & \textcolor{tabred}{-1.8} & \textcolor{tabgreen}{+7.4} & \textcolor{tabred}{-7.9} & \textcolor{tabgreen}{+6.1} & \textcolor{tabgreen}{+4.9} & \textcolor{tabgreen}{+0.6} & \textcolor{tabgreen}{+4.4} & \textcolor{tabgreen}{+5.0} & \textcolor{tabgreen}{+3.5} & \textcolor{tabgreen}{+2.2} & \textcolor{tabgreen}{+3.5} \\

\end{xltabular}

\end{document}